\documentclass{article}

\usepackage[main, final]{neurips_2025}

\usepackage[utf8]{inputenc} 
\usepackage[T1]{fontenc}    
\usepackage{hyperref}       
\usepackage{url}            
\usepackage{booktabs}       
\usepackage{siunitx}        
\usepackage{amsfonts}       
\usepackage{caption}        
\usepackage{newunicodechar}
\newunicodechar{₂}{$_2$}
\usepackage{amsthm}         
\usepackage{bm}             
\usepackage{bbm}            
\usepackage{multirow}       
\usepackage{tabularx}       
\usepackage{nicefrac}       
\usepackage{microtype}      
\usepackage{xcolor}         

\usepackage{amsmath}
\usepackage{amssymb}
\usepackage{mathtools}
\usepackage{amsthm}
\usepackage[utf8]{inputenc}
\newtheorem{theorem}{Theorem}[section] 
\usepackage{enumitem}
\theoremstyle{remark}

\newtheorem{proposition}{Proposition}[section]
\newtheorem{lemma}{Lemma}
\newtheorem{assumption}{Assumption}
\usepackage{tikz}
\usetikzlibrary{decorations.pathreplacing,calligraphy}

\newcommand{\V}[1]{\bm{#1} } 

\newcommand{\mR}{\mathbb{R}}

\newcommand{\lb}{\left(}
\newcommand{\rb}{\right)}
\newcommand{\lbb}{\left\{}
\newcommand{\rbb}{\right\}}

\newcommand{\Req}[1]{(\ref{eq:#1})}
\newcommand{\BReq}[1]{Equation\ (\ref{eq:#1})}
\newcommand{\NReq}[1]{(\ref{eq:#1})}
\newcommand{\NReqs}[2]{(\ref{eq:#1}), (\ref{eq:#2})}

\newcommand{\Rfig}[1]{Fig.\ \ref{fig:#1}}
\newcommand{\Rfigs}[2]{Figs.\ \ref{fig:#1} and \ref{fig:#2}}

\newcommand{\Leq}[1]{\label{eq:#1}}
\newcommand{\Rsec}[1]{Section\ \ref{sec:#1}}
\newcommand{\NRsec}[1]{\ref{sec:#1}}
\newcommand{\Lsec}[1]{\label{sec:#1}}
\newcommand{\be}{\begin{eqnarray}}
\newcommand{\ee}{\end{eqnarray}}
\newcommand{\ba}{\begin{array}}
\newcommand{\ea}{\end{array}}

\newcommand{\subbe}{\begin{subequations}}
\newcommand{\subee}{\end{subequations}}

\title{Neural Collapse in Cumulative Link Models \\ for Ordinal Regression: \\ An Analysis with Unconstrained Feature Model
}

\author{%
  Chuang~Ma$^{1}$ \quad Tomoyuki~Obuchi$^{1,2}$ \quad Toshiyuki~Tanaka$^{1}$\\
$^{1}$Kyoto University, $^{2}$RIKEN AIP,\\
  \texttt{\{ma.chuang.52h@st, obuchi@i, tt@i\}.kyoto-u.ac.jp}%
}

\begin{document}

\maketitle

\begin{abstract}
A phenomenon known as ``Neural Collapse (NC)'' in deep classification tasks, in which the penultimate-layer features and the final classifiers exhibit an extremely simple geometric structure, has recently attracted considerable attention, with the expectation that it can deepen our understanding of how deep neural networks behave. The Unconstrained Feature Model (UFM) has been proposed to explain NC theoretically, and there emerges a growing body of work that extends NC to tasks other than classification and leverages it for practical applications. In this study, we investigate whether a similar phenomenon arises in deep Ordinal Regression (OR) tasks, via combining the cumulative link model for OR and UFM. We show that a phenomenon we call Ordinal Neural Collapse (ONC) indeed emerges and is characterized by the following three properties: (ONC1) all optimal features in the same class collapse to their within-class mean when regularization is applied; (ONC2) these class means align with the classifier, meaning that they collapse onto a one-dimensional subspace; (ONC3) the optimal latent variables (corresponding to logits or preactivations in classification tasks) are aligned according to the class order, and in particular, in the zero-regularization limit, a highly local and simple geometric relationship emerges between the latent variables and the threshold values. We prove these properties analytically within the UFM framework with fixed threshold values and corroborate them empirically across a variety of datasets. We also discuss how these insights can be leveraged in OR, highlighting the use of fixed thresholds.
\end{abstract}

\section{Introduction}\Lsec{Introduction}
In classification tasks on balanced datasets, it has been observed that, after sufficient training, the outputs (or features) of the penultimate layer and the final classifier weights in sufficiently expressive Deep Neural Networks (DNNs) exhibit a remarkably simple symmetric structure. \citet{doi:10.1073/pnas.2015509117} conducted thorough experiments across architectures and datasets to corroborate this phenomenon, and identified its four intertwined signatures, which are summarized as \textbf{Neural Collapse (NC)}: (NC1) all feature vectors from the same class converge exactly onto their shared class mean, extinguishing within-class variance; (NC2) once these class means are recentered at the global mean, they occupy the vertices of a maximally symmetric Simplex Equiangular Tight Frame (Simplex ETF); (NC3) each classifier weight vector becomes parallel to its corresponding class mean vector, reflecting a self-dual alignment; (NC4) the network simply classifies by nearest class mean. NC, being considered to offer a valuable clue toward understanding DNNs, has inspired a number of theoretical studies~\citep{mixon2022neural,doi:10.1073/pnas.2103091118,zhu2021geometric,wojtowytsch2020emergence,lu2022neural} which introduced  {\bf Unconstrained Feature Model (UFM)} being a central theoretical tool in this context.

UFM has allowed extending NC to broader problem settings and facilitated the analytical investigation of its properties~\citep{zhou2022all,thrampoulidis2022imbalance,dang2023neural,dang2024neural,li2023neural}. For example, \citet{andriopoulos2024prevalence} applied this framework to multivariate regression, finding a phenomenon called Neural Regression Collapse (NRC) in which features collapse to a target subspace and weight vectors align accordingly. The proliferation of these extensions suggests that NC is a universal phenomenon in DNNs.  

Beyond classification and regression, there is a task called \textbf{Ordinal Regression (OR)} which aims to solve prediction tasks where labels are discrete categories with a natural order. Unlike classification, not all misclassifications are equally wrong in OR tasks; unlike regression, label values in OR do not bear quantitative information and only their ordering relationship is essential. A common approach to OR problems is to use threshold models~\citep{article} which assume that an unobserved continuous latent variable generates the observed ordinal response: the map to latent space is traditionally assumed to be linear with respect to (w.r.t.) input datapoints~\citep{Herbrich}. Even within this framework, more challenging scenarios can be addressed by applying an appropriate transformation (feature extractor) to the input datapoints, and the effectiveness of DNNs as feature extractors has been reported~\citep{Dorado-Moreno,vargas2020cumulative}. The latent variable in the threshold models occupies the same role as the logit in classification networks, prompting the question of whether phenomena similar to NC may also emerge in the feature space of OR.

To address this question, in this paper we explore phenomena analogous to NC within the context of OR. To that end, we adopt the Cumulative Link Model (CLM)~\citep{Agresti2010}---a classical type of threshold models---and analyze it in combination with UFM. As a result, we find that such a phenomenon indeed occurs when the $\ell_2$ regularization is applied, and we name it Ordinal Neural Collapse (ONC). ONC is characterized by the following three properties:
\begin{description}
\item[\textbf{(ONC1)}] \textbf{Within-class Mean Collapse:} \textit{all optimal features in the same class collapse to their within-class mean.}

\item[\textbf{(ONC2)}] \textbf{Collapse to One-Dimensional Subspace:} \textit{these class means align with the classifier, meaning that they collapse onto a one-dimensional subspace.}

\item[\textbf{(ONC3)}] \textbf{Collapse to Ordinal Structure:} \textit{the optimal latent variables are aligned according to the class order, and in particular, in the zero-regularization limit, a highly local and simple relationship emerges between the latent variables and the threshold values.}
\end{description}
We note that this result is obtained under the assumption that all the threshold values of CLM are fixed. Although it is not necessarily a standard assumption in recent studies, we argue that it is necessary for the emergence of ONC and, moreover, can be a practically meaningful assumption. A discussion about this point will be given later in \Rsec{Discussion}.

We also validated ONC through experiments using five imbalanced ordinal datasets and a DNN architecture. The result provides clear empirical evidence of ONC under fixed threshold values. Furthermore, our experiments with learnable thresholds still exhibit ONC, implying its robustness.

\section{Related work}
We cite here only recent results that are particularly relevant to the present work.

\noindent\textbf{UFM and related issues.}
UFM and the related models were proposed almost concurrently in a number of pieces of work~\citep{mixon2022neural,doi:10.1073/pnas.2103091118,zhu2021geometric,wojtowytsch2020emergence,lu2022neural}. The core idea of UFM lies in decoupling the features from the data by treating the output of a specific layer, typically the penultimate layer, as free learnable variables, while explicitly modeling the nonlinear functions and weight vectors in the subsequent layers. This simplifying assumption enables us to analytically derive nontrivial results. Even in such a simplified model, the analysis can remain nontrivial. For example, UFM generally admits multiple local minima in its loss landscape. Among them, it was shown by \cite{zhu2021geometric} that only the global minimum exhibits the NC structure in the case of balanced classification. Departing from the typical analysis of the single-layer UFM, the first investigation of UFM with multiple layers was carried out by \cite{tirer2022extended}, finding that NC still emerges as the unique global optimum. Furthermore, to understand the phenomenon called Deep NC, in which NC propagates not only to the final layer but also to intermediate layers~\citep{hui2022limitations,he2023law,rangamani2023feature,parker2023neural,masarczyk2023tunnel}, \citet{sukenik2023deep} extended UFM to multiple nonlinear layers and proved that, in the case of binary classification, Deep NC emerges as the unique global optimum. Their follow-up study \citep{sukenik2024neural} demonstrated that, in multi-layer architectures, increasing the number of classes causes Deep NC to cease being the optimal solution.

\noindent\textbf{NC extensions under varied settings on classification.} 
In classification, the NC framework has been adapted to different settings from the balanced case. \citet{zhou2022all} showed that cross‐entropy, focal loss, label smoothing, and even mean squared error lead to the same NC geometry using UFM. 
\citet{thrampoulidis2022imbalance} and \citet{JMLR:v25:23-1215} found that with cross‐entropy loss and $\ell_2$ regularization, class imbalance does not prevent NC1 but the global geometry generalizes from a Simplex ETF to a more general structure. In extremely imbalanced cases, \cite{doi:10.1073/pnas.2103091118} found a ``minority collapse'' phenomenon where minority-class features collapse onto a single point. \citet{dang2023neural} proved in deep linear UFM that every global minimizer forms orthogonal directions whose magnitudes scale proportionally to class sample sizes, and \citet{dang2024neural} proved essentially the same statement for rectified linear unit (ReLU) UFM.

\noindent\textbf{NC extensions beyond classification.} \citet{li2023neural} extended NC to multi‐label classification by showing that multi‐label embeddings lie in the linear span of label‐means. \citet{andriopoulos2024prevalence} generalized NC to multivariate regression to find NRC. \citet{wu2024linguistic} introduced the concept of ``linguistic collapse'' in large-scale language models, showing that token embeddings tend toward an approximately uniform distribution on the hypersphere as model scale increases. Furthermore, \citet{nguyen2024mitigating} demonstrated that NC-style embedding collapse also occurs in diffusion models. NC-like phenomena have also been observed in self-supervised learning~\citep{ben2023reverse} and in transfer learning~\citep{galanti2021role,li2022understanding}: the latter studies discussed the relationship between the degree of NC and transfer performance, and also proposed some strategy for leveraging NC insights to improve generalization performance.

\section{Formulation}\Lsec{Formulation}
\paragraph{OR.}
An OR task is formulated as follows. 
Let $\mathcal{X}$ be an input space and $\mathcal{Y}=\{1,2,\ldots,Q\}$ be an ordered label set with ordering $1<2<\cdots<Q$. Given a training set $D=\{(\bm{x}_{i},y_{i})\}_{i=1}^{N}$ with $(\bm{x}_{i},y_i)\in\mathcal{X}\times\mathcal{Y}$, our goal is to learn an order-respecting mapping $r:\mathcal{X}\to\mathcal{Y}$. For each label $q$, we let $D_q=\{(\bm{x}_{i},y_{i})\in D\mid y_{i}=q\}$ and $n_q$ its size, so that $\sum_{q=1}^{Q} n_q=N$ holds. 

\paragraph{CLM.}
To express the ordinal structure, threshold models introduce a latent variable $z\in \mR$ and also a strictly ordered set of ``thresholds'' $\V{b}=(b_0,b_1,\ldots,b_Q)$ which partitions the $z$-axis.
One typically assumes $(b_0,b_Q) = (-\infty,\infty)$ to partition $\mathbb{R}$ properly, and thus each interval is uniquely associated with one category
via the decision rule 
$y=q\Longleftrightarrow z\in(b_{q-1},b_{q}]$.

In CLMs, the probability of a specific category is expressed through a cumulative probability $P\bigl(y\le q \mid z \bigr)$ conditioned on the latent variable $z$, which is modeled by using a strictly monotone inverse link function \(g:\mathbb{R}\to(0,1)\) as 
\be
P\bigl(y\le q \mid z \bigr)
  =
  g\bigl(b_q - z \bigr).
\Leq{z2P}
\ee
There are several typical choices for $g$, including the logistic function $g(x)=(1+e^{-x})^{-1}$, the normal cumulative distribution function (CDF) $g(x)=\Phi(x)=\int_{-\infty}^{x}e^{-\frac{1}{2}z^2 }dz/\sqrt{2\pi}$, and the Gumbel CDF $g(x)=1-e^{-e^x}$, which correspond to the logit, probit, and clog-log models, respectively.

An input datapoint $\V{x}$ is transformed to a value $z$ in the latent space through a certain map. When using a feature extractor such as DNNs \citep{vargas2020cumulative}, the map is expressed as 
\be
z=f_{\bm{w},\theta}(\bm{x})=\bm{w}^{\top}\V{h}_{\theta}(\bm{x}),
\Leq{x2z}
\ee 
where $\V{w}\in \mR^p$ and $\V{h}_{\theta}:\mathcal{X} \to\mathbb{R}^{p}$ are the classifier weight vector and feature extractor, respectively. Here, $\theta$ represents the parameters of the feature extractor.

Under the model \NReqs{z2P}{x2z}, since the probability that $y$ belongs to class $q$ given $z$ is expressed as $P\bigl(y=q \mid z\bigr)= P\bigl(y\le q \mid z \bigr) - P\bigl(y\le q-1 \mid z \bigr)$, the empirical negative log-likelihood given the dataset $\{ D_q \}_{q=1}^{Q}$ becomes
\be
\Leq{nll_real}
\mathcal{L}_{\rm NLL}(\bm{w},\theta,\bm{b})
=\frac{1}{N}\sum_{q=1}^{Q}\sum_{(\bm{x}_i,y_i)\in D_q}
L(z_i,b_{q-1},b_q),
\qquad
     z_i = f_{\bm{w},\theta}(\bm{x}_i),
\ee
where we let $L(z,a,b):=-\log[g(b-z)-g(a-z)]$. 
As in recent practices using DNNs, we consider the $\ell_2$ regularization on the parameters:
\be
\mathcal{R}(\bm{w},\theta)
  =\tfrac{\lambda_{w}}{2}\|\bm{w}\|^{2}_2
  +\tfrac{\lambda_{\theta}}{2}\|\theta\|^{2}_{2}.
\ee
The overall optimization problem is therefore
\be\Leq{optimization_real}
\min_{\bm{w},\theta}
\bigl(
  \mathcal{L}_{\rm NLL}(\bm{w},\theta,\bm{b})
  +\mathcal{R}(\bm{w},\theta)
\bigr).
\Leq{optimization-general}
\ee

\paragraph{UFM for CLM-based OR.}
In the single-layer UFM, the feature vector $\V{h}_{\theta}(\V{x}_i)$ itself is treated as a free learnable variable. As a result, for each datapoint $\V{x}_{i}$, a free variable $\V{h}_i$ is associated. For notational simplicity, we relabel this variable as $\V{h}_{q,i}$, where $q$ indexes the class and $i = 1, \ldots, n_q$ indexes the datapoints within $D_q$. The regularization on the parameter $\theta$ is assumed to be converted to that on $H:=(\V{h}_{q,i})_{q,i}$.
UFM thus allows us to convert \Req{optimization-general} into 
\be
\Leq{optimization-UFM}
\min_{\bm{w},H}
\bigl(
  \mathcal{L}_{\rm NLL,UFM}(\bm{w},H,\bm{b})
  +\mathcal{R}_{\rm UFM}(\bm{w},H)
\bigr),
\ee
where 
\be 
&&
\mathcal{L}_{\rm NLL, UFM}(\bm{w},H,\bm{b})
= \frac{1}{N}\sum_{q=1}^{Q}\sum_{i=1}^{n_q}
L(\bm{w}^\top\bm{h}_{q,i},b_{q-1},b_q),
\\ &&
\mathcal{R}_{\rm UFM}(\bm{w},H)
=
\frac{\lambda_w}{2}\|\bm{w}\|^{2}_{2}+
\frac{\lambda_h}{2N}\sum_{q=1}^{Q}\sum_{i=1}^{n_q}\|\bm{h}_{q,i}\|^{2}_{2}.
\ee

\section{Theoretical results based on UFM analysis}\label{sec:theory}
Let us state our main theoretical results. Thanks to the structure of our CLM and UFM, \Req{optimization-UFM} can be decomposed into a multi-stage optimization as follows:
\be
\min_{w}
\lbb
\frac{\lambda_w}{2}w^2
+
\frac{1}{N}
\sum_{q=1}^{Q}\sum_{i=1}^{n_q}
\min_{\V{h}_{q,i}}
f_q(w,\V{h}_{q,i})
\rbb,
\Leq{optimization_multistage}
\ee
where
\be
&&
f_q(w,\V{h})=L(w\V{a}^{\top}\V{h},b_{q-1},b_{q})+\frac{\lambda_h}{2}\| \V{h} \|_2^2,
\ee
and where $w\ge0$ and $\V{a}$ are the norm of $\V{w}$ and the unit vector representing the direction of $\V{w}$, respectively, so that $\V{w}=w\V{a}$ and $\|\V{a}\|_2=1$ hold. Since our objective function to be minimized in \Req{optimization-UFM} is invariant under any orthogonal transformation $\V{w}\to O\V{w}, \V{h}\to O\V{h}$, $\forall O\in\mathrm{O}(p)$, we can fix the direction $\V{a}$ of $\V{w}$ without loss of generality. Furthermore, we assume that the derivative $g'$ of the inverse link function $g$ is logarithmically concave (log-concave): some standard choices of $g$ in OR such as the logistic function, the standard normal CDF, and the Gumbel CDF satisfy this assumption. 

Under these assumptions, the conditions of ONC can be derived. Before presenting the concrete statements, we first show the following theorem.
\begin{theorem}
  \label{thm:logcon}
  Let $p(x)$ be a log-concave function on $\mathbb{R}$, 
  and let $P(x)=\int_{-\infty}^xp(u)\,du$. 
  Then, for any $a<b$, the function $\rho(x)=P(b-x)-P(a-x)$ is log-concave.
\end{theorem}
\begin{proof}
One can write $\rho(x)$ as
  \begin{equation}
    \rho(x)=P(b-x)-P(a-x)=\int_{a-x}^{b-x}p(u)\,du
    =\int_a^bp(y+x)\,dy,
  \end{equation}
  where we let $y=u-x$. Since $p(y+x)$ is log-concave in $\mathbb{R}^2$, one can apply Theorem~\ref{th:Prekopa} with $A=[a,b]$ to conclude that $\rho(x)$ is log-concave in $x$. 
\end{proof}
This means that the log-concavity of $g'$ leads to the convexity of $L(z,a,b)=-\log[g(b-z)-g(a-z)]$ w.r.t. $z$ for any $(a,b)$ satisfying $b>a$. 
One can further show that, if $g'$ is \emph{strictly} log-concave, 
then $L(z,a,b)$ is \emph{strictly} convex in $z$:
see Appendix~\ref{sec:strictlogconc}. 

We are now ready to state the ONC theorem.
\begin{theorem}[ONC]
\label{thm:ONC}
Assume that the inverse link 
function $g(x)$ defined on $\mathbb{R}$ is differentiable,
and that its derivative $g'$ is log-concave. 
Consider \Req{optimization-UFM} with thresholds $\bm{b} = (b_0, b_1, \ldots, b_Q)$ satisfying $b_0 < b_1 < \cdots < b_Q$, and let $(\bm{w}^*,H^*)$ denote the global minimizer. Under the assumption $\lambda_w,\lambda_h>0$, the following three properties hold: 
\begin{description}
\item[\textbf{(ONC1)}] For any class $q\in\mathcal{Y}$, the optimal features $\{\bm{h}_{q, i}^*\}_i$ in class $q$ become identical: 
\begin{equation*}
\bm{h}_{q, i}^* = \bm{h}_q^*, \quad \forall i=1, \dots, n_q.
\end{equation*}
In other words, the optimal features collapse to their within-class mean $\bm{h}_q^*$.
\item[\textbf{(ONC2)}] \textit{For any class $q$, the class mean $\bm{h}_{q}^*$ becomes parallel to $\bm{w}^*$, meaning that all class means collapse onto the one-dimensional subspace spanned by $\bm{w}^*$. }
\item[\textbf{(ONC3)}] \textit{The optimal latent variables $z^*_q=(\V{w}^*)^{\top}\bm{h}_{q}^*$ satisfy $z_1^*\leq z_2^* \leq \cdots \leq z_Q^*$. Moreover, if $g'$ is strictly log-concave 
and if $\bm{w}^*\not=\mathbf{0}$, then these inequalities hold strictly. 
}
\end{description}
\begin{proof}
By a technical reason, we separately treat the two cases $w^*=0$ and $w^*\neq 0$, and here provide only the derivation of ONC1 and 2,  deferring the proof of ONC3 to Appendix~\ref{sec:Derivation of ONC3}. 
\\
Thanks to the structure of \Req{optimization_multistage}, for any fixed $w$ every $\bm{h}_{q,i}$ can be optimized separately from the other variables, and the objective function is identical for all $i\in\{1,\ldots,n_q\}$. Its explicit form is 
\begin{equation}
    \arg\min_{\bm{h}_{q,i}}f_q(w,\V{h}_{q,i})
    =\arg\min_{\bm{h}}\left(L(w\bm{a}^\top\bm{h},b_{q-1},b_q)
    +\frac{\lambda_h}{2}\|\bm{h}\|_2^2\right).
\end{equation}
First suppose $w\not=0$. Since $L(z,b_{q-1},b_q)$ is proven to be convex in $z$ through Theorem \ref{thm:logcon}, $L(w\bm{a}^\top\bm{h},b_{q-1},b_q)\eqqcolon L_q(w\bm{a}^\top\bm{h})$ is also convex in $\bm{h}$. Since the term $(\lambda_h/2)\| \V{h}\|_2^2$ is strictly convex, the total objective function to be minimized is strictly convex w.r.t. $\V{h}$. 
On the other hand, let $\bm{v}_q$ be the gradient of $L_q(w\bm{a}^\top\bm{h})$ at $\bm{h}=\mathbf{0}$.
Thanks to the convexity of $L_q(w\bm{a}^\top\bm{h})$, one has 
  \begin{equation}
      L_q(w\bm{a}^\top\bm{h})-L_q(0)\ge\bm{v}_q^\top\bm{h}
      ,\quad\forall\bm{h},
  \end{equation}
which implies that the objective function is bounded from below:
  \begin{align}
      L_q(w\bm{a}^\top\bm{h})+\frac{\lambda_h}{2}\|\bm{h}\|_2^2
      \ge L_q(0)+\frac{1}{2}\lambda_h
      \left\| \V{h}+\frac{\bm{v}_q}{\lambda_h} \right\|_2^2
      -\frac{1}{2\lambda_h} \|\bm{v}_q\|_2^2
      \ge L_q(0)-\frac{1}{2\lambda_h}\|\bm{v}_q\|_2^2>-\infty.
  \end{align}
Hence, the strict convexity and the boundedness of $L_q(w\bm{a}^\top\bm{h})+\frac{\lambda_h}{2}\|\bm{h}\|_2^2$ 
  guarantee the uniqueness of the minimizer, proving ONC1. The proof of ONC2 is more straightforward. Let $\V{h}_{\parallel}$ denote the projection of $\V{h}$ on $\V{a}$ and $\V{h}_{\perp}=\V{h}-\V{h}_{\parallel}$. Then we have
\begin{align}
L_q(w\bm{a}^\top\bm{h})+\frac{\lambda_h}{2}\|\bm{h}\|_2^2
&=
L_q(w\bm{a}^\top\bm{h}_{\parallel})
+\frac{\lambda_h}{2}\|\bm{h}_{\parallel}\|_2^2
+\frac{\lambda_h}{2}\|\bm{h}_{\perp}\|_2^2.
\end{align}
Hence, the minimization of this w.r.t. $\bm{h}_{\perp}$ yields $\bm{h}_{\perp}^*=\V{0}$, showing ONC2.
\\
Next we assume $w=0$. In this case, the dependence of the objective function on $\V{h}_{q,i}$ only appears in the regularization term and the optimization thus yields $\V{h}^*_{q,i}=\V{0}$ for all $q,i$. Hence, the ONC properties appear trivially.
\end{proof}
\end{theorem}

In contrast to ONC1 and 2, which only require the convexity of $L(z,a,b)$, ONC3 has a more quantitative information about the problem. Actually, the values of $w^*,\V{z}^*$ are determined from a set of equations deduced from the stationarity condition of \Req{optimization_multistage}. Borrowing the terminology from statistical physics, we call this set of equations {\bf Equations Of State (EOS)}. Analyzing EOS leads to a derivation of ONC3, but it is involved and is deferred to Appendix \NRsec{Derivation of ONC3}. The solution of EOS exhibits some singularity at certain parameter values, and also some simple behaviors in certain limits. The next theorem summarizes these findings. 
\begin{theorem}[EOS, phase transition, and some limiting behaviors]
\label{thm:EOS}
Consider the same situation as in Theorem \ref{thm:ONC}. If the optimal norm value satisfies $w^*>0$, $w^*$ and the optimal latent variables $\V{z}^*$ obey the following set of equations which we call EOS:
\subbe
\be
&&
\frac{g'(b_q-z_q^*)-g'(b_{q-1}-z_q^*)}{g(b_q-z_q^*)-g(b_{q-1}-z_q^*)}+\lambda_h\frac{z^*_q}{(w^*)^2}=0,\quad q=1,\ldots,Q,
\Leq{EOS_z}
\\ 
&&
\lambda_w w^* -\frac{\lambda_h }{(w^*)^3}
\sum_{q=1}^{Q}\alpha_q (z^*_q)^2=0,
\Leq{EOS_w}
\ee
\Leq{EOS}
\subee
where $\alpha_q=n_q/N$. Additionally assuming the continuity and monotonicity of $w^*$ w.r.t.\ $\lambda_h$ and $\lambda_w$, this EOS implies a phase transition with the phase boundary in the $(\lambda_h,\lambda_w)$-plane characterized by
\be
\label{eq:phaseboundary}
\lambda_h\lambda_{w}=C:=\sum_{q=1}^{Q}\alpha_q \lb \frac{g'(b_q)-g'(b_{q-1})}{g(b_q)-g(b_{q-1})} \rb^2.
\ee
Namely, for $\lambda_h\lambda_w \geq C$ the trivial solution $w^*=0,\V{z}^*=\V{0}$ becomes the optimal solution to \Req{optimization_multistage}, while for $\lambda_h\lambda_w < C$ the nontrivial solution $w^*>0,\V{z}^*\neq 0$, which obey EOS, becomes the optimal one.  

Moreover, this EOS admits a simple behavior emerging in the limit where the product $\lambda_h\lambda_w$ approaches zero. In that limit, $\V{z}^*$ is determined by
\be
g'(b_q-z_q^*)=g'(b_{q-1}-z_q^*),
\quad q=1,\ldots,Q,
\Leq{EOS-z^*_zeroreg}
\ee
and one has $w^*=\Theta\bigl((\lambda_h/\lambda_w)^{1/4}\bigr)$. 
\begin{proof}
Applying ONC2, we have $z_q=w\V{a}^{\top}\V{h}_q$.
Thus we may rewrite the squared norm $\| \V{h}_q\|_2^2$ as $z_q^2/w^2$ and the optimization w.r.t.\ $\V{w}$ and $(\V{h}_q)_q$ in \Req{optimization_multistage} can be reduced to those w.r.t.\ $w,\V{z}$. Taking the stationarity condition w.r.t.\ $\V{z}$ and $w$ lead to EOS. 

Next, we examine the phase transition and the phase boundary. One subtlety in analyzing the nature of the phase transition is that the trivial solution $(w^*,\V{z}^*)=(0,\V{0})$ does not satisfy EOS within the whole parameter region where it is optimal. However, thanks to the assumed continuity and monotonicity of $w^*$, exactly on the phase boundary the trivial solution must satisfy EOS. Therefore, we search for a condition under which EOS admits the trivial solution. By substituting \Req{EOS_z} into \Req{EOS_w} to eliminate the explicit dependence on $w^*$, we obtain the following equation:
\be
\lambda_w=\frac{1}{\lambda_h}
\sum_{q=1}^{Q} \alpha_q\left(\frac{g'(b_q-z_q^*)-g'(b_{q-1}-z_q^*)}{g(b_q-z_q^*)-g(b_{q-1}-z_q^*)}\right)^2.
\ee
Inserting the trivial solution $\bm{z}^*=\mathbf{0}$ into this leads to the phase boundary~\eqref{eq:phaseboundary}. Thanks to the assumed monotonicity, once the solution becomes the trivial one, it continues to be so above the boundary. 

Finally, the limiting behavior is investigated. By inserting \Req{EOS_w} into \Req{EOS_z} to eliminate the explicit dependence on $w^*$, we have
\be
\frac{g'(b_q-z_q^*)-g'(b_{q-1}-z_q^*)}{g(b_q-z_q^*)-g(b_{q-1}-z_q^*)}+\sqrt{\lambda_w \lambda_h}\frac{z_q^*}{\sqrt{\sum_{q'=1}^{Q}\alpha_{q'} (z^*_{q'})^2}}=0,\quad q=1,\ldots,Q.
\ee
This yields \Req{EOS-z^*_zeroreg} when $\lambda_h\lambda_w$ is sent to zero, as long as $\V{z}^*\neq \V{0}$. Then, solving \Req{EOS_w} w.r.t.\ $w^*$, we have the scaling $w^*=\Theta\bigl((\lambda_h/\lambda_w)^{1/4}\bigr)$.
\end{proof}
\end{theorem}

The scaling $w^*=\Theta\bigl((\lambda_h/\lambda_w)^{1/4}\bigr)$
means that $w^*$ in the limit $\lambda_h$ and/or $\lambda_w\to0$
may diverge, vanish, or remain finite depending on how one takes the limit. 

\BReq{EOS-z^*_zeroreg} in the vanishing regularization limit is fairly striking since it provides a simple local relation between $\V{b}$ and $\V{z}^*$. Especially, if the inverse link function satisfies a symmetry $1-g(x)=g(-x)$, which is the case for the logit and probit models,
one has $g'(x)=g'(-x)$, and \Req{EOS-z^*_zeroreg} thus implies 
\be
z_q^*=\frac{b_q+b_{q-1}}{2}.
\Leq{ONC3-simple}
\ee
This simple relation will be verified later in experiments using real-world datasets.

\begin{figure}[htbp]
  \centering
  \includegraphics[width=0.44\textwidth]{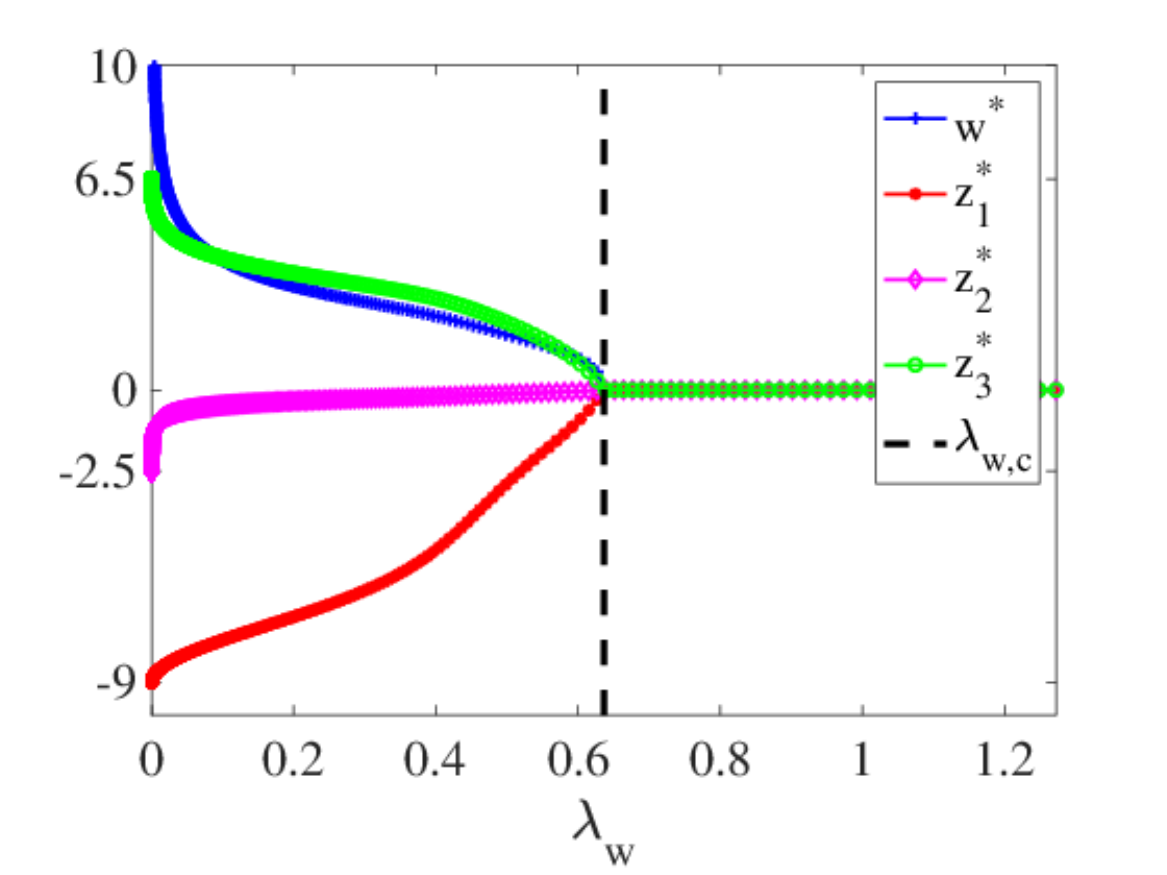}
  \includegraphics[width=0.44\textwidth]{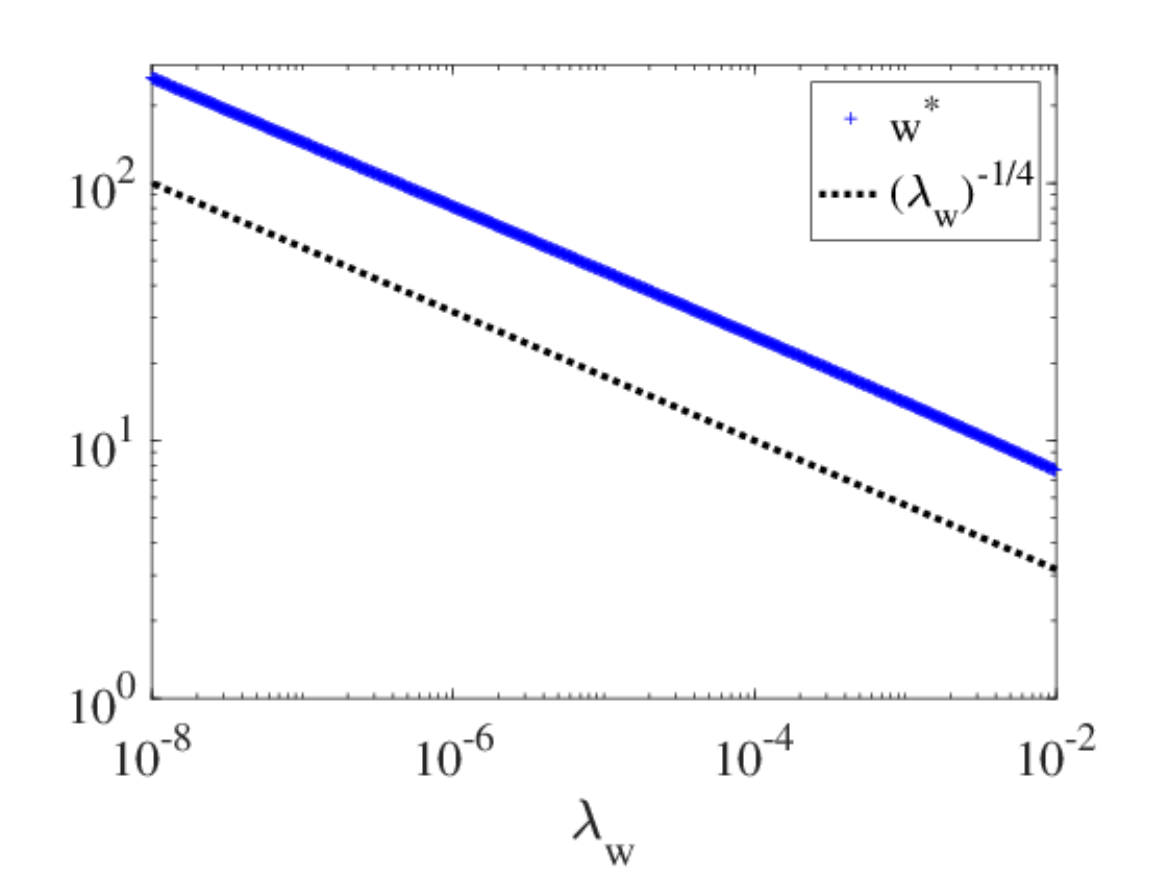}
  \caption{Solution behavior of EOS in the logit model for $Q=3$ with $\V{b}=(-10,-8,3,10)$ at $\lambda_h=1$.
  (Left) $w^*$ and  $\V{z}^*$ are plotted against $\lambda_w$ on a linear scale.  A clear phase transition appears at $\lambda_{w,c}=C/\lambda_h$ (vertical broken line), and the values of $\V{z}^*$ in the limit $\lambda_w \to 0$ match well with the theoretical prediction ($z_q^*=(b_q+b_{q-1})/2$).  (Right) $w^*$ is plotted on a log-log scale in the small-$\lambda_w$ region. A power-law divergence with exponent $-1/4$, corresponding to the scaling $w^*=\Theta\bigl((\lambda_h/\lambda_w)^{1/4}\bigr)$ with fixed $\lambda_h$, is clearly observed. 
  } \label{fig:EOS-behavior}
\end{figure}

For illustration, we numerically solved \Req{EOS} for the logit model and plotted the solution in \Rfig{EOS-behavior}.
The analytical prediction about the critical point and the limiting behaviors were certainly reproduced. 

\section{Experiments}\Lsec{experiments}
\subsection{Experimental setting}

\paragraph{Inverse link functions.}
Two symmetric inverse link functions, the logistic function and the normal CDF, which correspond to the logit and probit models, respectively, were treated in the experiment. 
\paragraph{Datasets and neural networks.}
We used five tabular OR datasets with the largest number of data points---ER, LE, SW, CA, and WR---among those publicly available from \citet{gutierrez2016}. Thirty pre-defined training–validation splits with identical label distributions have been officially released, and we used them as-is in this study. Additionally, we conducted experiments on the UTKFace age estimation dataset \citep{zhang2017age}, which contains face images labeled with ages. We grouped ages into classes with five-year intervals. For the tabular datasets, we employed a multilayer perceptron with residual connections, while for UTKFace, we used ResNet101 and ResNet50 \citep{he2016deep}, and DenseNet201 \citep{huang2017densely} as backbones. The weight decay coefficient was set to small values, with the exact choice varying by condition. The motivation for this setting is that, in the small-regularization limit, a very simple result emerges as shown in \Req{EOS-z^*_zeroreg}, which facilitates experimental verification. Further details are provided in Appendix~\ref{sec:Details of the Experimental Setup}. 
\paragraph{Treatment of thresholds.}
We considered two cases: fixed and learnable thresholds.

For the fixed case, to ensure that the ignored tail probabilities are sufficiently small, the edge thresholds $b_0$ and $b_Q$ were symmetrically fixed ($b_0 = -b_Q$) to sufficiently large values. The remaining thresholds were evenly spaced over the interval $[b_0,b_Q]$. Under this setting, we solved \Req{optimization-general}, where $\theta$ denotes the DNN parameters.

In the learnable case, we set \(b_{0}=-\infty\) and \(b_{Q}=+\infty\) and learned \(b_q\) with \(q=1,\dots,Q-1\). To guarantee the strict ordering between the threshold values, we parameterized them with \(\bm{s}\in\mathbb{R}^{Q-1}\) as
\be
        b_q(\boldsymbol{s})
        :=\sum_{j=1}^{q}\log\bigl(1+e^{s_j}\bigr)
        -\frac{1}{Q-1}\sum_{j=1}^{Q-1}\log\bigl(1+e^{s_j}\bigr),\quad q=1,\dots,Q-1.
\ee
Correspondingly, we solved $\min_{\bm{w},\theta,\V{s}}
\bigl(
  \mathcal{L}_{\rm NLL}(\bm{w},\theta,\bm{b}(\V{s}))
  +\mathcal{R}(\bm{w},\theta)
\bigr)
$
instead of \Req{optimization-general}.

\paragraph{Evaluation metrics.}
We used two basic training metrics for evaluation: $\mathcal{L}_{\rm NLL}$ and the mean absolute error (MAE) for label prediction, 
\(
  \mathrm{MAE}
  = \tfrac{1}{N}\sum_{i=1}^{N}\bigl|\hat y_i - y_i\bigr|
\)
 where \(\hat y_i\) is the predicted label.
Additionally, we evaluated classification accuracy, quadratic weighted kappa (QWK), within-1 accuracy (proportion of predictions within one class of the true label), and minimum sensitivity (worst-case per-class recall).
 Besides, let $\V{h}_{\theta}(\cdot)$ denote the penultimate-layer output of our DNN, and let
$\bar{\bm{h}}_q = (1/n_q)\sum_{(\bm{x}_i,y_i)\in D_q}\bm{h}_\theta(\bm{x}_i)$
and
$\bar{\bm{h}}   = (1/N)\sum_{i=1}^{N}\bm{h}_\theta(\bm{x}_i) $
represent the class-wise and global feature means, respectively. Using these, we introduced the following four quantitative indicators for ONC: 
\be
&&
\underline{\mathrm{ONC}_1} = 
\frac{
(1/Q)\sum_{q=1}^Q \frac{1}{N_q}\sum_{(\bm{x}_i,y_i)\in D_q}\|\bm{h}_\theta(\bm{x}_i)-\bar{\bm{h}}_q\|_2
}{
(1/N)\sum_{i=1}^N\|\bm{h}_\theta(\bm{x}_i)-\bar{\bm{h}}\|_2
},
\\ &&
\underline{\mathrm{ONC}_{2\text{-}1}} = \frac{\sum_{q=1}^Q\|(\bar{\bm{h}}_q-\bar{\bm{h}}) - 
\lb \bm{u}^{\top}(\bar{\bm{h}}_q-\bar{\bm{h}}) \rb \V{u}\|_2^2}{\sum_{q=1}^Q\|\bar{\bm{h}}_q-\bar{\bm{h}}\|_2^2},
\quad
\underline{\mathrm{ONC}_{2\text{-}2}} = 1-\left| \frac{\bm{w}^\top\bm{u}}{\|\bm{w}\|_2 } \right|,
\\ &&
\underline{\mathrm{ONC}_3} = 
\frac{
\sum_{q=1}^{Q-1}|b_q - (z_q+z_{q+1})/2|
}{
\sum_{q=1}^{Q-1}(b_{q+1}-b_q)
}
\Leq{ONC3}
,
\ee
where $\V{u}$ is the unit first principal component of $\{\bar{\bm{h}}_q-\bar{\bm{h}}\}_{q=1}^Q$. $\underline{\mathrm{ONC}_1}$ is the indicator for ONC1 and becomes zero when ONC1 exactly happens; $\underline{\mathrm{ONC}_{2-1}}$ quantifies whether each class mean collapses onto the dominant one-dimensional subspace represented by \( \V{u} \), while $\underline{\mathrm{ONC}_{2-2}}$ measures whether \( \V{w} \) also collapses onto the same subspace; since our experiments focus on the small-regularization regime under the use of symmetric $g$,  ONC3 is expected to emerge in the form \NReq{ONC3-simple}, and accordingly $\underline{\mathrm{ONC}_{3}}$ serves as an appropriate indicator for it.

\subsection{Results}
In the main text, the results for the ER and UTKFace datasets are shown; the logit model is used in both the cases, and ResNet101 among three backbones mentioned above is used for the UTKFace case. Experiments conducted under different settings also yielded consistent results and are reported in Appendix \NRsec{Additional}. The results obtained using the clog-log model, as well as the quantitative comparison between the theoretical and experimental results--where the latter even exhibits phase transitions--are also reported in the same section.

\begin{figure}[!htbp]
  \vspace{-6pt}                 
  \centering
  \includegraphics[width=\textwidth]{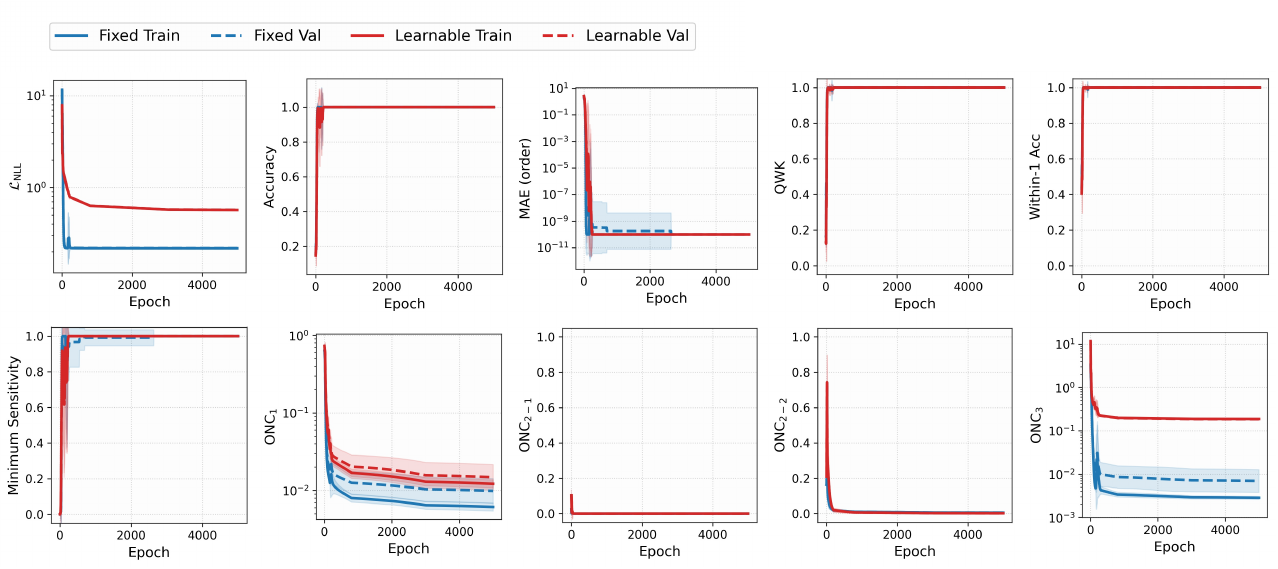}
  \caption{Epoch-wise average metrics curves for the ER dataset with the logit model.}
  \label{fig:metric_curves_logit_ER}
  \vspace{-6pt}                                 
\end{figure}

\begin{figure}[!htbp]
  \vspace{-6pt}                 
  \centering
  \includegraphics[width=\textwidth]{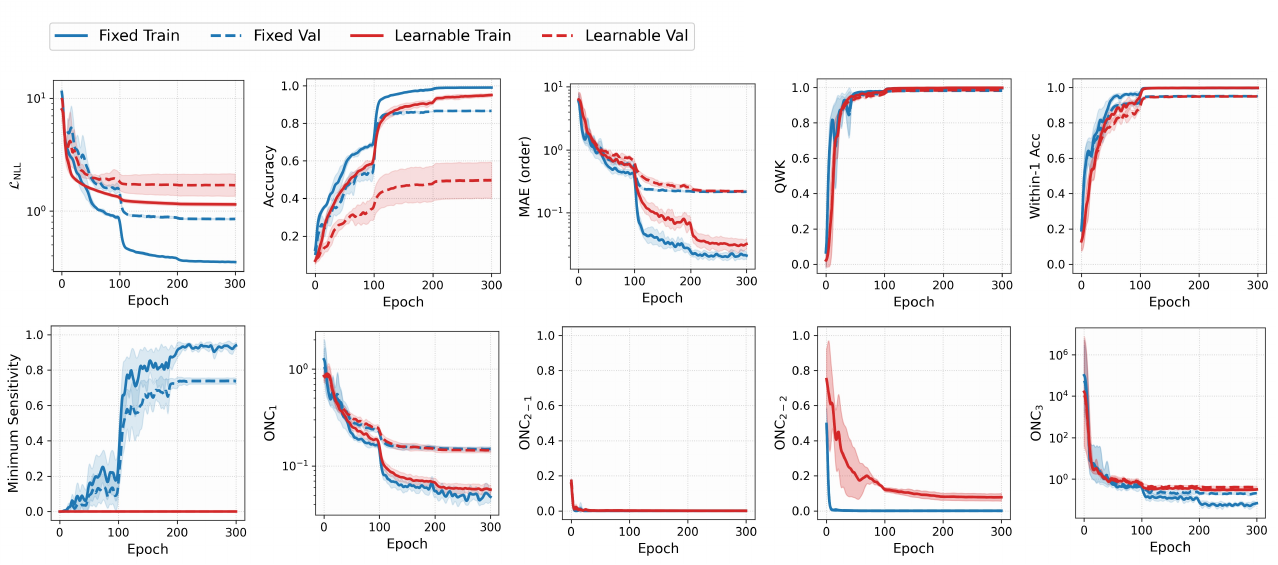}
  \caption{Epoch-wise average metrics curves for the UTKFace dataset with ResNet101 backbone.}
  \label{fig:metric_curves_utkface_resnet101}
  \vspace{-6pt}                 
\end{figure}

Figures~\ref{fig:metric_curves_logit_ER} and \ref{fig:metric_curves_utkface_resnet101} plot the evolution of all evaluation metrics for ER and UTKFace, respectively.

For the ER dataset, both training and validation $\mathrm{MAE}$ approached zero, while accuracy, within-1 accuracy, QWK, and minimum sensitivity all approached one, indicating that all samples were correctly classified. \(\underline{\mathrm{ONC}_{2\text{-}1}}\) and \(\underline{\mathrm{ONC}_{2\text{-}2}}\) rapidly approached zero, showing that the feature vectors collapsed onto the one-dimensional subspace spanned by \(\bm{w}\).  
As training proceeded, \(\underline{\mathrm{ONC}_{1}}\) decreased steadily, confirming that features collapsed toward their class means. For \(\underline{\mathrm{ONC}_{3}}\) we observed a clear difference between the two threshold strategies. With fixed thresholds, $\underline{\mathrm{ONC}_{3}}$ took a small value from an early stage and continued to decrease as training progressed. In contrast, with learnable thresholds, $\underline{\mathrm{ONC}_{3}}$ seemed to converge to a non-zero value. These observations indicate that the simple form of ONC3, given by \Req{ONC3-simple}, practically holds under fixed thresholds but does not hold under learnable ones.

For the more complex UTKFace dataset, the ONC metrics exhibit the same trends as observed on the ER dataset. Across metrics including $\mathcal{L}_{\rm NLL}$, accuracy, MAE, and minimum sensitivity, fixed thresholds demonstrate better performance. Notably, both training and validation minimum sensitivity for learnable thresholds remain at zero throughout training, indicating that at least one class is completely ignored by the model. This phenomenon is consistently observed across the other two backbones presented in Appendix~\ref{sec:utkface}.

\begin{figure}[htbp]
  \vspace{-6pt} 
  \centering
  \includegraphics[width=\textwidth]{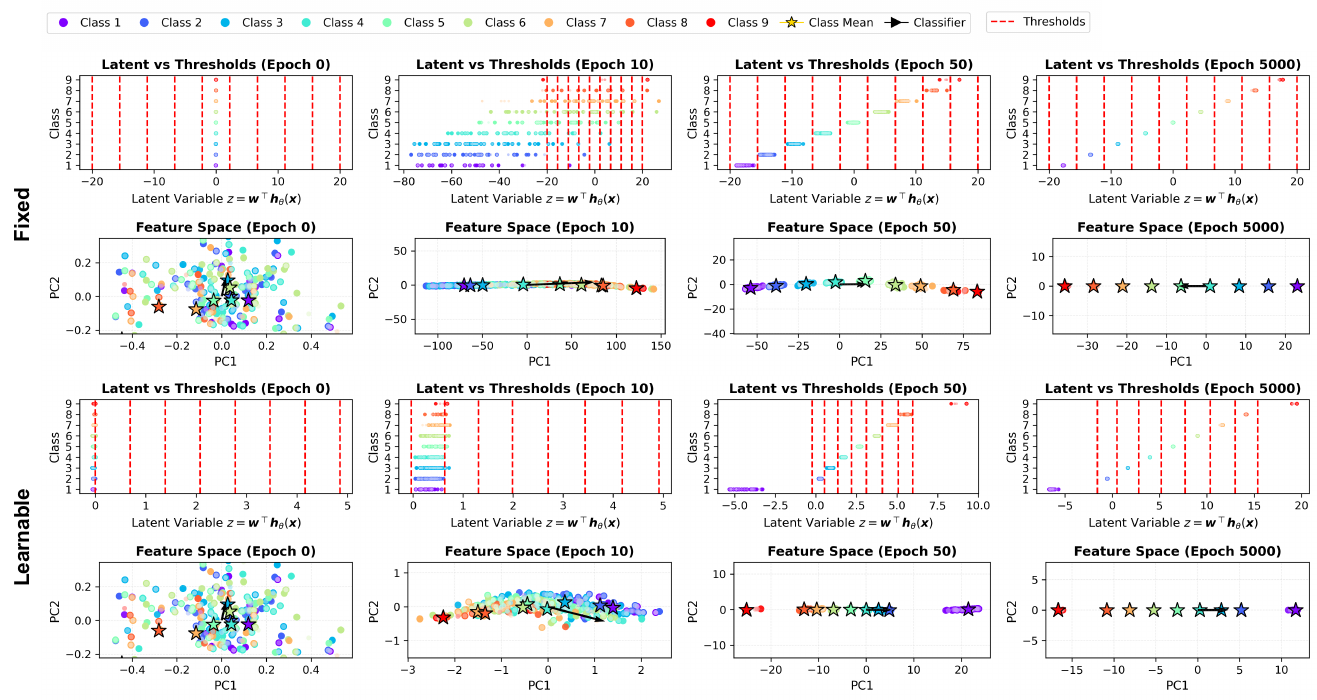}
  \captionsetup{skip=3pt}
  \caption{Latent and feature space visualization for the ER dataset with the logit model.}
  \label{fig:vis_logit_ER}
  \vspace{-6pt} 
\end{figure}

To illustrate the ONC emergence, a visualization of the feature and latent space evolution throughout training are shown in \Rfigs{vis_logit_ER}{vis_utkface_resnet101}. The red dashed lines denote the thresholds. Feature points are two-dimensional PCA-projected feature vectors and are color-coded by class, with validation features in lighter shades. Class means are highlighted with star markers, and the black arrow indicates the classifier weight. This visualization demonstrates the emergence of ONC1--3, but the ONC3 behavior is clearly different between the two threshold strategies, as observed in the latent space.

\begin{figure}[htbp]
  \vspace{-6pt} 
  \centering
  \includegraphics[width=\textwidth]{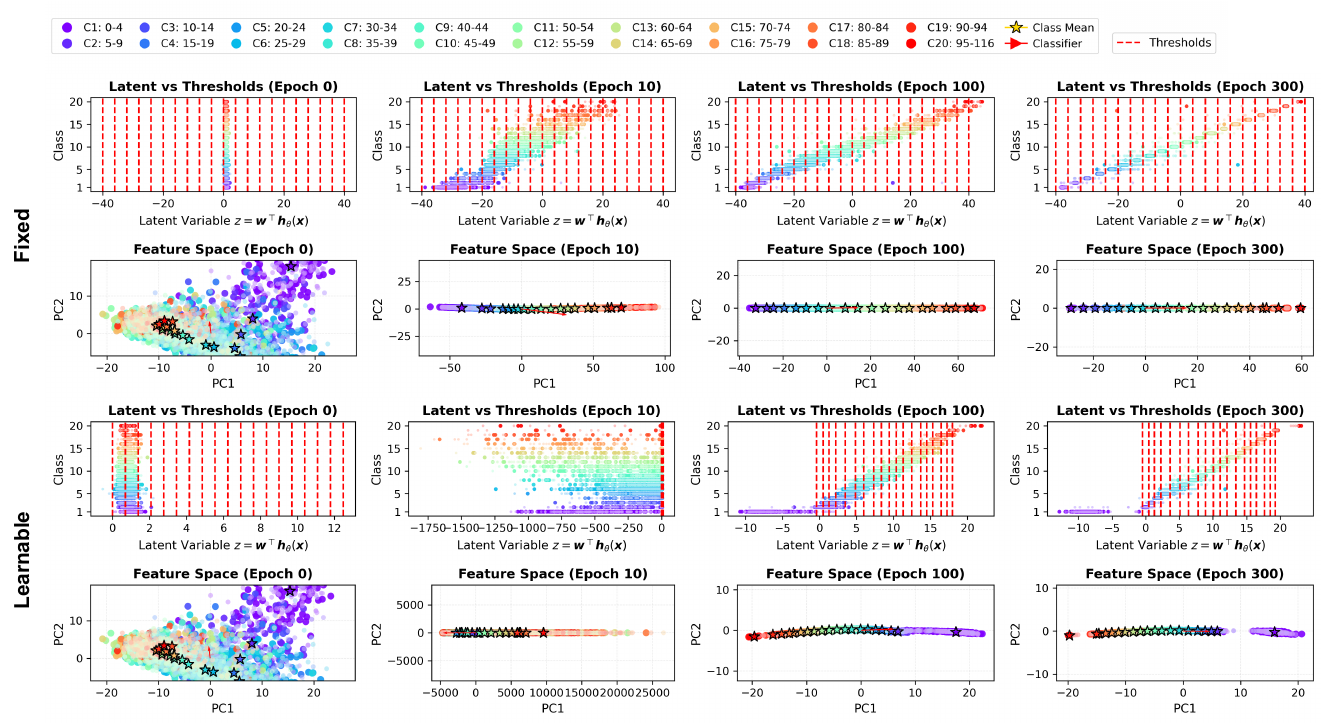}
  \captionsetup{skip=3pt}
  \caption{Latent and feature space visualization for the UTKFace dataset with ResNet101.}
  \label{fig:vis_utkface_resnet101}
  \vspace{-6pt} 
\end{figure}


\section{Discussion}\Lsec{Discussion}
\paragraph{Perspectives and future directions.}
ONC can also offer practical advantages. Although fixed thresholds were introduced primarily to establish ONC, as shown in \Rfig{metric_curves_logit_ER}, they tended to yield faster and more stable convergence compared with learnable thresholds. Moreover, as shown in \Rfig{metric_curves_utkface_resnet101}, for the more complex UTKFace dataset, it can be seen that using fixed thresholds leads to a dramatic improvement in classification accuracy. Judging from the behavior of the minimum sensitivity, this improvement is largely attributed to better classification of the minority classes. This is actually natural since fixed thresholds provide a fairer allocation of the latent space---and hence of predicted probabilities---across all classes and they can offer greater robustness and generalization under label imbalance or label shift. We believe that this insight will be valuable also for practitioners.

Furthermore, the geometric structure induced by ONC can be directly utilized in the design of regularization terms or loss functions. For instance, adding lightweight penalties that attract each class mean toward the classifier axis or to the corresponding threshold midpoint may accelerate training, especially in scenarios with scarce labels or significant class imbalance. We leave such extensions as promising directions for future exploration.

\paragraph{Limitations.}
The theoretical development in \Rsec{theory} assumes that the thresholds $\bm{b}$ are fixed.
Although our experimental results suggest that ONC1--2 also emerges even when $\bm{b}$ is learnable, this has not yet been theoretically established.
Moreover, we believe that there exist only two phases---one with the trivial solution and the other with a non-trivial solution---but we have not succeeded in rigorously proving this.
Instead, in Theorem \ref{thm:EOS}, we circumvented this gap by assuming the continuity and monotonicity of $w^*$ w.r.t. $\lambda_w$ and $\lambda_h$.

During the revision, we realized that the current definition of $\underline{\mathrm{ONC}_3}$ \Req{ONC3} is only valid when the thresholds are fixed and evenly spaced in the latent space, which is the case for our fixed threshold experiments but is not for the others. Even though, our overall conclusion for the learnable threshold case remains unaffected because it is clear from other analyses including the latent space visualization that ONC3 behavior does not occur in the same manner as with fixed thresholds. Developing generally applicable ONC3 metrics is possible but the corresponding experiment is left for future work.

\paragraph{Conclusion.}
This study extended the NC theory to CLM-based OR through the UFM framework, to find ONC involving three hallmark properties. In the small-regularization limit, we also showed that a highly local and simple relation between the thresholds and the latent variables emerges. Experiments on real-world datasets confirmed the theoretical predictions and additionally demonstrated that fixed thresholds not only exhibit ONC but also lead to training speedup and better classification accuracy. These suggest that ONC can be an effective concept for generic OR tasks.

\newpage

\begin{ack}
This work was partially supported by JST, CREST Grant number JPMJCF1862, Japan (TO, TT), JSPS KAKENHI under Grant 22K12179 (TO), Grant-in-Aid for Transformative Research Areas (A), ``Foundation of machine learning physics’' (22H05117) (TO), and Grant-in-Aid for Transformative Research Areas (A), ``Shin-biology regulated by protein lifetime’' (24H01895) (TT).

\end{ack}

\bibliographystyle{plainnat}
\bibliography{references}


\newpage
\appendix

\section{Log-concavity}\Lsec{logconcavity}
\subsection{Pr\'{e}kopa's theorem}
We first recall a theorem, due to \cite{Prekopa1973},
which is derived from the Pr\'{e}kopa-Leindler inequality
(Theorem~\ref{th:plineq})
on log-concave functions. 
It basically states 
that marginalization preserves the log-concavity. 
We use it to prove Theorem~\ref{thm:logcon} in the main text. 




\begin{theorem}[Theorem 6 of~\cite{Prekopa1973}]
  \label{th:Prekopa}
  Let $f(\bm{x},\bm{y})$ be a function of $(n+m)$ variables where $\bm{x}$
  and $\bm{y}$ are $n$- and $m$-dimensional, respectively.
  Suppose that $f$ is log-concave
  on $\mathbb{R}^{n+m}$ and let $A$ be a convex subset of $\mathbb{R}^m$.
  Then the function of $\bm{x}$ defined by
  \begin{equation}
    \int_Af(\bm{x},\bm{y})\,d\bm{y}
  \end{equation}
  is log-concave on $\mathbb{R}^n$. 
\end{theorem}

\subsection{Strict log-concavity}
\label{sec:strictlogconc}
Here we show that the strict log-concavity of $g'(x)$ on $\mathbb{R}$ 
ensures the strict convexity of $L(z,a,b)$ in $z$. 
Although proving it would be easy if one can assume 
differentiability of $g'$,
as demonstrated in Appendix~\ref{sec:Derivation of ONC3},
it holds even without 
the differentiability assumption, as shown in the following.

We start by recalling the Pr\'{e}kopa-Leindler inequality~\citep{Prekopa1973,Leindler1972}. 
\begin{theorem}[Theorem 3 of \protect\cite{Prekopa1973}]
\label{th:plineq}
Suppose that $\lambda_i$, $i=1,\ldots,k$, are positive constants 
satisfying $\sum_{i=1}^k\lambda_i=1$.
Let $f_1,\ldots,f_k$ be nonnegative and Borel measurable functions
on $\mathbb{R}^n$, and let
\begin{equation}
  r(\bm{t})=\sup_{\sum_{i=1}^k\lambda_i\bm{x}_i=\bm{t}}f_1(\bm{x}_1)\cdots f_k(\bm{x}_k)
  ,\quad \bm{t}\in\mathbb{R}^n.
\end{equation}
Then the function $r(\bm{t})$ is Lebesgue measurable
and we have
\begin{equation}
  \int_{\mathbb{R}^n}r(\bm{t})\,d\bm{t}
  \ge\prod_{i=1}^k\left(\int_{\mathbb{R}^n}f_i^{1/\lambda_i}(\bm{x}_i)\,d\bm{x}_i\right)^{\lambda_i}.
\end{equation}
\end{theorem}

We need some more preparation. 
\begin{lemma}
  \label{lm:sp}
  Let $f:D\to[0,\infty)$ be a strictly log-concave function
  on a convex set $D$.
  If there exists $\bm{x}_0\in D$ for which $f(\bm{x}_0)=0$, then
  $\bm{x}_0$ is an extreme point of $D$.
\end{lemma}
\begin{proof}
  We prove that if $\bm{x}_0$ is not an extreme point of $D$
  (that is, $\bm{x}_0$ can be represented as a convex combination
  of $\bm{x},\bm{y}\in D$ such that $\bm{x}_0=\lambda\bm{x}+(1-\lambda)\bm{y}$,
  $\exists\lambda\in(0,1)$), then $f(\bm{x}_0)$ cannot be zero.
  From the strict log-concavity of $f$, one has
  \begin{equation}
    f(\bm{x}_0)=f(\lambda\bm{x}+(1-\lambda)\bm{y})
    >f(\bm{x})^\lambda f(\bm{y})^{1-\lambda}\ge0,
  \end{equation}
  showing that $f(\bm{x}_0)$ is positive.
\end{proof}
It should be noted that
Lemma~\ref{lm:sp} implies that any strictly log-concave function
on $\mathbb{R}^n$ is strictly positive. 

Let $A$ be a subset of $\mathbb{R}^n$.
For $\bm{a}\in\mathbb{R}^n$,
let $A_{\bm{a}}=\{\bm{x}+\bm{a}\mid\bm{x}\in A\}$
and $A_{\backslash\bm{a}}=A\backslash A_{\bm{a}}=A\cap\overline{(A_{\bm{a}})}$.
\begin{lemma}
  Let $A$ be a convex subset of $\mathbb{R}^n$
  and $\bm{a}\in\mathbb{R}^n$ be a non-zero vector.
  Then one has:
  \begin{itemize}
  \item[(a)] If $A_{\backslash\bm{a}}=\emptyset$, then one has
    $\{\bm{x}-t\bm{a}\mid\bm{x}\in A,t\in[0,\infty)\}\subset A$. 
  \item[(b)] Consider the set $A_{\backslash t\bm{a}}$
  indexed by $t\in\mathbb{R}$. 
  If $A_{\backslash t_0\bm{a}}=\emptyset$ for some $t_0>0$,
    then $A_{\backslash t\bm{a}}=\emptyset$ holds for any $t\ge0$.
  \item[(c)] $A_{\backslash t\bm{a}}$ is nondecreasing on $[0,\infty)$
  (that is, for any $0\le t_1<t_2$ one has
  $A_{\backslash t_1\bm{a}}\subset A_{\backslash t_2\bm{a}}$)
  and nonincreasing on $(-\infty,0]$.
  \end{itemize}
\end{lemma}
\begin{proof}
  We first prove (a).
  $A_{\backslash\bm{a}}=\emptyset$ implies $A\subset A_{\bm{a}}$,
  that is, for any $\bm{x}\in A$ one has $\bm{x}-\bm{a}\in A$.
  By induction, for any $\bm{x}\in A$ one has $\bm{x}-m\bm{a}\in A$. 
  for any $m\in\{0,1,\ldots\}$.
  Due to the convexity of $A$, one obtains
  $\bm{x}-t\bm{a}\in A$ for any $t\ge0$, proving (a). 

  We next prove (b).
  From (a), the condition $A_{\backslash t_0\bm{a}}=\emptyset$
  implies that 
  for any $\bm{x}\in A$ one has $\bm{x}\in A_{t\bm{a}}$ for any $t\ge0$, 
  which in turn yields $A_{\backslash t\bm{a}}=\emptyset$ for any $t\ge0$.
  
  We now prove (c). 
  Fix $0\le t_1<t_2$.
  Take any point $\bm{x}\in A_{\backslash t_1\bm{a}}$, which implies that
  $\bm{x}\in A$ and $\bm{x}-t_1\bm{a}\not\in A$.
  We show that $\bm{x}-t_2\bm{a}\not\in A$. 
  For this purpose, assume, to the contrary,
  that $\bm{x}-t_2\bm{a}\in A$ holds.
  One has
  \begin{equation}
    \bm{x}-t_1\bm{a}=\left(1-\frac{t_1}{t_2}\right)(\bm{x}-t_2\bm{a})
    +\frac{t_1}{t_2}\bm{x},
  \end{equation}
  which shows that $\bm{x}-t_1\bm{a}$ is a convex combination
  of $\bm{x}$ and $\bm{x}-t_2\bm{a}$, both lying in $A$.
  Due to the convexity of $A$, it would follow that
  $\bm{x}-t_1\bm{a}\in A$, which is a contradiction.
  We have thus proved that for any $\bm{x}\in A_{\backslash t_1\bm{a}}$,
  one has $\bm{x}-t_2\bm{a}\not\in A$,
  and hence $\bm{x}\in A_{\backslash t_2\bm{a}}$.
  It in turn implies $A_{\backslash t_1\bm{a}}\subset A_{\backslash t_2\bm{a}}$,
  proving that $A_{\backslash t\bm{a}}$ is nondecreasing on $[0,\infty)$.
  The statement that $A_{\backslash t\bm{a}}$ is nonincreasing on $(-\infty,0]$ can be proved 
  in the same manner. 
\end{proof}

For a subset $A$ of $\mathbb{R}^n$, let $\mu(A)$ denote its volume.
\begin{lemma}
  \label{lm:0plaw}
  Let $A$ be a convex subset of $\mathbb{R}^n$,
  and $\bm{a}\in\mathbb{R}^n$ be a non-zero vector.
  \begin{itemize}
  \item[(a)] If $\mu(A_{\backslash t_0\bm{a}})=0$ for some $t_0>0$,
  then one has $\mu(A_{\backslash t\bm{a}})=0$ for all $t\ge0$.
  \item[(b)] If $\mu(A_{\backslash t_0\bm{a}})>0$ for some $t_0>0$,
  then one has $\mu(A_{\backslash t\bm{a}})>0$ for all $t>0$.
  \end{itemize}
\end{lemma}
\begin{proof}
  We first prove (a). 
  The condition $\mu(A_{\backslash t_0\bm{a}})=0$ implies that
  for almost every $\bm{x}\in A$ one should have $\bm{x}\in A_{t_0\bm{a}}$,
  or eqivalently, $\bm{x}-t_0\bm{a}\in A$.
  One then has $\bm{x}-2t_0\bm{a}\in A$ for almost every such
  $\bm{x}\in A$ satisfying $\bm{x}-t_0\bm{a}\in A$.
  By induction, for almost every $\bm{x}\in A$
  one has $\bm{x}-mt_0\bm{a}\in A$ for any $m\in\{1,2,\ldots\}$.
  Due to the convexity of $A$, for almost every $\bm{x}\in A$
  one has $\{\bm{x}-t\bm{a}\mid t\ge0\}\subset A$.
  It then implies that, for any $t\ge0$
  and for almost every $\bm{x}\in A$,
  one has $\bm{x}\in A_{t\bm{a}}$, showing that
  $\mu(A_{\backslash t\bm{a}})=0$ holds for all $t\ge0$.

  Proving (b) is straightforward by noting that
  if one has $\mu(A_{\backslash t_0\bm{a}})>0$ and $\mu(A_{\backslash t_1\bm{a}})=0$
  for $t_0,t_1>0$, $t_0\not=t_1$, the latter would imply, via (a), 
  $\mu(A_{\backslash t\bm{a}})=0$ for all $t\ge0$,
  which is in conflict with the former. 
\end{proof}
We need the following assumption.
\begin{assumption}
  \label{asm:assumption_A}
  The subset $A$ of $\mathbb{R}^n$ is convex. 
  Furthermore, for any non-zero vector $\bm{a}\in\mathbb{R}^n$,
  either $\mu(A_{\backslash\bm{a}})$ or $\mu(A_{\backslash-\bm{a}})$
  is positive. 
\end{assumption}
The last positivity assumption in Assumption~\ref{asm:assumption_A}
does not always hold (consider the case where $A$ is a half-space
and $\bm{a}$ is parallel to the boundary of $A$).
Lemma~\ref{lm:sufcond} below gives a necessary and sufficient condition
for that. 
\begin{lemma}
  \label{lm:sufcond}
  Let $A$ be a convex subset of $\mathbb{R}^n$ with $\mu(A)>0$.
  Assume that $A$ has $n$ supporting hyperplanes $H_1,\ldots,H_n$ with
  their normal vectors $\bm{n}_1,\ldots,\bm{n}_n$
  forming a basis of $\mathbb{R}^n$.
  Then for any non-zero vector $\bm{a}\in\mathbb{R}^n$,
  either $\mu(A_{\backslash\bm{a}})$ or  $\mu(A_{\backslash-\bm{a}})$ is positive.
  Conversely, if the normal vectors of the supporting hyperplanes of $A$
  do not span $\mathbb{R}^n$, then there exists $\bm{a}\not=\mathbf{0}$
  with which $\mu(A_{\backslash\bm{a}})=\mu(A_{\backslash-\bm{a}})=0$ holds. 
\end{lemma}
\begin{proof}
  We first prove the former statement. 
  As $\{\bm{n}_1,\ldots\bm{n}_n\}$ is a basis of $\mathbb{R}^n$,
  not all the inner products $\{\langle\bm{n}_i,\bm{a}\rangle\}_{i=1,\ldots,n}$
  are simultaneously zero for any non-zero vector $\bm{a}$. 
  Assume without loss of generality that $\langle\bm{n}_1,\bm{a}\rangle\not=0$.
  Let $\bm{p}\in H_1$ be a boundary point of $A$.
  Then one has $H_1=\{\bm{x}\in\mathbb{R}^n\mid
  \langle\bm{n}_1,\bm{x}-\bm{p}\rangle=0\}$.
  Take $-\bm{n}_1$ in place of $\bm{n}_1$ if necessary,
  one can assume that 
  $\{\bm{x}\in\mathbb{R}^n\mid\langle\bm{n}_1,\bm{x}-\bm{p}\rangle\ge0\}
  \supset A$ holds. 

  Take a point $\bm{z}$ in the interior of $A$,
  and a ball $B_r(\bm{z})$ of small enough radius $r>0$ centered at $\bm{z}$,
  so that $B_r(\bm{z})\subset A$ holds.
  Consider the hyperplane $H=\{\bm{x}\in\mathbb{R}^n\mid
  \langle\bm{n}_1,\bm{x}-\bm{z}\rangle=0\}$. 
  It is parallel to $H_1$ and passes through $\bm{z}$.
  It cuts the set $A$ into two parts, each of which contains
  the half of the ball $B_r(\bm{x})$ and is thus of positive volume.
  Let $A'=\{\bm{x}\in A\mid\langle\bm{n}_1,\bm{x}-\bm{z}\rangle<0\}$,
  that is, $A'$ is the set of points in $A$ which resides on the same
  side of $H$ as the point $\bm{p}$.
  One consequently has $\mu(A')>0$ since $A'$ contains the half-ball.

  Let
  \begin{equation}
    \bm{p}'=\bm{p}+t\bm{a},\quad
    t=\frac{\langle\bm{n}_1,\bm{z}-\bm{p}\rangle}{\langle\bm{n}_1,\bm{a}\rangle}.
  \end{equation}
  The point $\bm{p}'$ is on $H$ because
  \begin{equation}
    \label{eq:onH}
    \langle\bm{n}_1,\bm{p}'-\bm{z}\rangle
    =\langle\bm{n}_1,\bm{p}-\bm{z}\rangle
    +\frac{\langle\bm{n}_1,\bm{z}-\bm{p}\rangle}{\langle\bm{n}_1,\bm{a}\rangle}\langle\bm{n}_1,\bm{a}\rangle
    =0
  \end{equation}
  holds.
  Furthermore, $H$ is a supporting hyperplane of $A_{t\bm{a}}$:
  noting that $\bm{p}'+\bm{p}+t\bm{a}\in A_{t\bm{a}}=\{\bm{x}+t\bm{a}\mid\bm{x}\in A\}$, 
  one has 
  \begin{align}
    \langle\bm{n}_1,\bm{x}+t\bm{a}-\bm{z}\rangle
    &=\langle\bm{n}_1,\bm{x}-\bm{p}\rangle+\langle\bm{n}_1,\bm{p}+t\bm{a}-\bm{z}\rangle
    \nonumber\\
    &=\langle\bm{n}_1,\bm{x}-\bm{p}\rangle+\langle\bm{n}_1,\bm{p}'-\bm{z}\rangle
    \ge0,
  \end{align}
  where the last inequality is due to
  $A\subset\{\bm{x}\mid\langle\bm{n}_1,\bm{x}-\bm{p}\rangle\ge0\}$
  and~\eqref{eq:onH}. 
  It implies that $A_{t\bm{a}}\subset\{\bm{x}\mid\langle\bm{n}_1,\bm{x}-\bm{z}\rangle\ge0\}$ holds.
  It should be noted that the point $\bm{p}$ lies on the opposite side
  of $H$ to $A_{t\bm{a}}$,
  which can be confirmed by noting that
  $\bm{z}\in A\subset\{\bm{x}\mid\langle\bm{n}_1,\bm{x}-\bm{p}\rangle\ge0\}$
  implies $\langle\bm{n}_1,\bm{p}-\bm{z}\rangle<0$. 
  One consequently has $A_{t\bm{a}}\cap A'=\emptyset$,
  which furthermore implies $A'\subset\overline{(A_{t\bm{a}})}$. 

  One therefore has 
  $A_{\backslash t\bm{a}}=A\cap\overline{(A_{t\bm{a}})}\supset A'$,
  and thus $\mu(A_{\backslash t\bm{a}})\ge\mu(A')>0$. 
  One can then apply Lemma~\ref{lm:0plaw} to show that
  either $\mu(A_{\backslash\bm{a}})$ or $\mu(A_{\backslash-\bm{a}})$ is positive.

  We next prove the converse.
  Let $W$ be the linear span of the normals
  of the supporting hyperplanes of $A$,
  and let $\pi_W:\mathbb{R}^n\to W$ be the orthogonal projection onto $W$.
  The supporting hyperspace of $A$
  with the normal $\bm{n}\in W$ can be represented as
  $c_{\bm{n}}\ge\langle\bm{n},\bm{x}\rangle=\langle\bm{n},\pi_W(\bm{x})\rangle$.
  Since the closure $A^\circ$ of $A$ is convex and thus is the intersection
  of its supporting hyperspaces~\cite[Theorem 4.5]{Brondsted1983},
  one has
  \begin{align}
    A^\circ
    &=\bigcap_{\bm{n}\in W}\{\bm{x}\mid\langle\bm{n},\bm{x}\rangle\le c_{\bm{n}}\}
    \nonumber\\
    &=\bigcap_{\bm{n}\in W}\{\bm{x}\mid\langle\bm{n},\pi_W(\bm{x})\rangle\le c_{\bm{n}}\}
    \nonumber\\
    &=\bigcap_{\bm{n}\in W}
    \bigl(\{\bm{y}\in W\mid\langle\bm{n},\bm{y}\rangle\le c_{\bm{n}}\}
    +W^\perp\bigr)
    \nonumber\\
    &=\left(\bigcap_{\bm{n}\in W}
    \{\bm{y}\in W\mid\langle\bm{n},\bm{y}\rangle\le c_{\bm{n}}\}\right)
    +W^\perp,
  \end{align}
  showing that $A^\circ$ is a cylinder
  since $W^\perp\not=\{\mathbf{0}\}$ . 
  It immediately implies that for any $\bm{a}\in W^\perp$
  one has $(A^\circ)_{\backslash\bm{a}}=\emptyset$,
  and consequently, $\mu(A_{\backslash\bm{a}})=0$. 
\end{proof}
The positivity assumption in Assumption~\ref{asm:assumption_A} is
automatically satisfied if $A$ is convex and $\mu(A)\in(0,\infty)$,
as shown in the next lemma. 
\begin{lemma}
  \label{lm:pv}
  Let $A$ be a convex subset of $\mathbb{R}^n$,
  and $\bm{a}\in\mathbb{R}^n$ be a non-zero vector. 
  Assume $\mu(A)\in(0,\infty)$. 
  Then one has $\mu(A_{\backslash\bm{a}})>0$. 
\end{lemma}
\begin{proof}
  The condition $\mu(A)\in(0,\infty)$ implies that $A$ is bounded
  and that its closure $A^\circ$ is compact.
  For any nonzero $\bm{n}\in\mathbb{R}^n$,
  the function $\bm{x}\mapsto\langle\bm{n},\bm{x}\rangle$
  is continuous, and it attains its maximum $c_{\bm{n}}$ on $A^\circ$.
  Then $\langle\bm{n},\bm{x}\rangle=c_{\bm{n}}$ is
  the supporting hyperplane of $A$ with normal $\bm{n}$.
  One can then observe that, 
  by taking $\bm{n}_1=\bm{a}=\bm{n}$ in the proof of Lemma~\ref{lm:sufcond}, 
  one has $\mu(A_{\backslash\bm{n}})>0$. 
\end{proof}

We now state the main theorem.
\begin{theorem}
\label{th:slcn}
Assume that $f(\bm{x})$ is strictly log-concave on $\mathbb{R}^n$.
For a subset $A$ of $\mathbb{R}^n$ satisfying Assumption~\ref{asm:assumption_A},
let $\rho(\bm{x})$ be defined by
\begin{equation}
  \rho(\bm{x})=\int_{A}f(\bm{u}-\bm{x})\,d\bm{u}.
\end{equation}
Then $\rho(\bm{x})$ is also strictly log-concave.
\end{theorem}
\begin{proof}
Take arbitrary $\bm{x}_0,\bm{x}_1\in\mathbb{R}^n$
with $\bm{x}_1-\bm{x}_0=\bm{\delta}\not=\mathbf{0}$ and 
let $\bm{x}_\lambda=(1-\lambda)\bm{x}_0+\lambda\bm{x}_1$.
We will show the strict inequality
$\rho(\bm{x}_\lambda)>\rho(\bm{x}_0)^{1-\lambda}\rho(\bm{x}_1)^\lambda$ to hold
for any $\lambda\in(0,1)$, which proves the theorem. 
Let
\begin{equation}
  \label{eq:defFln}
  F_\lambda(\bm{u})=f(\bm{u}-\bm{x}_\lambda)\mathbbm{1}(\bm{u}\in A).
\end{equation}
One then has
\begin{equation}
  \rho(\bm{x}_\lambda)=\int_Af(\bm{u}-\bm{x}_\lambda)\,d\bm{u}
  =\int_{\mathbb{R}^n}F_\lambda(\bm{u})\,d\bm{u}.
\end{equation}
$F_\lambda(\bm{u})$ is log-concave in $\bm{u}$
because it is a product of the two log-concave functions
$f(\bm{u}-\bm{x}_\lambda)$ and $\mathbbm{1}(\bm{u}\in A)$
(note that $A$ is convex due to Assumption~\ref{asm:assumption_A}),
and because multiplication of log-concave functions preserves log-concavity.
\begin{lemma}
  \label{lm:sineqn}
  For any $\lambda\in(0,1)$ and any $\bm{u}_0,\bm{u}_1\in\mathbb{R}^n$,
  let $\bm{u}_\lambda=(1-\lambda)\bm{u}_0+\lambda\bm{u}_1$. 
  One then has
  \begin{equation}
    \label{eq:lcFn}
    F_\lambda(\bm{u}_\lambda)\ge F_0(\bm{u}_0)^{1-\lambda}F_1(\bm{u}_1)^\lambda,
  \end{equation}
  with strict inequality when $F_\lambda(\bm{u}_\lambda)>0$
  and $\bm{u}_1-\bm{u}_0\not=\bm{\delta}$. 
\end{lemma}
\begin{proof}[Proof of Lemma~\ref{lm:sineqn}]
  Fix $\lambda\in(0,1)$.
  Assume that $\bm{u}_0,\bm{u}_1$ are such that $\bm{u}_\lambda\in A$
  and $\bm{u}_1-\bm{u}_0\not=\bm{\delta}$ hold.
  One then has $F_\lambda(\bm{u}_\lambda)=f(\bm{u}_\lambda-\bm{x}_\lambda)>0$,
  where the positivity of $f$ is due to Lemma~\ref{lm:sp}. 
  For such $\bm{u}_0,\bm{u}_1$, one has 
\begin{align}
  \label{eq:slc}
  F_\lambda(\bm{u}_\lambda)
  &=f(\bm{u}_\lambda-\bm{x}_\lambda)
  =f\bigl((1-\lambda)(\bm{u}_0-\bm{x}_0)+\lambda(\bm{u}_1-\bm{x}_1)\bigr)
  \nonumber\\
  &>f(\bm{u}_0-\bm{x}_0)^{1-\lambda}f(\bm{u}_1-\bm{x}_1)^\lambda
  \nonumber\\
  &\ge F_0(\bm{u}_0)^{1-\lambda}F_1(\bm{u}_1)^\lambda,
\end{align}
where the first inequality is due to the strict log-concavity of $f$
and the condition $\bm{u}_0-\bm{x}_0\not=\bm{u}_1-\bm{x}_1$
which is derived from $\bm{u}_1-\bm{u}_0\not=\bm{\delta}=\bm{x}_1-\bm{x}_0$,
and where the second inequality results from multiplication
with the indicator functions.

For $\bm{u}_0,\bm{u}_1$ such that $\bm{u}_\lambda\in A$
and $\bm{u}_1-\bm{u}_0=\bm{\delta}$ hold,
one has $\bm{u}_1-\bm{x}_1=\bm{u}_0-\bm{x}_0$ and
$\bm{u}_\lambda-\bm{x}_\lambda=(1-\lambda)(\bm{u}_1-\bm{x}_1)+\lambda(\bm{u}_0-\bm{x}_0)=\bm{u}_0-\bm{x}_0$, 
so that $f(\bm{u}_\lambda-\bm{x}_\lambda)$ is constant for $\lambda\in[0,1]$,
and hence the inequality~\eqref{eq:lcFn} holds. 
For $\bm{u}_0,\bm{u}_1$ such that $\bm{u}_\lambda\not\in A$ holds,
either $\bm{u}_0$ or $\bm{u}_1$ should lie outside $A$,
so that one has the equality 
$F_\lambda(\bm{u}_\lambda)=F_0(\bm{u}_0)^{1-\lambda}F_1(\bm{u}_1)^\lambda=0$.
\end{proof}

We return to the proof of Theorem~\ref{th:slcn}. 
Let
\begin{equation}
  r(\bm{u})=\sup_{\bm{u}_0,\bm{u}_1:(1-\lambda)\bm{u}_0+\lambda\bm{u}_1=\bm{u}}
  F_0(\bm{u}_0)^{1-\lambda}F_1(\bm{u}_1)^\lambda.
\end{equation}
Then for $\bm{u}\not\in A$ one has $F_\lambda(\bm{u})=r(\bm{u})=0$.
On the other hand,
for $\bm{u}\in A_{\backslash\lambda\bm{\delta}}$, one cannot have $\bm{u}_0,\bm{u}_1\in A$
which satisfy both $\bm{u}_1-\bm{u}_0=\bm{\delta}$ and $(1-\lambda)\bm{u}_0+\lambda\bm{u}_1=\bm{u}$:
indeed, since $\bm{u}_0\in A$, if one lets $\bm{u}_1=\bm{u}_0+\bm{\delta}$,
one has $\bm{u}=(1-\lambda)\bm{u}_0+\lambda\bm{u}_1=\bm{u}_0+\lambda\bm{\delta}\in A_{\lambda\bm{\delta}}$, 
which should not lie in $A_{\backslash\lambda\bm{\delta}}$. 
This, together with Lemma~\ref{lm:sineqn},
in turn implies the strict inequality $F_\lambda(\bm{u})>r(\bm{u})$
for $\bm{u}\in A_{\backslash\lambda\bm{\delta}}$. 
One can similarly show the strict inequality $F_\lambda(\bm{u})>r(\bm{u})$
to hold for $\bm{u}\in A_{\backslash-(1-\lambda)\bm{\delta}}$ as well.

As we have shown the strict inequality $F_\lambda(\bm{u})>r(\bm{u})$
to hold for $\bm{u}$ in the set $A_{\backslash\lambda\bm{\delta}}\cup A_{\backslash-(1-\lambda)\bm{\delta}}$ which has a positive volume
for $\bm{\delta}\not=\mathbf{0}$ due to Assumption~\ref{asm:assumption_A},
one has 
\begin{align}
  \rho(\bm{x}_\lambda)
  &=\int_{\mathbb{R}^n}F_\lambda(\bm{u})\,d\bm{u}
  \nonumber\\
  &>\int_{\mathbb{R}^n}r(\bm{u})\,d\bm{u}
  \nonumber\\
  &\ge\left(\int_{\mathbb{R}^n}F_0(\bm{u})\,d\bm{u}\right)^{1-\lambda}
  \left(\int_{\mathbb{R}^n}F_1(\bm{u})\,d\bm{u}\right)^\lambda
  \nonumber\\
  &=\rho(\bm{x}_0)^{1-\lambda}\rho(\bm{x}_1)^\lambda,
\end{align}
where the first inequality is due to the strict inequality
shown above, and where the second inequality is derived
by applying the Pr\'{e}kopa-Leindler inequality (Theorem~\ref{th:plineq}).
This proves the strict log-concavity of $\rho$. 
\end{proof}

Strict log-concavity of the function 
\begin{equation}
  g(b-z)-g(a-z)=\int_a^bg'(u-z)\,du
\end{equation}
in $z$ under the assumption of strict log-concavity of $g'$
is immediate from Theorem~\ref{th:slcn}.
It proves the strict convexity of $L(z,a,b)=-\log[g(b-z)-g(a-z)]$
in $z$ as well. 

It should be noted that the function $f(\bm{u}-\bm{x})\mathbbm{1}(\bm{u}\in A)$,
whose integral with respect to $\bm{u}\in\mathbb{R}^n$ yields $\rho(\bm{x})$, 
is itself log-concave but not strictly log-concave in $(\bm{x},\bm{u})$,
so that one cannot apply 
the argument in~\cite{Prekopa1973} on 
the strict log-concavity to our case.

\section{Derivation of ONC3}\Lsec{Derivation of ONC3}
We show in this section the ordering 
$z_1^*\le z_2^*\le\cdots\le z_Q^*$ of the optimal latent variables. 
For this purpose, 
we discuss how the minimizer of the optimization problem
\begin{equation}
  \label{eq:optL}
  \min_x\left(L(x,a,b)+\frac{\lambda}{2}x^2\right),\quad\lambda>0,
  \ a<b,
\end{equation}
with the function $L(x,a,b)$ being of the form 
\begin{equation}
  \label{eq:defL}
  L(x,a,b)=-\log[P(b-x)-P(a-x)],
\end{equation}
behaves as one changes $a,b$,
where $P$ is an indefinite integral of a function $p$,
which is log-concave on $\mathbb{R}$. 
It is because the optimal latent variable $z_q^*=w\bm{a}^\top\bm{h}_q^*$ given $w>0$
is determined as the optimal solution of the following minimization: 
\begin{equation}
    \min_{x}\left(L(x,b_{q-1},b_q)+\frac{\lambda_h}{2w^2}x^2\right),
\end{equation}
with $L(x,a,b)$ defined as in~\eqref{eq:defL} using $P(z)=\int_{-\infty}^zg(u)\,du$. 

Theorem~\ref{thm:logcon} ensures that $L(x,a,b)$ is convex in $x$ for any $a,b$, 
which in turn ensures that the minimizer $\hat{x}$ of~\eqref{eq:optL} is unique.
We show in the following that the minimizer $\hat{x}=\hat{x}(a,b)$
is monotonically non-decreasing in $a$ and $b$.
This monotonicity will prove the desired ordering of $\{z_q^*\}_q$. 

The minimizer $\hat{x}$ satisfies the stationarity condition
\begin{equation}
  \label{eq:optimalitycond}
  L_x(\hat{x},a,b)+\lambda\hat{x}=0,
\end{equation}
where the subscript $x$ of $L$ denotes the partial derivative
of $L$ with respect to $x$. 
It should be noted that particularizing \eqref{eq:optimalitycond} 
in the optimization of the latent variable $z_q$ 
yields EOS~\eqref{eq:EOS_z}. 

We first assume differentiability of $p$. 
Taking the derivative of both sides of~\eqref{eq:optimalitycond} 
w.r.t.\ $a$, one has
\begin{equation}
  L_{xx}(\hat{x},a,b)\hat{x}_a+L_{xa}(\hat{x},a,b)+\lambda\hat{x}_a=0,
\end{equation}
yielding
\begin{equation}
  \label{eq:hatxa}
  \hat{x}_a=-\frac{L_{xa}(\hat{x},a,b)}{L_{xx}(\hat{x},a,b)+\lambda}.
\end{equation}
Similarly, one has 
\begin{equation}
  \label{eq:hatxb}
  \hat{x}_b=-\frac{L_{xb}(\hat{x},a,b)}{L_{xx}(\hat{x},a,b)+\lambda}.
\end{equation}
Since $L(x,a,b)$ is convex in $x$,
one has $L_{xx}(x,a,b)\ge0$.
One also has 
\begin{align}
  L_x(x,a,b)
  &=\frac{p(b-x)-p(a-x)}{P(b-x)-P(a-x)},
  \nonumber\\
  \label{eq:Lxa}
  L_{xa}(x,a,b)
  &=-\frac{p'(a-x)}{P(b-x)-P(a-x)}
  +\frac{[p(b-x)-p(a-x)]p(a-x)}{[P(b-x)-P(a-x)]^2}
  \nonumber\\
  &=-\frac{p(a-x)}{P(b-x)-P(a-x)}
  \left[\frac{p'(a-x)}{p(a-x)}-\frac{p(b-x)-p(a-x)}{P(b-x)-P(a-x)}
    \right],
  \\
  \label{eq:Lxb}
  L_{xb}(x,a,b)
  &=\frac{p'(b-x)}{P(b-x)-P(a-x)}
  -\frac{[p(b-x)-p(a-x)]p(b-x)}{[P(b-x)-P(a-x)]^2}
  \nonumber\\
  &=\frac{p(b-x)}{P(b-x)-P(a-x)}
  \left[\frac{p'(b-x)}{p(b-x)}-\frac{p(b-x)-p(a-x)}{P(b-x)-P(a-x)}\right],
  \\
  \label{eq:Lxx}
  L_{xx}(x,a,b)
  &=-\frac{p'(b-x)-p'(a-x)}{P(b-x)-P(a-x)}
  +\frac{[p(b-x)-p(a-x)]^2}{[P(b-x)-P(a-x)]^2}
  \nonumber\\
  &=-[L_{xa}(x,a,b)+L_{xb}(x,a,b)].
\end{align}
Since $p(u)$ is assumed log-concave, $(\log p(u))'=p'(u)/p(u)$ is
monotonically non-increasing. 
One therefore has, via the technique used in~\cite[Proof of Lemma 1]{Dierker1991}
and \cite{BagnoliBergstrom2005}, 
\begin{align}
  \label{eq:Lxapos}
  \frac{p'(a-x)}{p(a-x)}[P(b-x)-P(a-x)]
  &=\frac{p'(a-x)}{p(a-x)}\int_{a-x}^{b-x}p(u)\,du
  \nonumber\\
  &\ge\int_{a-x}^{b-x}\frac{p'(u)}{p(u)}p(u)\,du
  \nonumber\\
  &=\int_{a-x}^{b-x}p'(u)\,du
  \nonumber\\
  &=p(b-x)-p(a-x),
\end{align}
which, together with~\eqref{eq:Lxa}, implies that $L_{xa}(x,a,b)\le0$ holds.
Combined with $L_{xx}(x,a,b)\ge0$ and~\eqref{eq:hatxa},
it in turn proves $\hat{x}_a\ge0$. 
Similarly, one has 
\begin{align}
  \label{eq:Lxbpos}
  \frac{p'(b-x)}{p(b-x)}[P(b-x)-P(a-x)]
  &=\frac{p'(b-x)}{p(b-x)}\int_{a-x}^{b-x}p(u)\,du
  \nonumber\\
  &\le\int_{a-x}^{b-x}\frac{p'(u)}{p(u)}p(u)\,du
  \nonumber\\
  &=\int_{a-x}^{b-x}p'(u)\,du
  \nonumber\\
  &=p(b-x)-p(a-x),
\end{align}
which, together with~\eqref{eq:Lxb}, implies that $L_{xb}(x,a,b)\le0$ holds. 
Combined with \eqref{eq:hatxb}, it in turn 
proves $\hat{x}_b\ge0$.

We next discuss the case where $p$ is not necessarily 
differentiable. 
Since $p(u)$ is log-concave, $q(u)=-\log p(u)$ is convex, 
so that it is continuous and differentiable 
except on a countable set. 
Let $\phi(u)$ be any function such that 
$\phi(u)$ takes a value in the subderivative of $q(u)$ 
for any $u$. 
As $p(u)=e^{-q(u)}$, one has, at any point $u$ at which 
$p(u)$ is differentiable, 
\begin{equation}
    p'(u)=-q'(u)e^{-q(u)}=-\phi(u)p(u).
\end{equation}
One therefore has, for any $b>a$, 
\begin{equation}
    p(b)-p(a)=-\int_a^b\phi(u)p(u)\,du,
\end{equation}
which can be proved in the same way as~\cite[Theorem 1.28]{Simon2011}. 
On the other hand, 
we know that $\phi(u)$ is monotonically non-decreasing in $u$. 
One thus has
\begin{equation}
    \label{eq:pos}
    \phi(a)[P(b)-P(a)]
    =\phi(a)\int_a^b p(u)\,du
    \le\int_a^b\phi(u)p(u)\,du
    \le\phi(b)\int_a^b p(u)\,du
    =\phi(b)[P(b)-P(a)],
\end{equation}
which, via replacing $(a,b)$ with $(a-x,b-x)$, 
proves inequalities corresponding to~\eqref{eq:Lxapos} 
and \eqref{eq:Lxbpos}.

One has therefore proven the following proposition.
\begin{proposition}
\label{prop:monotone}
  For $a<b$, let $L(x,a,b)$ be as defined in~\eqref{eq:defL}, and 
  let $\hat{x}(a,b)$ be the minimizer
  of the optimization problem~\eqref{eq:optL}.
  Then, for $a<b$ and $a'<b'$ with $a\le a'$ and $b\le b'$,
one has $\hat{x}(a,b)\le\hat{x}(a',b')$.
\end{proposition}
We would like to note that if $p$ is differentiable and
\emph{strictly} log-concave,
then we have strict inequalities in the above proposition, 
which in turn implies, via~\eqref{eq:Lxx}, 
that $L_{xx}(x,a,b)\gneq0$. 
This constitutes an alternative proof 
of the strict convexity of $L(x,a,b)$ in $x$ 
under the differentiability assumption.
Even without the differentiability assumption, 
one can note that, under the strict log-concavity 
of $p$, $\phi(u)$ is increasing, 
so that the strict inequalities hold in~\eqref{eq:pos},
proving the strict inequalities in the above proposition as well.

The above argument proves the first half of ONC3, that is,
for any fixed $w$
one has $z_q^*\le z_{q+1}^*$ for $q\in\{1,\ldots,Q-1\}$.
Furthermore, if $g'$ is strictly log-concave, 
one has the strict ordering when $w\not=0$: 
$z_q^*<z_{q+1}^*$ for $q\in\{1,\ldots,Q-1\}$,
thereby proving the latter half of ONC3.

\def\putabove#1#2{\setbox0=\hbox{$#1$}%
  \hbox{\vbox to\ht0{\vss\hbox to\wd0{\hss$#2$\hss}\vspace{1mm}\hbox to\wd0{\hss$\vee$\hss}\vspace{1mm}%
      \hbox{$#1$}}}}

\section{Details of the Experimental Setup}\Lsec{Details of the Experimental Setup}

\paragraph{Dataset statistics.}
We used the five publicly available real-world OR tabular datasets of \citet{gutierrez2016}\footnote{\url{https://www.uco.es/grupos/ayrna/orreview}.}---ER, LE, and SW (Employee rejection/acceptance, Lecturers evaluation, and Social workers decisions (public domain)~\citep{BenDavid1992}), CA (Car evaluation (CC BY 4.0)~\citep{Bohanec1988}), and WR (Wine quality---Red (CC BY 4.0)~\citep{Cortez2009})---exactly as released.  
The website offers 30 pre-defined training–validation hold-out splits whose label distributions are identical across the two partitions.  
We ran our experiments on all the 30 splits for each dataset and report the averages. 

Table~\ref{tab:datasets} summarizes the key statistics of the five datasets used in our study.  
For consistency, all ordinal labels were remapped to consecutive integers starting from one. We constructed input vectors by concatenating two types of preprocessed attributes: one-hot encoded categorical attributes and normalized numerical attributes.
\begin{table}[htbp]
  \centering
  \caption{Summary of tabular datasets used in the experiments. Attr. denotes attributes and Input dim. denotes the dimension of input vectors.}
  \label{tab:datasets}
  \scriptsize
  \begin{tabularx}{\textwidth}{@{} l l r r r r X @{} }
    \toprule
    \textbf{Dataset (code)} & \textbf{Subset} & \textbf{\#Samples} & \textbf{\#Attr.} & \textbf{Input dim.} & \textbf{\#Classes} & \textbf{Distribution} \\
    & & (per split) & & & (Q) & (counts per label) \\
    \midrule
    \multirow{2}{*}{ERA (ER)} 
      & Train &  750 & \multirow{2}{*}{4}  & \multirow{2}{*}{13} & \multirow{2}{*}{9} & [1:69, 2:106, 3:136, 4:129, 5:118, 6:89, 7:66, 8:23, 9:14] \\
      & Val  &  250 & & & & [1:23, 2:36, 3:45, 4:43, 5:40, 6:29, 7:22, 8:8, 9:4] \\
    \addlinespace
    \multirow{2}{*}{LEV (LE)} 
      & Train &  750 & \multirow{2}{*}{4}  & \multirow{2}{*}{9} & \multirow{2}{*}{5} & [0:70, 1:210, 2:302, 3:148, 4:20] \\
      & Val  &  250 & & & & [0:23, 1:70, 2:101, 3:49, 4:7] \\
    \addlinespace
    \multirow{2}{*}{SWD (SW)} 
      & Train &  750 & \multirow{2}{*}{10} & \multirow{2}{*}{14} & \multirow{2}{*}{4} & [2:24, 3:264, 4:299, 5:163] \\
      & Val  &  250 & & & & [2:8, 3:88, 4:100, 5:54] \\
    \addlinespace
    \multirow{2}{*}{Car (CA)} 
      & Train & 1296 & \multirow{2}{*}{6}  & \multirow{2}{*}{21} & \multirow{2}{*}{4} & [acc:288, good:52, unacc:907, vgood:49] \\
      & Val  &  432 & & & & [acc:96,  good:17,  unacc:303, vgood:16] \\
    \addlinespace
    \multirow{2}{*}{Wine (WR)} 
      & Train & 1199 & \multirow{2}{*}{11} & \multirow{2}{*}{17} & \multirow{2}{*}{6} & [3:8, 4:39, 5:510, 6:479, 7:150, 8:13] \\
      & Val  &  400 & & & & [3:2, 4:14, 5:171, 6:159, 7:49, 8:5] \\
    \bottomrule
  \end{tabularx}
\end{table}

Additionally, we conducted experiments on the UTKFace dataset (non-commercial research use)~\citep{zhang2017age}, a large-scale face image dataset for age estimation. The dataset contains 23,708 facial images with age labels. We grouped ages into 20 ordinal classes using 5-year intervals. The dataset exhibits natural class imbalance, with the most frequent class (25--29 years) containing 5,034 samples and the rarest class (95--116 years) containing only 67 samples. Following standard practice, we performed a stratified 80/20 train-validation split to maintain class distribution across subsets. All images were resized to $224 \times 224$ pixels and normalized using ImageNet statistics for pre-trained backbone compatibility. Table~\ref{tab:utkface_class_dist} presents the detailed class distribution.

\begin{table}[htbp]
  \centering
  \caption{Class distribution of UTKFace dataset with 5-year age grouping.}
  \label{tab:utkface_class_dist}
  \scriptsize
  \begin{tabular}{c l r r | c l r r}
    \toprule
    \textbf{Class} & \textbf{Age} & \textbf{Train} & \textbf{Val} & \textbf{Class} & \textbf{Age} & \textbf{Train} & \textbf{Val} \\
    \midrule
    0 & [0--4] & 1,733 & 434 & 10 & [50--54] & 1,076 & 269 \\
    1 & [5--9] & 716 & 179 & 11 & [55--59] & 763 & 191 \\
    2 & [10--14] & 471 & 118 & 12 & [60--64] & 586 & 146 \\
    3 & [15--19] & 753 & 189 & 13 & [65--69] & 469 & 117 \\
    4 & [20--24] & 1,848 & 462 & 14 & [70--74] & 298 & 75 \\
    5 & [25--29] & 4,027 & 1,007 & 15 & [75--79] & 261 & 65 \\
    6 & [30--34] & 1,832 & 458 & 16 & [80--84] & 190 & 47 \\
    7 & [35--39] & 1,797 & 450 & 17 & [85--89] & 214 & 53 \\
    8 & [40--44] & 945 & 236 & 18 & [90--94] & 82 & 20 \\
    9 & [45--49] & 851 & 213 & 19 & [95--116] & 54 & 13 \\
    \midrule
    \multicolumn{4}{c|}{\textbf{Total:}} & \multicolumn{2}{c}{18,966} & \multicolumn{2}{c}{4,742} \\
    \bottomrule
  \end{tabular}
\end{table}

\textbf{Overparameterized network.}
Here we describe the neural network architecture used in our experiments.
For the tabular datasets, we employed a multilayer perceptron with residual connections as follows:
\begin{itemize}[leftmargin=*]
  \item An input $\bm{x}\in\mathbb{R}^{d}$ is first mapped to a 128-dimensional representation by a linear layer, then passed through a parametric rectified linear unit (PReLU)~\citep{he2015delving}.
  \item It then passes through four residual blocks, each defined as
        \[
          \bm{x}\;\mapsto\;\bm{x}+\mathrm{PReLU}\bigl(W_{2}\,
                 \mathrm{PReLU}(W_{1}\bm{x}+\bm{b}_{1})+\bm{b}_{2}\bigr),
          \quad
                  W_{1},W_{2} \in \mathbb{R}^{128\times128},\;
                  \bm{b}_{1},\bm{b}_{2} \in \mathbb{R}^{128}.
        \]
  \item It subsequently passes through three consecutive linear layers with linear activation, yielding the 64-dimensional feature $\bm{h}_\theta(\bm{x})$.
  \item Finally, a linear layer without bias---whose weight vector is the classifier weight analyzed in this study---maps the $\bm{h}_\theta(\bm{x})$ to the one-dimensional latent variable
        $z = \bm{w}^{\top}\bm{h}_\theta(\bm{x})$.
\end{itemize}

\begin{figure}[!h]
  \vspace{-6pt}                 
  \centering
  \includegraphics[width=\textwidth]{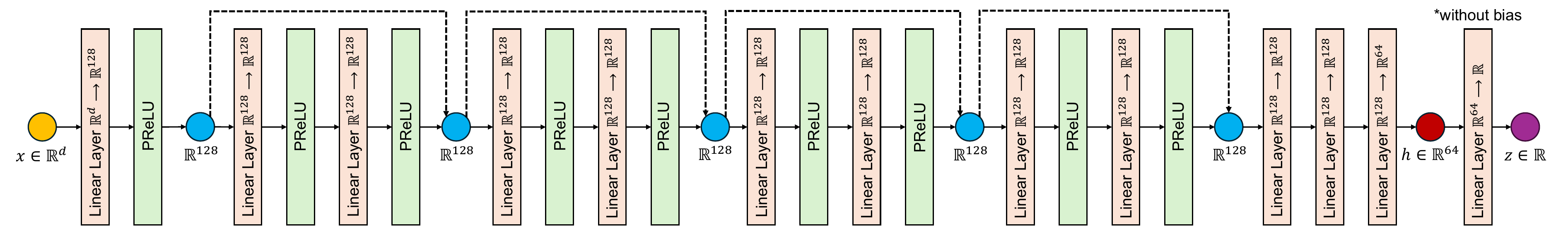}
  \caption{Architecture of the overparameterized network for tabular datasets.}
  \label{fig:neural_network}
  \vspace{-6pt}                 
\end{figure}

The PReLU activations together with the linear tail give the network enough flexibility to map inputs to any location in the feature space, aligning with the UFM assumption. Figure~\ref{fig:neural_network} illustrates the architecture of the network.

For the UTKFace dataset, we employed pre-trained ResNet50, ResNet101~\citep{he2016deep}, and DenseNet201~\citep{huang2017densely} as backbones. To better satisfy the UFM assumption, we augmented each backbone with two additional linear layers that maintain the same dimension as the backbone output. The mapped features $\bm{h}_\theta(\bm{x})$ are then passed through a final linear layer without bias to produce the one-dimensional latent variable $z = \bm{w}^{\top}\bm{h}_\theta(\bm{x})$, where $\bm{w}$ is the classifier weight analyzed in this study.

\textbf{Other experimental settings.}
For the tabular datasets, all the models were trained for \num{5000} epochs with the Adam optimizer,
starting from the initial learning rates listed in Table~\ref{tab:exp-settings}.
We made the learning rate to decay by a factor of~\num{0.1} at epochs~\num{200}, \num{800}, and~\num{3000}.
We used a batch size of~\num{2048} and applied a weight decay
of \(5\times10^{-3}\) to all network parameters
(except that the learnable thresholds, when present,
received zero weight decay). For each dataset we tested the four possible combinations 
of the three link functions (logit, probit, clog-log) and the two
threshold strategies (fixed, learnable).
Running every configuration on the 30 predefined hold-out splits
yielded a total of 600 training–validation runs.

To ensure that the training error reached (near-)zero---the regime where
feature collapse is observed---we tuned only the initial learning rate and,
for fixed thresholds, the threshold range.
Because the logistic function has heavier tails than the normal CDF,
the logit runs used a wider fixed range \([-20,20]\),
whereas the probit runs used \([-2,2]\).
The complete hyper-parameter grid is summarized in
Table~\ref{tab:exp-settings}.

For the UTKFace dataset, models were trained for 300 epochs using the Adam optimizer with a batch size of 512 per GPU across 5 GPUs (total effective batch size 2560). We tested two threshold strategies (fixed, learnable), using only the logit link function. Starting from the initial learning rates listed in Table~\ref{tab:utkface-settings}, the learning rate was decayed by a factor of 0.1 at epochs 100 and 200. Each configuration was repeated with three different random seeds. The other hyper-parameter settings for each backbone are summarized in Table~\ref{tab:utkface-settings}.

\paragraph{Experiments compute resources.}
For the tabular datasets, every run was executed on a single NVIDIA RTX~A6000 (48 GB) GPU, taking roughly 5–10 minutes and requiring only a few hundred megabytes of GPU memory. For the UTKFace dataset, training was performed across 5 NVIDIA RTX~A6000 GPUs using distributed data parallel, with each configuration taking approximately 6 hours to complete 300 epochs.

\begin{table}[htbp]
  \centering
  \caption{Hyper-parameter settings for tabular datasets with three link functions (logit, probit, clog-log) and two threshold strategies (fixed, learnable).}
  \label{tab:exp-settings}
  \scriptsize
  \begin{tabular}{ccccc}
    \toprule
    \textbf{Dataset} & \textbf{Link function} & \textbf{Thresholds} &
    \textbf{Threshold range (when fixed)} & \textbf{Initial learning rate} \\
    \midrule
    \multirow{6}{*}{ER}
      & logit    & fixed     & $[-20,\,20]$ & $1\times10^{-2}$ \\ 
      & logit    & learnable & --           & $1\times10^{-2}$ \\ 
      & probit   & fixed     & $[-2,\,2]$   & $1\times10^{-3}$ \\ 
      & probit   & learnable & --           & $1\times10^{-3}$ \\ 
      & clog-log & fixed     & $[-2,\,2]$   & $5\times10^{-3}$ \\ 
      & clog-log & learnable & --           & $5\times10^{-3}$ \\ 
    \addlinespace
    \multirow{6}{*}{LE}
      & logit    & fixed     & $[-20,\,20]$ & $1\times10^{-2}$ \\ 
      & logit    & learnable & --           & $1\times10^{-2}$ \\ 
      & probit   & fixed     & $[-2,\,2]$   & $5\times10^{-3}$ \\ 
      & probit   & learnable & --           & $5\times10^{-3}$ \\ 
      & clog-log & fixed     & $[-2,\,2]$   & $5\times10^{-3}$ \\ 
      & clog-log & learnable & --           & $5\times10^{-3}$ \\ 
    \addlinespace
    \multirow{6}{*}{SW}
      & logit    & fixed     & $[-20,\,20]$ & $1\times10^{-2}$ \\ 
      & logit    & learnable & --           & $1\times10^{-2}$ \\ 
      & probit   & fixed     & $[-2,\,2]$   & $5\times10^{-3}$ \\ 
      & probit   & learnable & --           & $5\times10^{-3}$ \\ 
      & clog-log & fixed     & $[-2,\,2]$   & $5\times10^{-3}$ \\ 
      & clog-log & learnable & --           & $5\times10^{-3}$ \\ 
    \addlinespace
    \multirow{6}{*}{CA}
      & logit    & fixed     & $[-20,\,20]$ & $1\times10^{-2}$ \\ 
      & logit    & learnable & --           & $1\times10^{-2}$ \\ 
      & probit   & fixed     & $[-2,\,2]$   & $5\times10^{-3}$ \\ 
      & probit   & learnable & --           & $5\times10^{-3}$ \\ 
      & clog-log & fixed     & $[-2,\,2]$   & $5\times10^{-3}$ \\ 
      & clog-log & learnable & --           & $5\times10^{-3}$ \\ 
    \addlinespace
    \multirow{6}{*}{WR}
      & logit    & fixed     & $[-20,\,20]$ & $1\times10^{-2}$ \\ 
      & logit    & learnable & --           & $1\times10^{-2}$ \\ 
      & probit   & fixed     & $[-2,\,2]$   & $1\times10^{-3}$ \\ 
      & probit   & learnable & --           & $1\times10^{-3}$ \\ 
      & clog-log & fixed     & $[-2,\,2]$   & $1\times10^{-3}$ \\ 
      & clog-log & learnable & --           & $1\times10^{-3}$ \\ 
    \bottomrule
  \end{tabular}
\end{table}

\begin{table}[htbp]
  \centering
  \caption{Hyper-parameter settings for UTKFace experiments with logit link function and two threshold strategies (fixed, learnable).}
  \label{tab:utkface-settings}
  \scriptsize
  \begin{tabular}{ccccc}
    \toprule
    \textbf{Backbone} & \textbf{Thresholds} & \textbf{Threshold range (when fixed)} & \textbf{Initial learning rate} & \textbf{Weight decay} \\
    \midrule
    \multirow{2}{*}{ResNet50}
      & fixed     & $[-40,\,40]$ & $1\times10^{-3}$ & $1\times10^{-4}$ \\ 
      & learnable & --           & $1\times10^{-3}$ & $1\times10^{-4}$ \\ 
    \addlinespace
    \multirow{2}{*}{ResNet101}
      & fixed     & $[-40,\,40]$ & $1\times10^{-3}$ & $1\times10^{-3}$ \\ 
      & learnable & --           & $1\times10^{-3}$ & $1\times10^{-3}$ \\ 
    \addlinespace
    \multirow{2}{*}{DenseNet201}
      & fixed     & $[-40,\,40]$ & $5\times10^{-4}$ & $1\times10^{-4}$ \\ 
      & learnable & --           & $5\times10^{-4}$ & $1\times10^{-4}$ \\ 
    \bottomrule
  \end{tabular}
\end{table}


\renewcommand\topfraction{.99}
\renewcommand\textfraction{.01}
\newpage

\section{Additional experimental results}\Lsec{Additional}

\subsection{Results with the logit model}
\begin{figure}[!h]
  \vspace{-6pt}                 
  \centering
  \includegraphics[width=\textwidth]{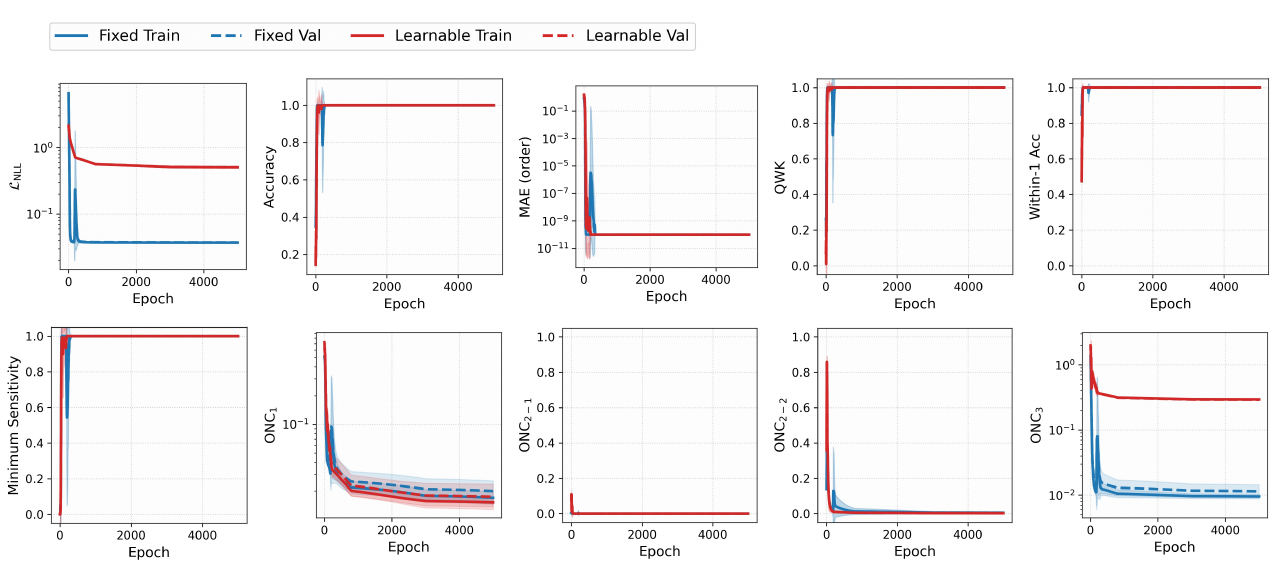}
  \caption{Epoch-wise average metrics curves for the LE dataset with the logit model, comparing fixed- and learnable-threshold models.}
  \label{fig:metric_curves_logit_LE}
  \vspace{-6pt}                 
\end{figure}

\begin{figure}[!h]
  \vspace{-6pt}                 
  \centering
  \includegraphics[width=\textwidth]{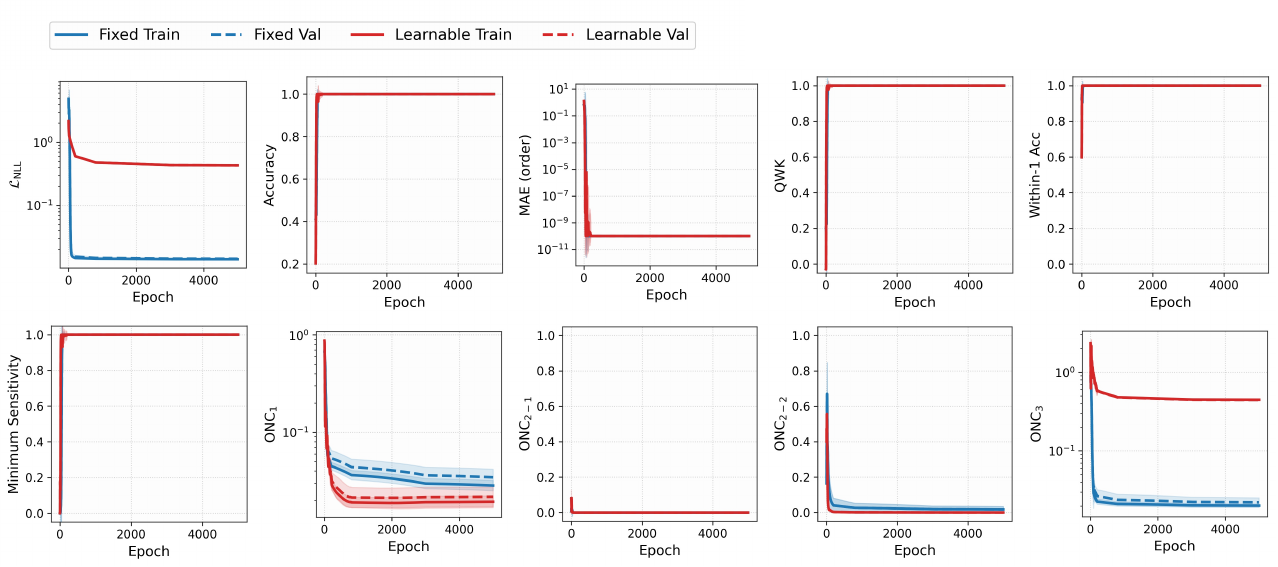}
  \caption{Epoch-wise average metrics curves for the SW dataset with the logit model, comparing fixed- and learnable-threshold models.}
  \label{fig:metric_curves_logit_SW}
  \vspace{-6pt}                 
\end{figure}

\begin{figure}[!h]
  \vspace{-6pt}                 
  \centering
  \includegraphics[width=\textwidth]{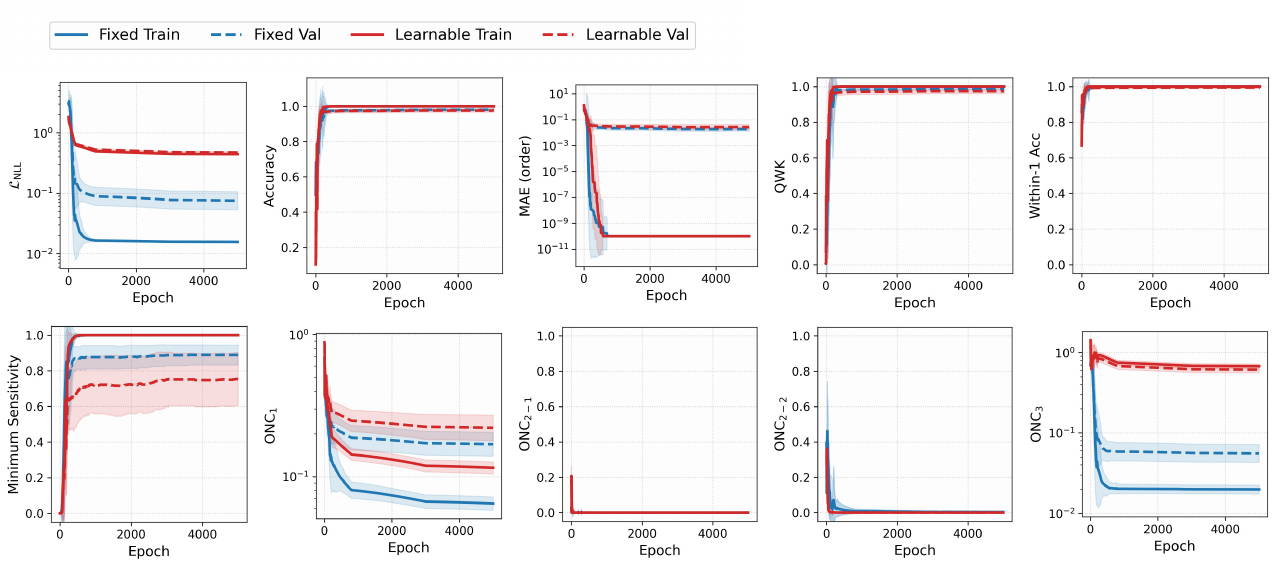}
  \caption{Epoch-wise average metrics curves for the CA dataset with the logit model, comparing fixed- and learnable-threshold models.}
  \label{fig:metric_curves_logit_CA}
  \vspace{-6pt}                 
\end{figure}

\begin{figure}[!h]
  \vspace{-6pt}                 
  \centering
  \includegraphics[width=\textwidth]{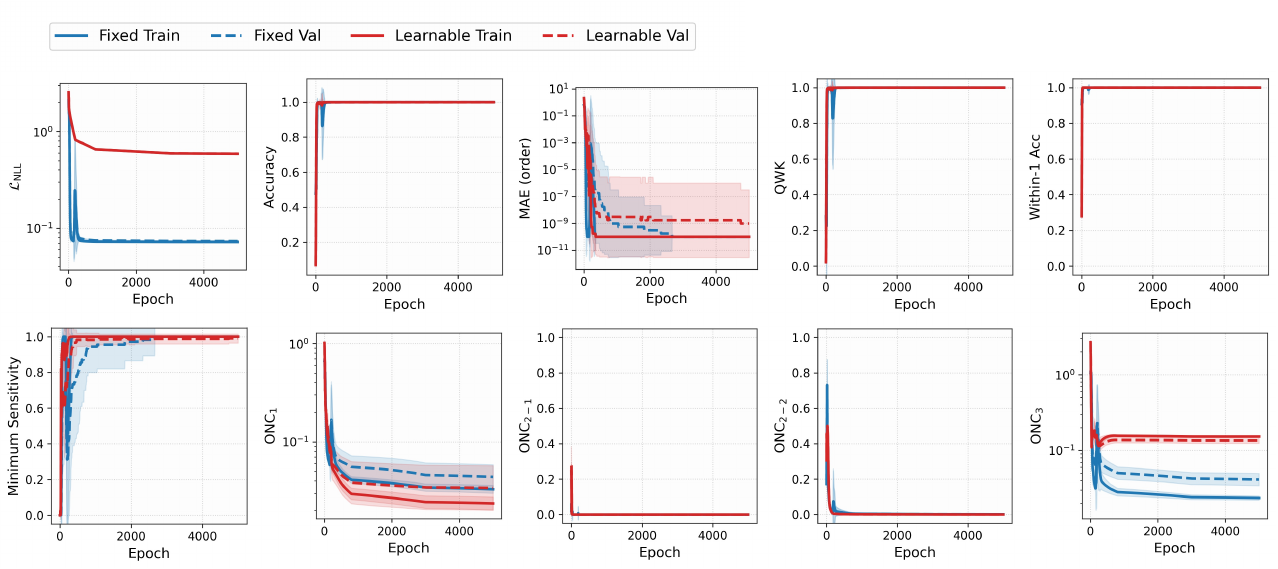}
  \caption{Epoch-wise average metrics curves for the WR dataset with the logit model, comparing fixed- and learnable-threshold models.}
  \label{fig:metric_curves_logit_WR}
  \vspace{-6pt}                 
\end{figure}

\begin{figure}[!h]
  \vspace{-6pt} 
  \centering
  \includegraphics[width=\textwidth]{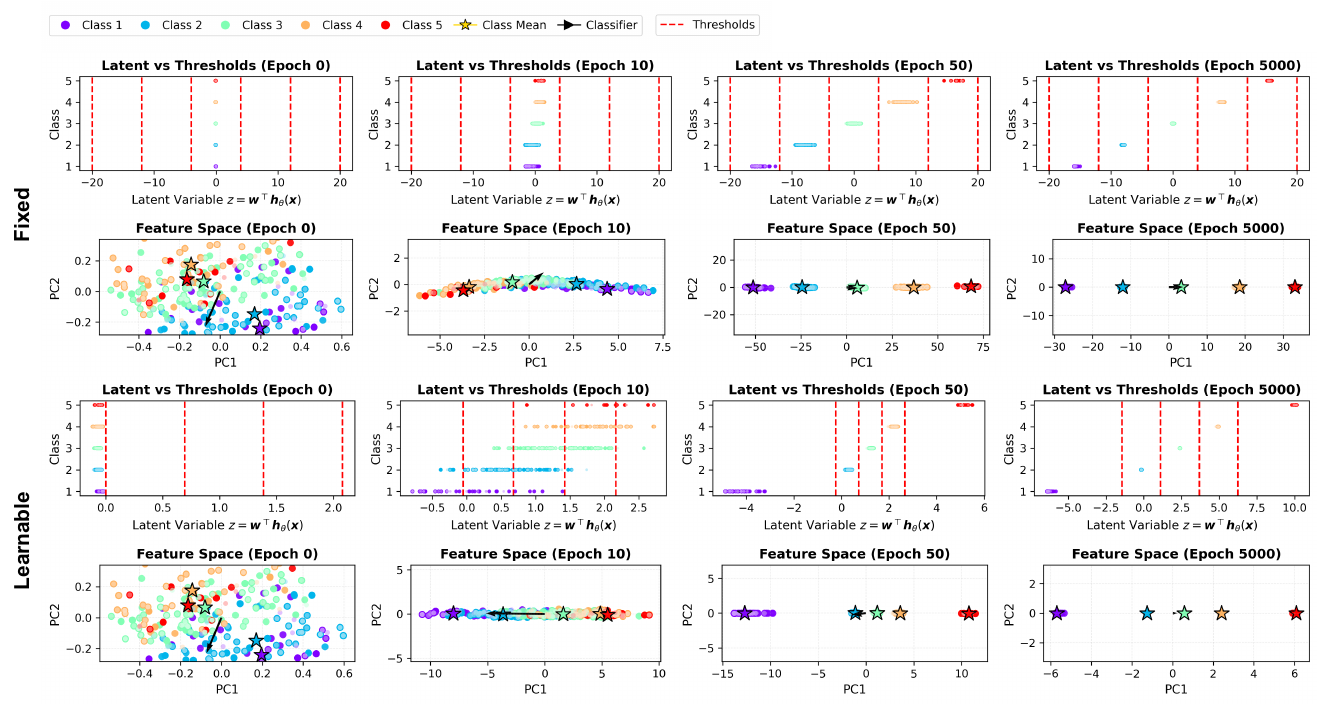}
  \captionsetup{skip=3pt}
  \caption{Visualization of the latent and feature spaces for the LE dataset using the logit model, comparing fixed- and learnable-threshold models.}
  \label{fig:vis_logit_LE}
  \vspace{-6pt} 
\end{figure}

\begin{figure}[!h]
  \vspace{-6pt} 
  \centering
  \includegraphics[width=\textwidth]{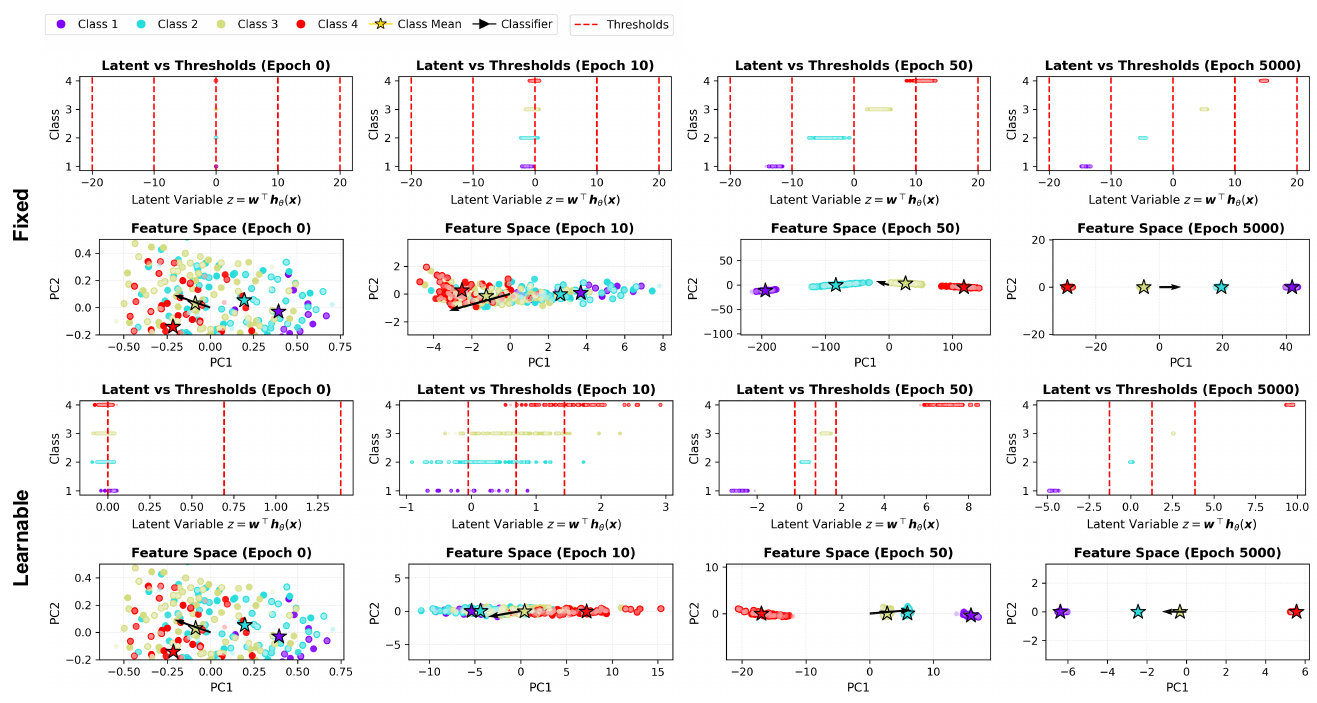}
  \captionsetup{skip=3pt}
  \caption{Visualization of the latent and feature spaces for the SW dataset using the logit model, comparing fixed- and learnable-threshold models.}
  \label{fig:vis_logit_SW}
  \vspace{-6pt} 
\end{figure}

\begin{figure}[!h]
  \vspace{-6pt} 
  \centering
  \includegraphics[width=\textwidth]{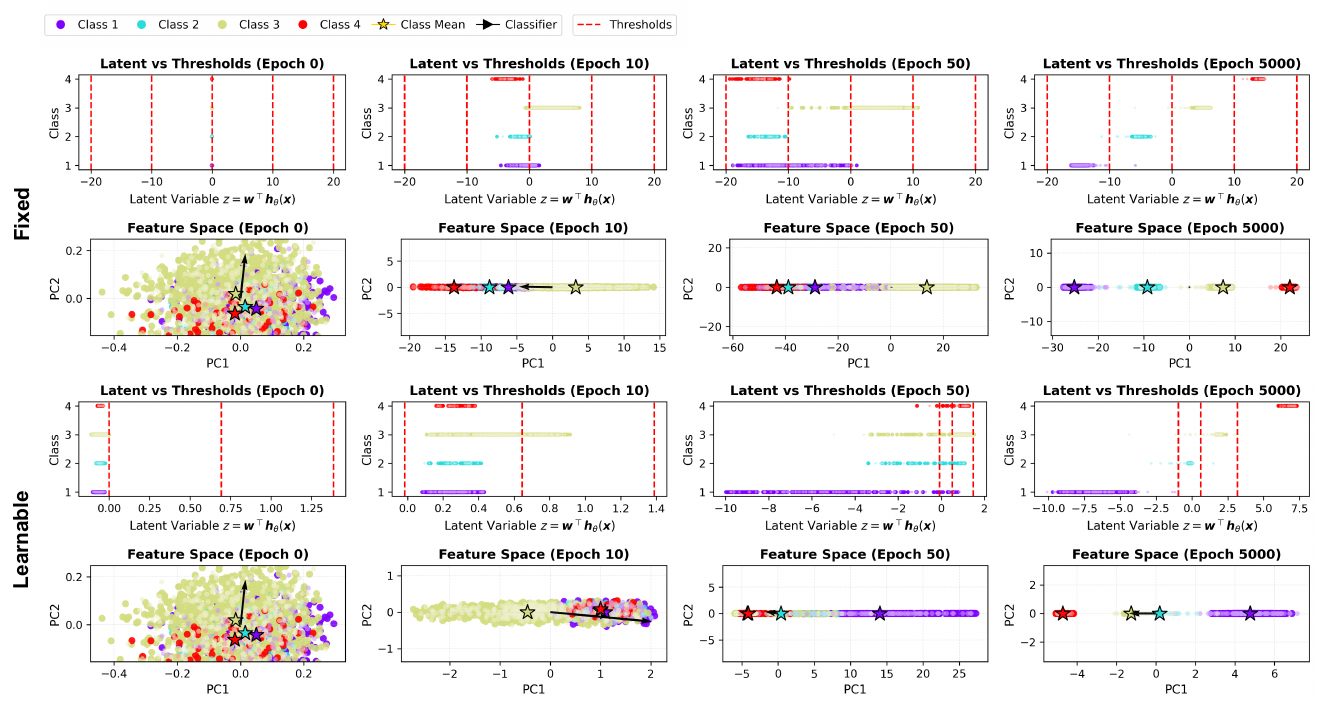}
  \captionsetup{skip=3pt}
  \caption{Visualization of the latent and feature spaces for the CA dataset using the logit model, comparing fixed- and learnable-threshold models.}
  \label{fig:vis_logit_CA}
  \vspace{-6pt} 
\end{figure}

\begin{figure}[!h]
  \vspace{-6pt} 
  \centering
  \includegraphics[width=\textwidth]{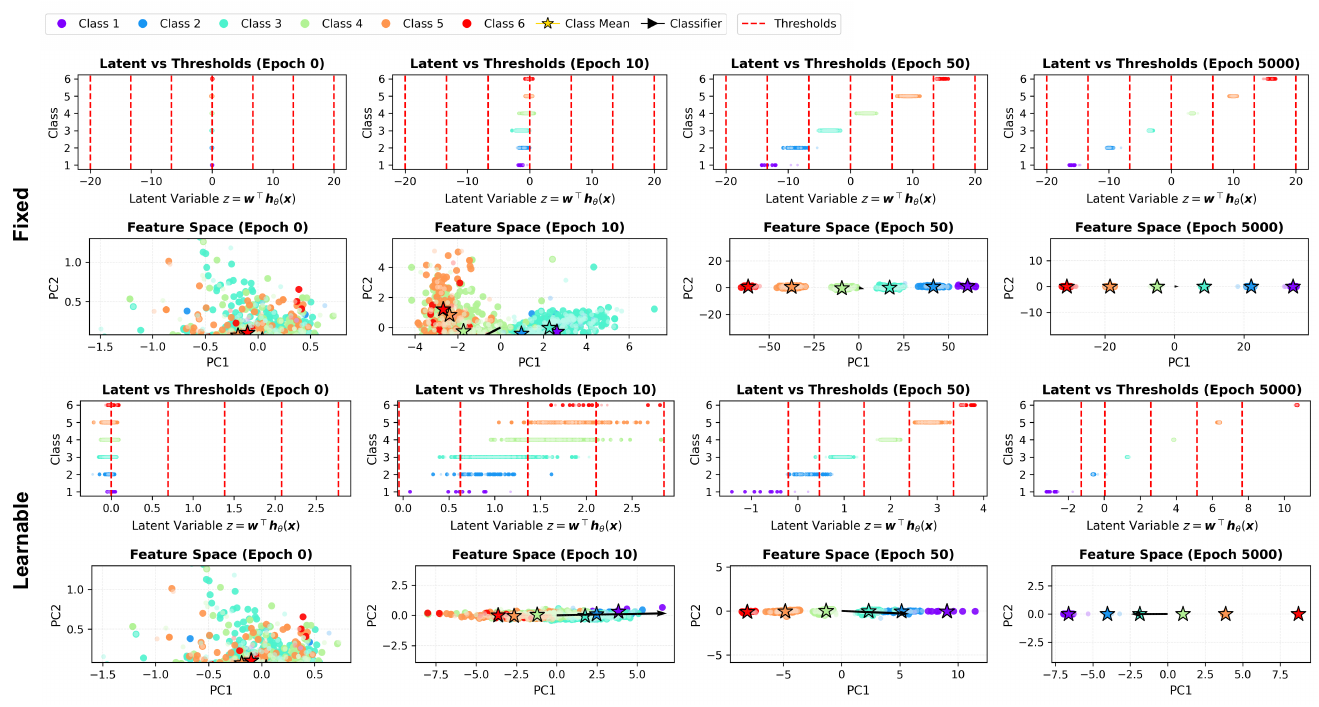}
  \captionsetup{skip=3pt}
  \caption{Visualization of the latent and feature spaces for the WR dataset using the logit model, comparing fixed- and learnable-threshold models.}
  \label{fig:vis_logit_WR}
  \vspace{-6pt} 
\end{figure}

\Lsec{logit}
This section presents the experimental outcomes obtained using the logistic function (i.e., $g(x)=(1+e^{-x})^{-1}$), which corresponds to the logit model.
Figures~\ref{fig:metric_curves_logit_LE}--\ref{fig:metric_curves_logit_WR} show 
the evolution of evaluation-metric curves 
for the datasets LE, SW, CA, and WR, respectively.
Figures~\ref{fig:vis_logit_LE}--\ref{fig:vis_logit_WR} 
show visualization of the latent and feature spaces 
for the datasets LE, SW, CA, and WR, respectively.
These exhibit a consistent behavior with the one of the ER dataset in the main text (Figs.~\ref{fig:metric_curves_logit_ER} and \ref{fig:vis_logit_ER}).

\clearpage 

\subsection{Results with the probit model}
\begin{figure}[!h]
  \vspace{-6pt}                 
  \centering
  \includegraphics[width=\textwidth]{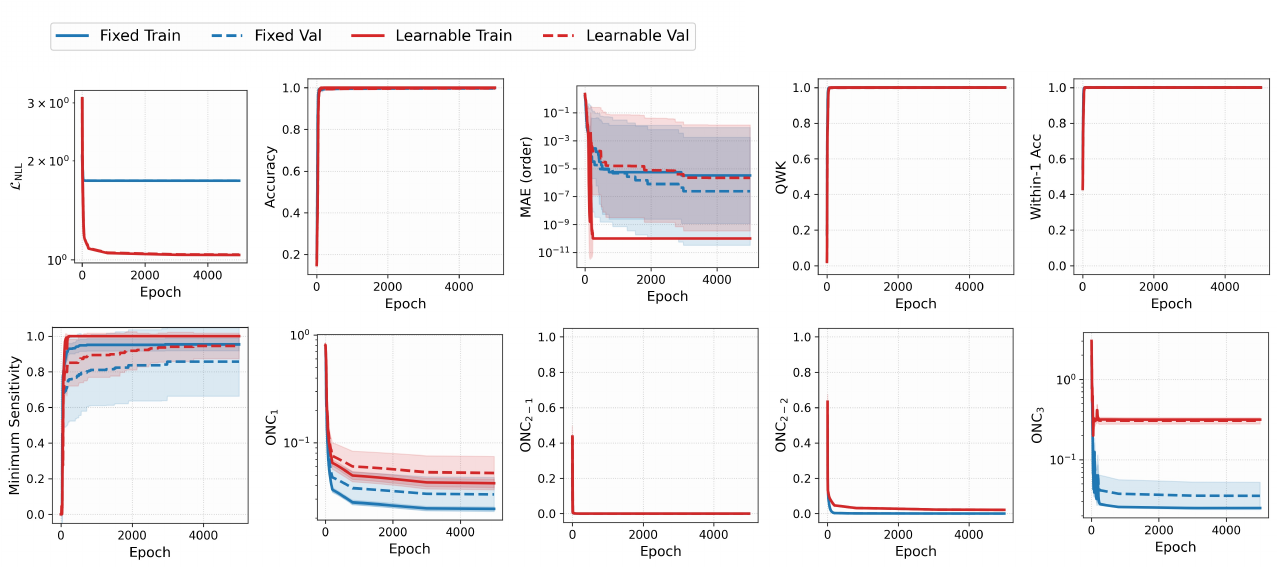}
  \caption{Epoch-wise average metrics curves for the ER dataset with the probit model, comparing fixed- and learnable-threshold models.}
  \label{fig:metric_curves_probit_ER}
  \vspace{-6pt}                 
\end{figure}

\begin{figure}[!h]
  \vspace{-6pt}                 
  \centering
  \includegraphics[width=\textwidth]{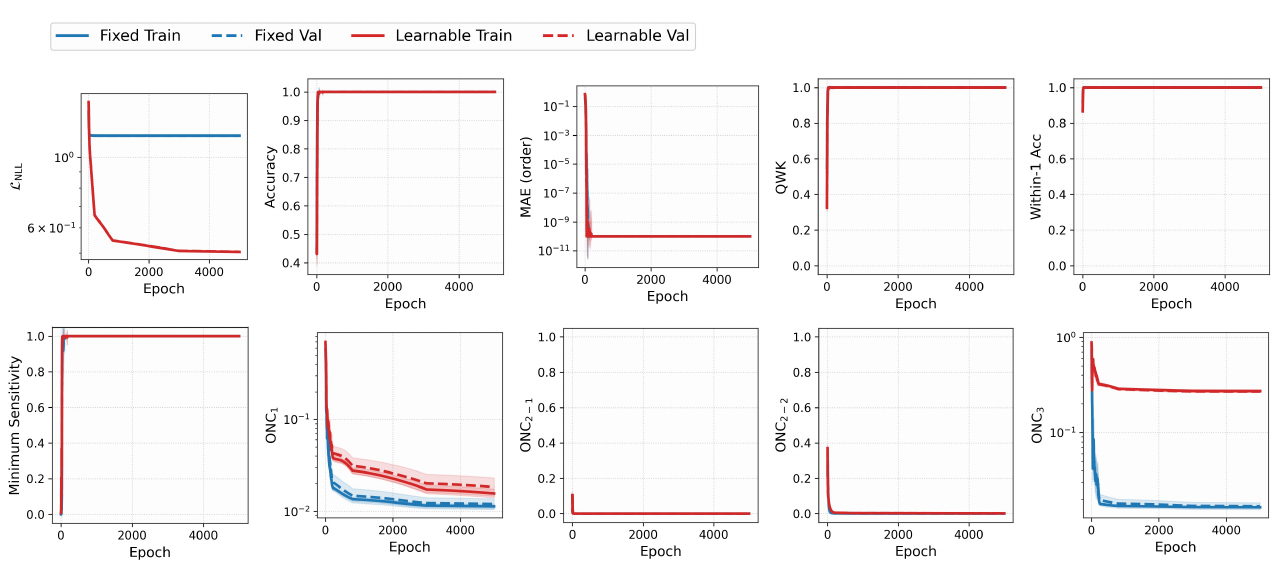}
  \caption{Epoch-wise average metrics curves for the LE dataset with the probit model, comparing fixed- and learnable-threshold models.}
  \label{fig:metric_curves_probit_LE}
  \vspace{-6pt}                 
\end{figure}

\begin{figure}[!htbp]
  \vspace{-6pt}                 
  \centering
  \includegraphics[width=\textwidth]{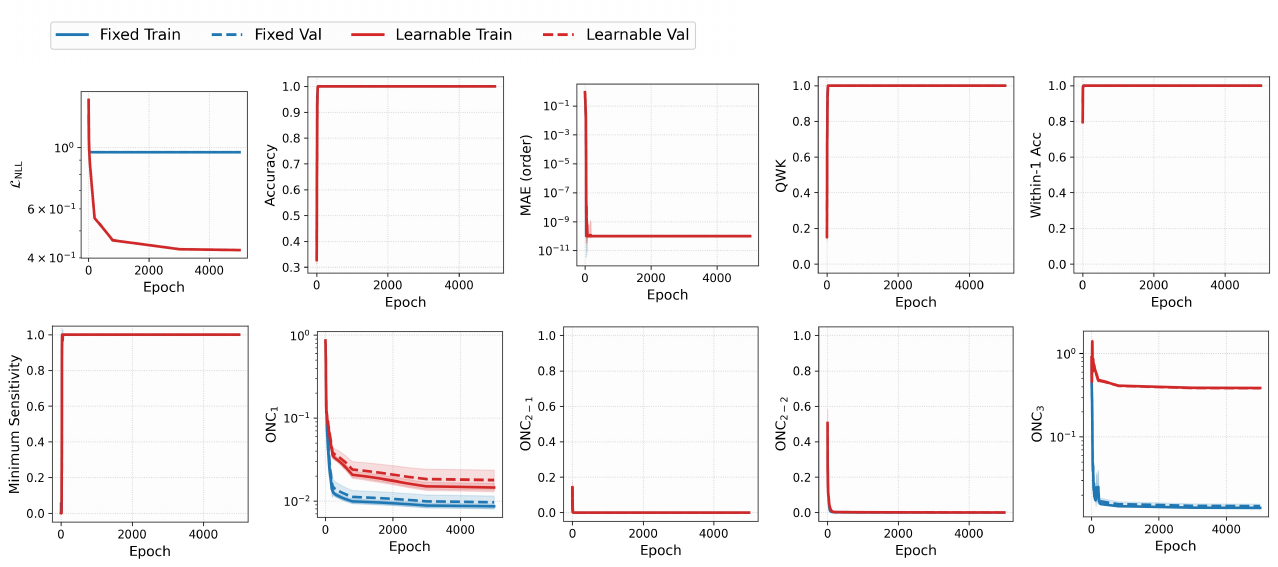}
  \caption{Epoch-wise average metrics curves for the SW dataset with the probit model, comparing fixed- and learnable-threshold models.}
  \label{fig:metric_curves_probit_SW}
  \vspace{-6pt}                 
\end{figure}

\begin{figure}[!htbp]
  \vspace{-6pt}                 
  \centering
  \includegraphics[width=\textwidth]{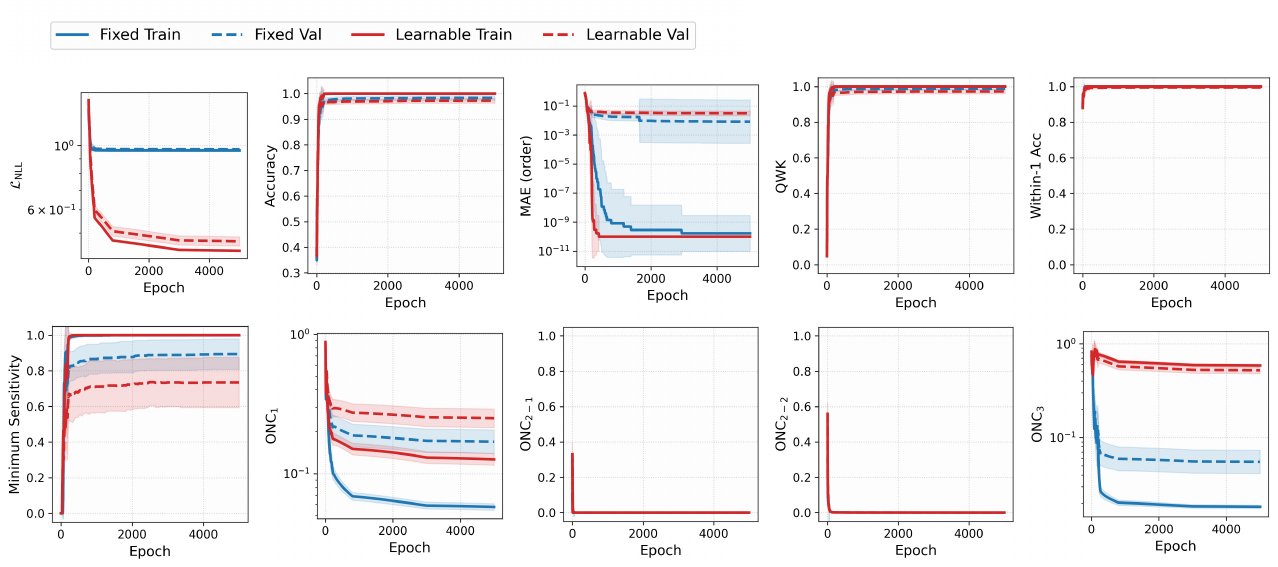}
  \caption{Epoch-wise average metrics curves for the CA dataset with the probit model, comparing fixed- and learnable-threshold models.}
  \label{fig:metric_curves_probit_CA}
  \vspace{-6pt}                 
\end{figure}

\begin{figure}[!htbp]
  \vspace{-6pt}                 
  \centering
  \includegraphics[width=\textwidth]{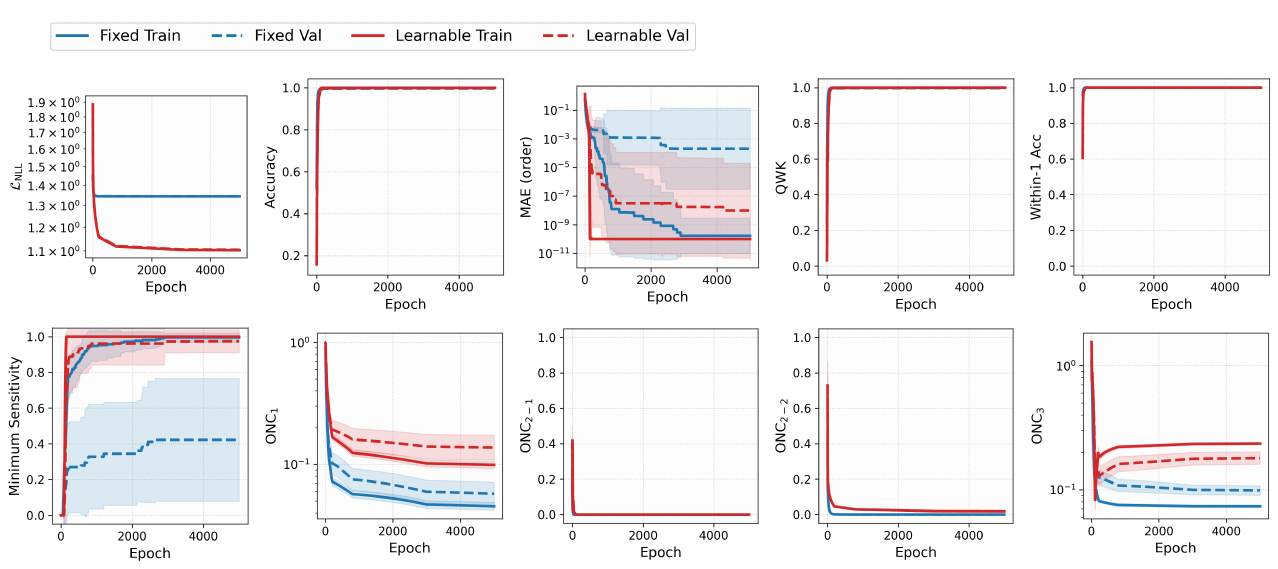}
  \caption{Epoch-wise average metrics curves for the WR dataset with the probit model, comparing fixed- and learnable-threshold models.}
  \label{fig:metric_curves_probit_WR}
  \vspace{-6pt}                 
\end{figure}

\begin{figure}[htbp]
  \vspace{-6pt} 
  \centering
  \includegraphics[width=\textwidth]{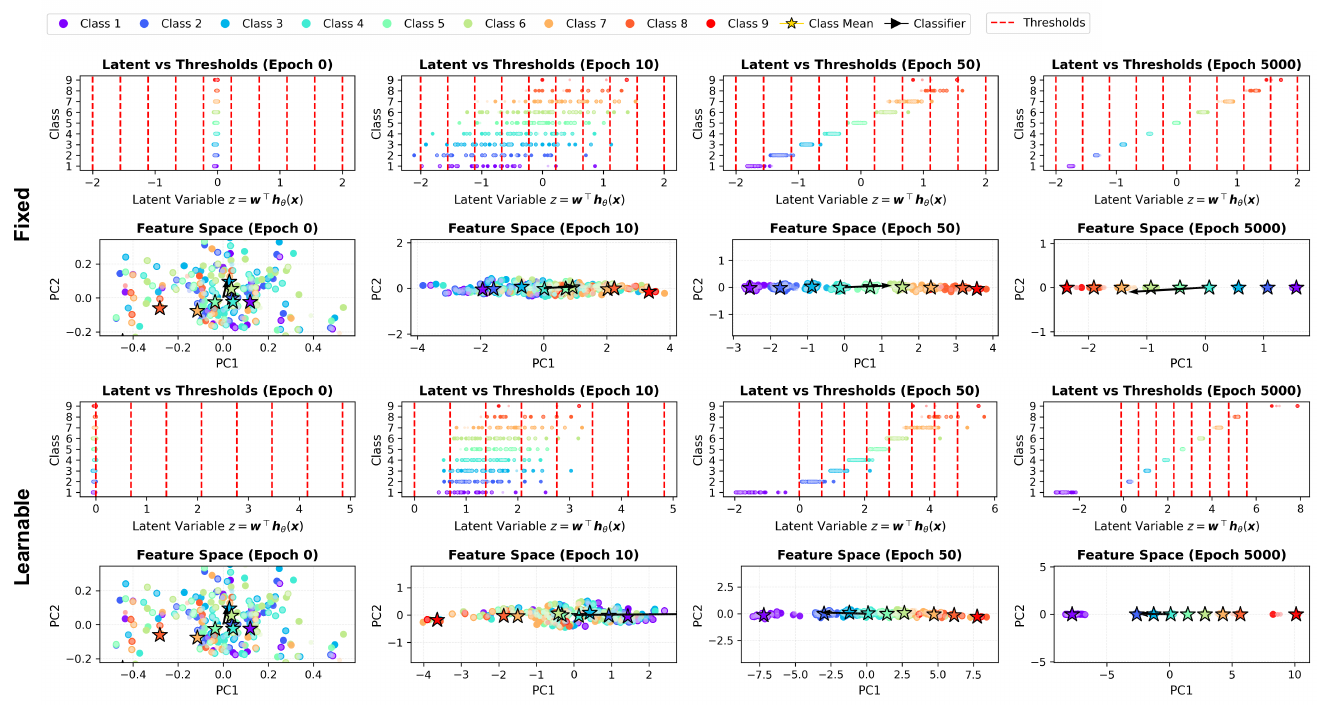}
  \captionsetup{skip=3pt}
  \caption{Visualization of the latent and feature spaces for the ER dataset using the probit model, comparing fixed- and learnable-threshold models.}
  \label{fig:vis_probit_ER}
  \vspace{-6pt} 
\end{figure}

\begin{figure}[htbp]
  \vspace{-6pt} 
  \centering
  \includegraphics[width=\textwidth]{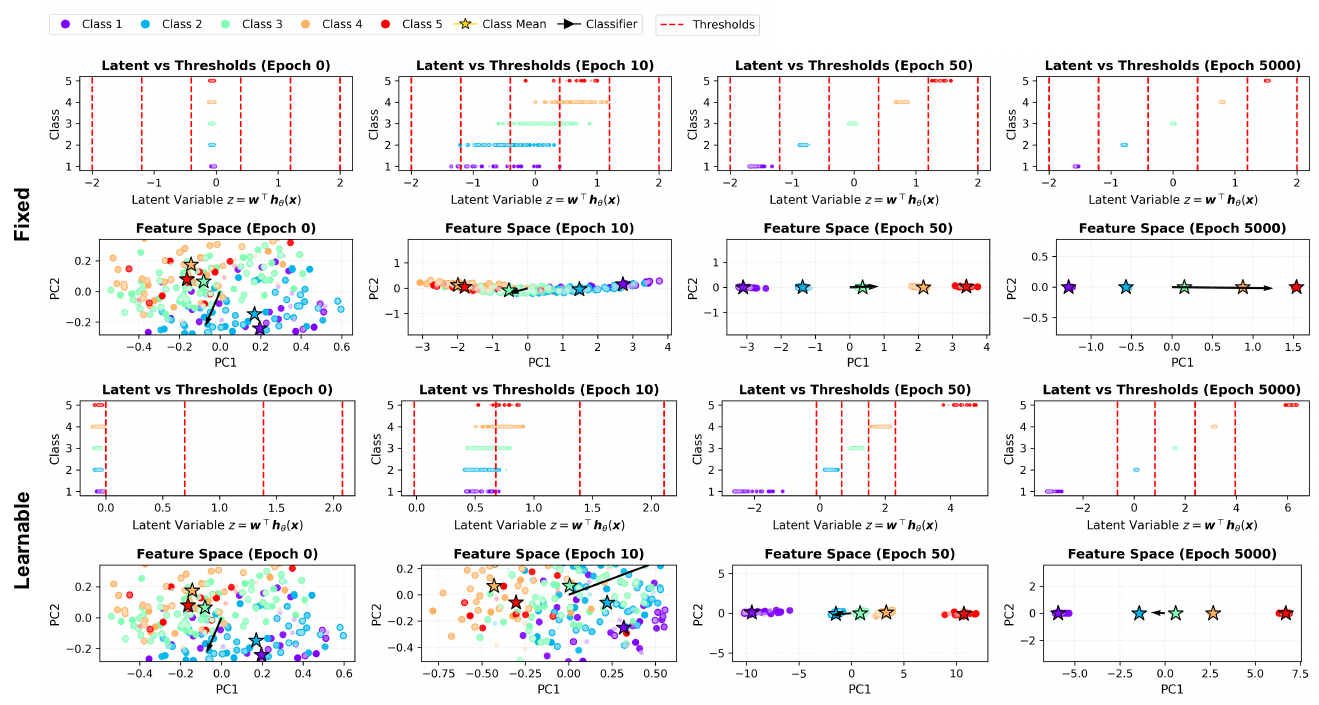}
  \captionsetup{skip=3pt}
  \caption{Visualization of the latent and feature spaces for the LE dataset using the probit model, comparing fixed- and learnable-threshold models.}
  \label{fig:vis_probit_LE}
  \vspace{-6pt} 
\end{figure}

\begin{figure}[htbp]
  \vspace{-6pt} 
  \centering
  \includegraphics[width=\textwidth]{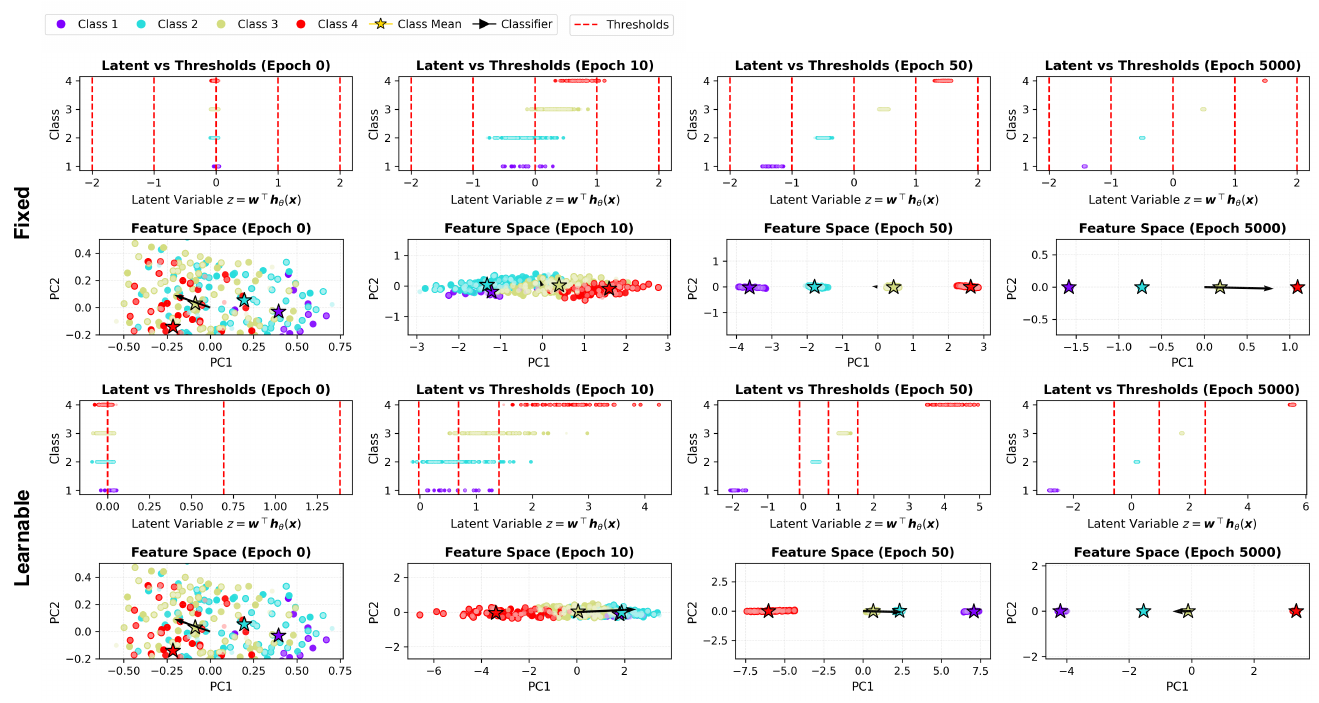}
  \captionsetup{skip=3pt}
  \caption{Visualization of the latent and feature spaces for the SW dataset using the probit model, comparing fixed- and learnable-threshold models.}
  \label{fig:vis_probit_SW}
  \vspace{-6pt} 
\end{figure}

\begin{figure}[htbp]
  \vspace{-6pt} 
  \centering
  \includegraphics[width=\textwidth]{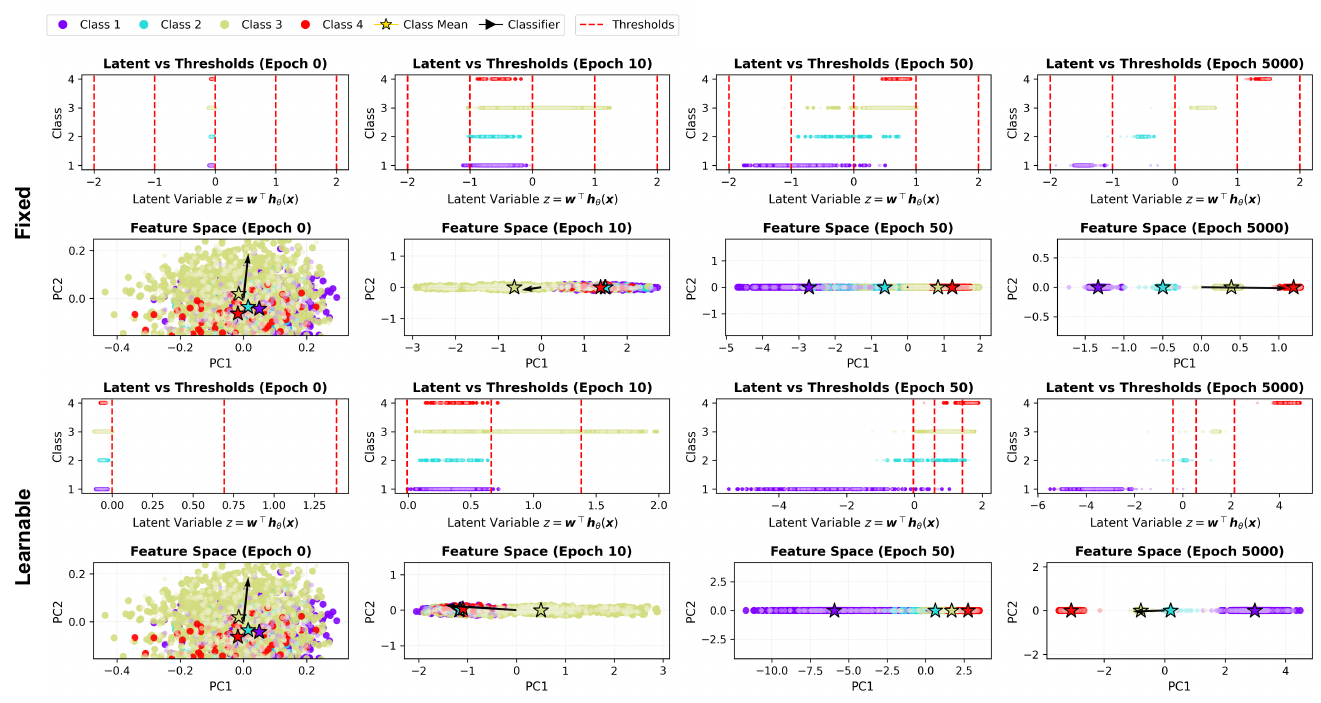}
  \captionsetup{skip=3pt}
  \caption{Visualization of the latent and feature spaces for the CA dataset using the probit model, comparing fixed- and learnable-threshold models.}
  \label{fig:vis_probit_CA}
  \vspace{-6pt} 
\end{figure}

\begin{figure}[htbp]
  \vspace{-6pt} 
  \centering
  \includegraphics[width=\textwidth]{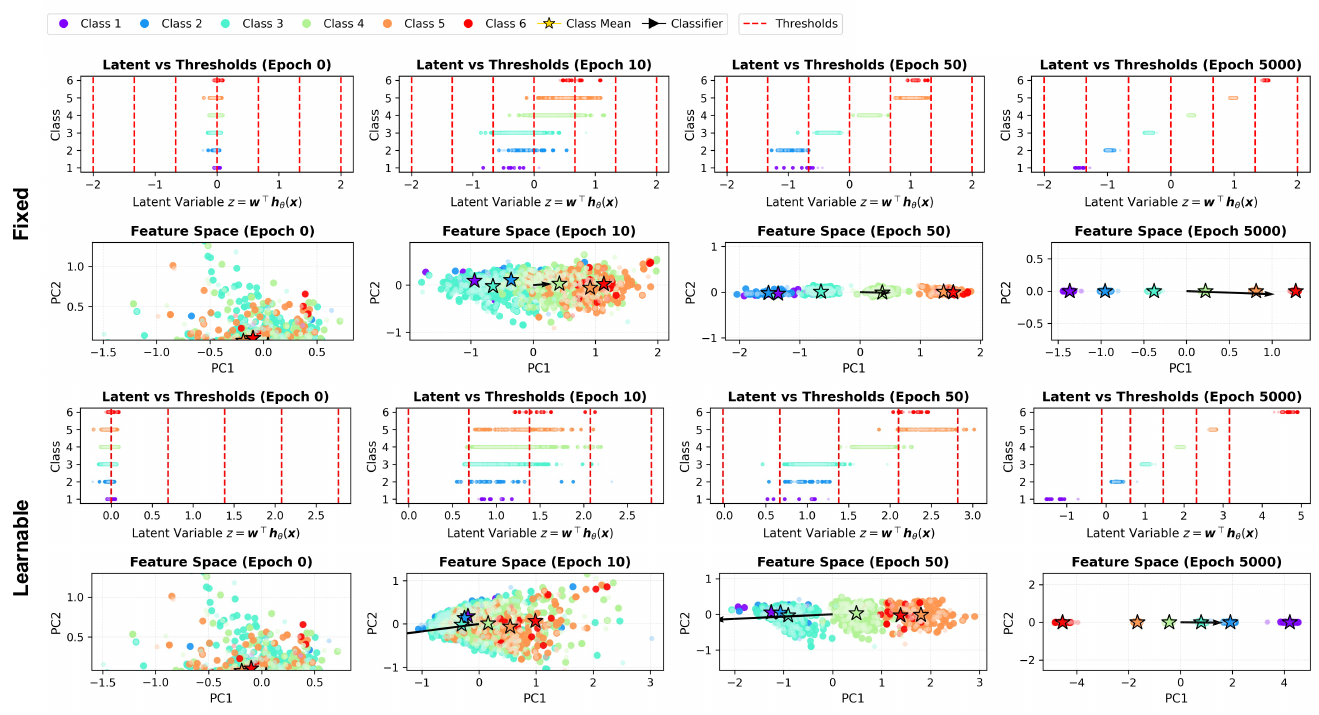}
  \captionsetup{skip=3pt}
  \caption{Visualization of the latent and feature spaces for the WR dataset using the probit model, comparing fixed- and learnable-threshold models.}
  \label{fig:vis_probit_WR}
  \vspace{-6pt} 
\end{figure}

\Lsec{probit}
This section presents the experimental outcomes obtained using the normal CDF (i.e., $g(x)=\Phi(x)$), which corresponds to the probit model. Figures~\ref{fig:metric_curves_probit_ER}--\ref{fig:metric_curves_probit_WR} show the evolution of evaluation-metric curves 
for the datasets ER, LE, SW, CA, and WR, respectively. Figures~\ref{fig:vis_probit_ER}--\ref{fig:vis_probit_WR} show visualization of the latent and feature spaces 
for the datasets ER, LE, SW, CA, and WR, respectively. These again show a consistent behavior with that in the main text.
 

\clearpage
\subsection{Results with the clog-log model}\Lsec{cloglog}
This section presents the experimental outcomes obtained using the Gumbel CDF (i.e., $g(x)=1-e^{-e^x}$), which corresponds to the clog-log model.

Unlike the symmetric link functions (logit and probit), the clog-log link is asymmetric. The solution to \Req{EOS-z^*_zeroreg} thus no longer places optimal latent variables at the simple midpoint between adjacent thresholds, but shifts them by an intrinsic offset. 

As the examined fixed threshold cases so far, we assume the uniformly spaced fixed thresholds $b_q^{\text{fix}} = b_0^{\text{fix}} + q\Delta_b^{\text{fix}}$ where $\Delta_b^{\text{fix}} = (b_Q^{\text{fix}} - b_0^{\text{fix}})/Q$. In this case, we have the optimal solution as 
\be
z_q^* = \frac{b_q^{\text{fix}} + b_{q-1}^{\text{fix}}}{2} + \sigma
\ee
for a constant offset $\sigma$. To verify this, we substitute the above expression into \Req{EOS-z^*_zeroreg}, which requires $g'(b_q^{\text{fix}} - z_q^*) = g'(b_{q-1}^{\text{fix}} - z_q^*)$. Computing the arguments gives:
\be
b_q^{\text{fix}} - z_q^* = b_q^{\text{fix}} - \frac{b_q^{\text{fix}} + b_{q-1}^{\text{fix}}}{2} - \sigma = \frac{\Delta_b^{\text{fix}}}{2} - \sigma,
\ee
and
\be
b_{q-1}^{\text{fix}} - z_q^* = b_{q-1}^{\text{fix}} - \frac{b_q^{\text{fix}} + b_{q-1}^{\text{fix}}}{2} - \sigma = -\frac{\Delta_b^{\text{fix}}}{2} - \sigma.
\ee
Substituting these into \Req{EOS-z^*_zeroreg}, we obtain:
\be
g'\left(\frac{\Delta_b^{\text{fix}}}{2} - \sigma\right) = g'\left(-\frac{\Delta_b^{\text{fix}}}{2} - \sigma\right).
\ee
Since this equation depends only on $\Delta_b^{\text{fix}}$ and not on $q$, the same offset $\sigma$ applies uniformly to all class intervals. In practice, $\sigma$ can be computed as $\sigma = z_1^* - (b_1^{\text{fix}} + b_0^{\text{fix}})/2$ once $z_1^*$ is determined from \Req{EOS-z^*_zeroreg}.

Accordingly, for the uniformly spaced fixed thresholds, we introduce a modified ONC3 metric that accounts for the asymmetry of the link function:
\be
\underline{\mathrm{ONC}_{3,\text{clog-log}}} = 
\frac{
\sum_{q=1}^{Q-1}\left|b_q^{\text{fix}} - b_q^{\text{ideal}}\right|
}{
\sum_{q=1}^{Q-1}(b_{q+1}^{\text{fix}}-b_q^{\text{fix}})
},
\ee
where the ideal threshold position $b_q^{\text{ideal}}$ is defined as
\be
b_q^{\text{ideal}} = \frac{z_q+z_{q+1}}{2} - \sigma.
\ee

\noindent\textbf{Note:} To derive this expression, recall that for uniformly spaced fixed thresholds, the optimal latent variables satisfy $z_q^* = (b_q^{\text{fix}} + b_{q-1}^{\text{fix}})/2 + \sigma$ and $z_{q+1}^* = (b_{q+1}^{\text{fix}} + b_q^{\text{fix}})/2 + \sigma$. Adding these two equations yields:
\be
z_q^* + z_{q+1}^* = \frac{b_q^{\text{fix}} + b_{q-1}^{\text{fix}}}{2} + \frac{b_{q+1}^{\text{fix}} + b_q^{\text{fix}}}{2} + 2\sigma = \frac{b_{q-1}^{\text{fix}} + 2b_q^{\text{fix}} + b_{q+1}^{\text{fix}}}{2} + 2\sigma.
\ee
For uniformly spaced thresholds with $b_{q+1}^{\text{fix}} - b_q^{\text{fix}} = b_q^{\text{fix}} - b_{q-1}^{\text{fix}} = \Delta_b^{\text{fix}}$, we have $b_{q-1}^{\text{fix}} + b_{q+1}^{\text{fix}} = 2b_q^{\text{fix}}$. Substituting this gives:
\be
z_q^* + z_{q+1}^* = \frac{4b_q^{\text{fix}}}{2} + 2\sigma = 2b_q^{\text{fix}} + 2\sigma.
\ee
Rearranging, we obtain:
\be
b_q^{\text{fix}} = \frac{z_q^* + z_{q+1}^*}{2} - \sigma.
\ee
This establishes the form of the ideal threshold position.

In the above analysis, the assumption that fixed thresholds are evenly spaced, but this assumption does not hold for learnable thresholds. Even for symmetric link functions such as the logit model, the same limitation applies; however, in the main text, we have presented the results without resolving this issue. The rationale for this choice is discussed in \Rsec{Discussion}.

Fortunately, for symmetric cases, the error caused by this simplification is not severe, and therefore the conclusions in the main text remain valid. In contrast, for asymmetric link functions, the error introduced by this issue is expected to be more significant. Thus, for the general case--including learnable thresholds--it is necessary to appropriately extend ONC3 and define a new quantity that can properly quantify the phenomenon. However, doing so would make systematic comparison with the existing experiments in the main text difficult, as it would create a mismatch from the experimental setup in the main text. Therefore, for the clog-log model, we only quantify ONC3 for the fixed-threshold case and omit the ONC3 quantification for the learnable-threshold case.

\begin{figure}[!h]
  \vspace{-6pt}                 
  \centering
  \includegraphics[width=\textwidth]{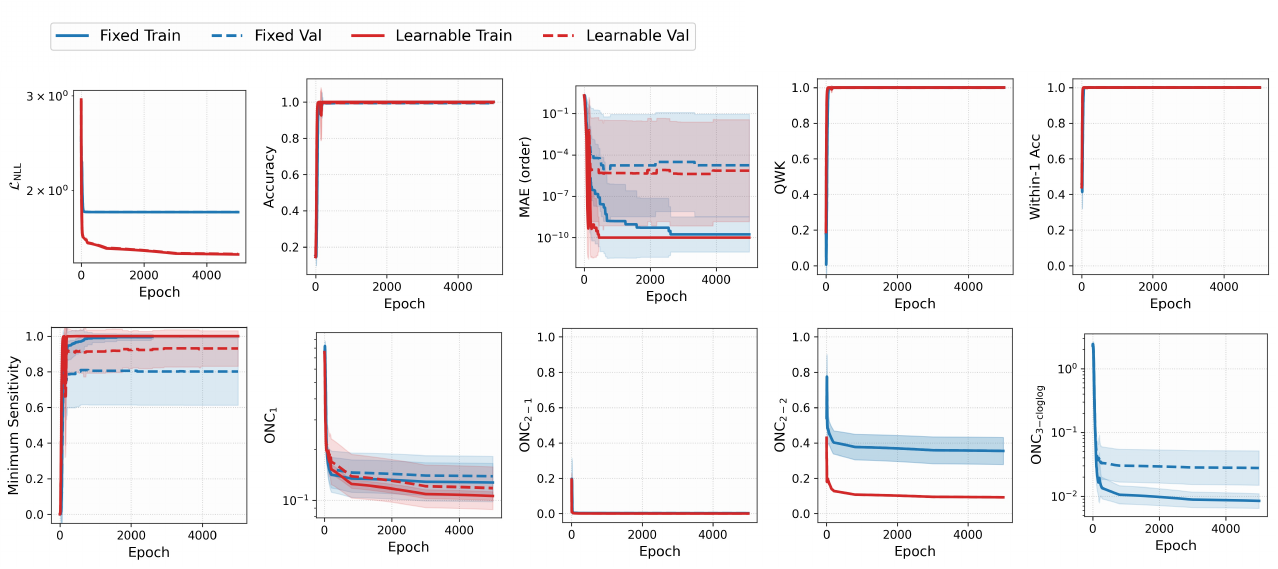}
  \caption{Epoch-wise average metrics curves for the ER dataset with the clog-log model, comparing fixed- and learnable-threshold models.}
  \label{fig:metric_curves_cloglog_ER}
  \vspace{-6pt}                 
\end{figure}

\begin{figure}[!h]
  \vspace{-6pt}                 
  \centering
  \includegraphics[width=\textwidth]{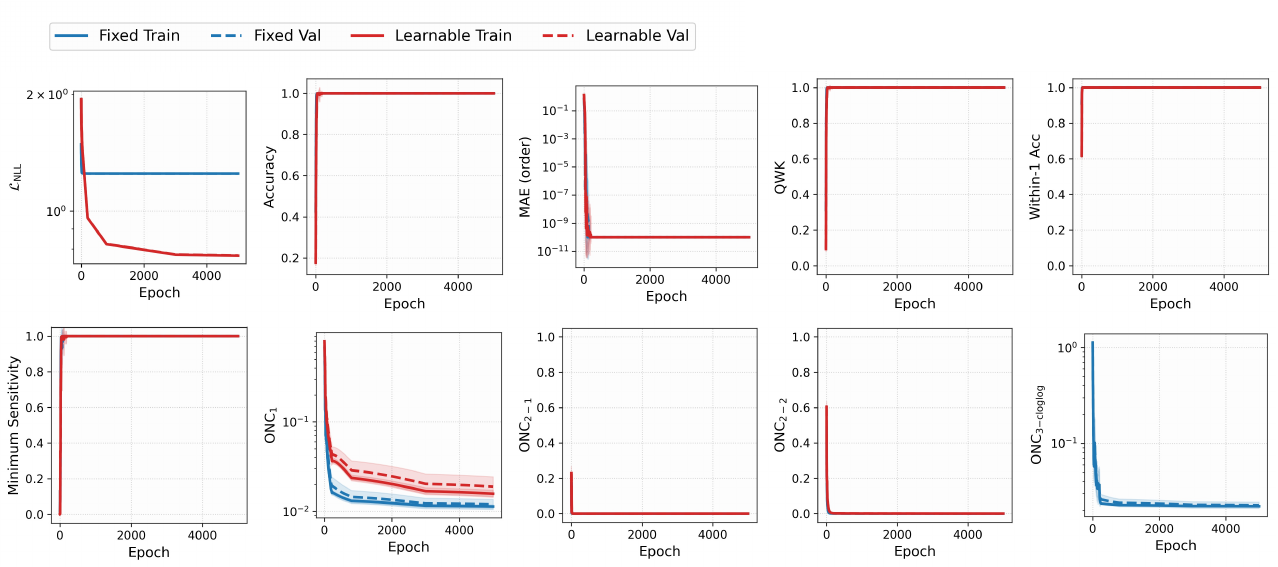}
  \caption{Epoch-wise average metrics curves for the LE dataset with the clog-log model, comparing fixed- and learnable-threshold models.}
  \label{fig:metric_curves_cloglog_LE}
  \vspace{-6pt}                 
\end{figure}

\begin{figure}[!htbp]
  \vspace{-6pt}                 
  \centering
  \includegraphics[width=\textwidth]{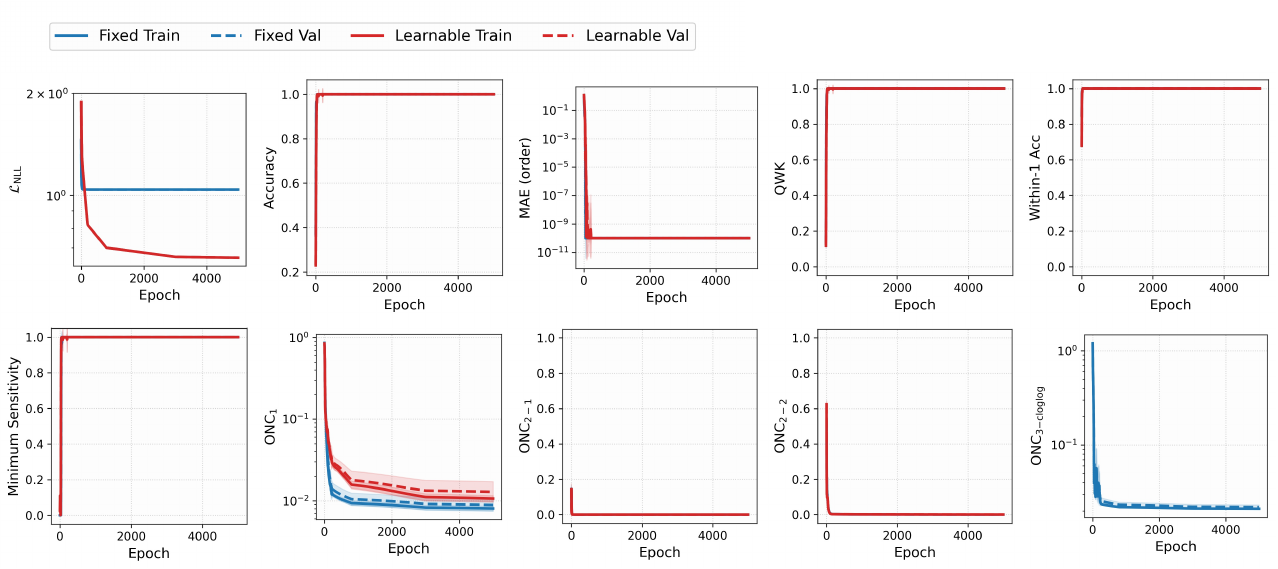}
  \caption{Epoch-wise average metrics curves for the SW dataset with the clog-log model, comparing fixed- and learnable-threshold models.}
  \label{fig:metric_curves_cloglog_SW}
  \vspace{-6pt}                 
\end{figure}

\begin{figure}[!htbp]
  \vspace{-6pt}                 
  \centering
  \includegraphics[width=\textwidth]{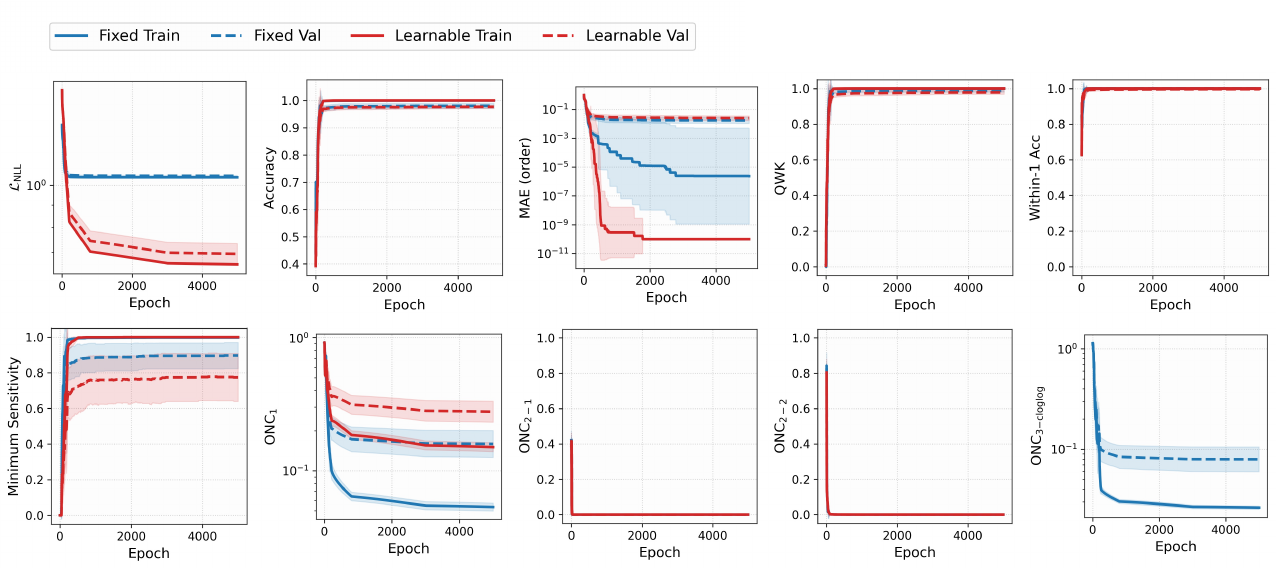}
  \caption{Epoch-wise average metrics curves for the CA dataset with the clog-log model, comparing fixed- and learnable-threshold models.}
  \label{fig:metric_curves_cloglog_CA}
  \vspace{-6pt}                 
\end{figure}

\begin{figure}[!htbp]
  \vspace{-6pt}                 
  \centering
  \includegraphics[width=\textwidth]{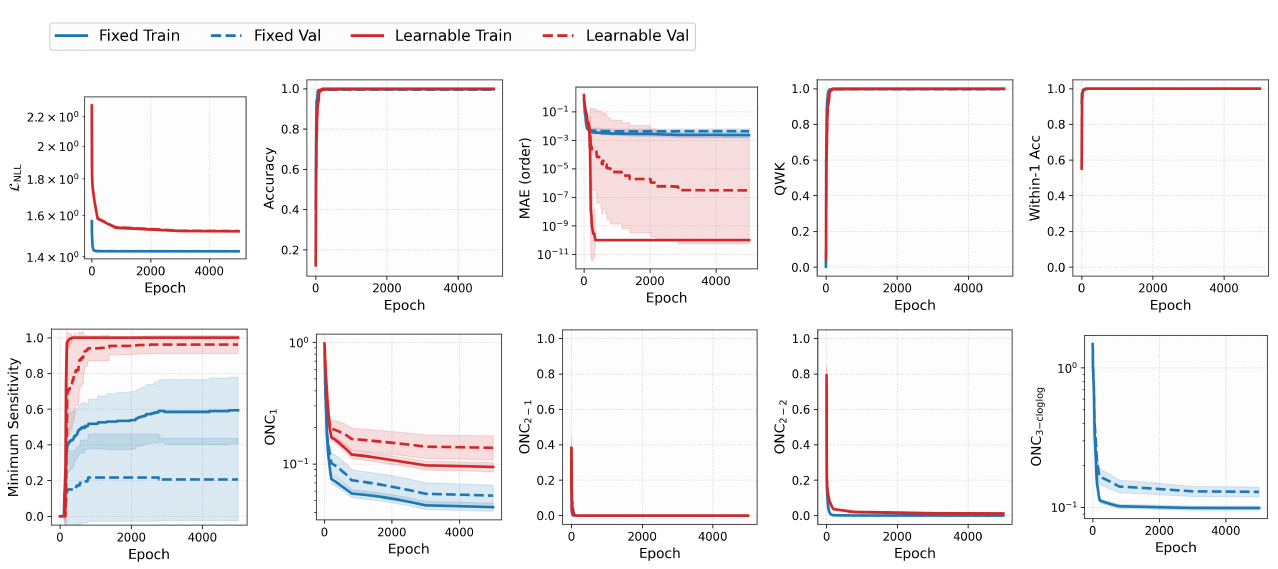}
  \caption{Epoch-wise average metrics curves for the WR dataset with the clog-log model, comparing fixed- and learnable-threshold models.}
  \label{fig:metric_curves_cloglog_WR}
  \vspace{-6pt}                 
\end{figure}

\begin{figure}[htbp]
  \vspace{-6pt} 
  \centering
  \includegraphics[width=\textwidth]{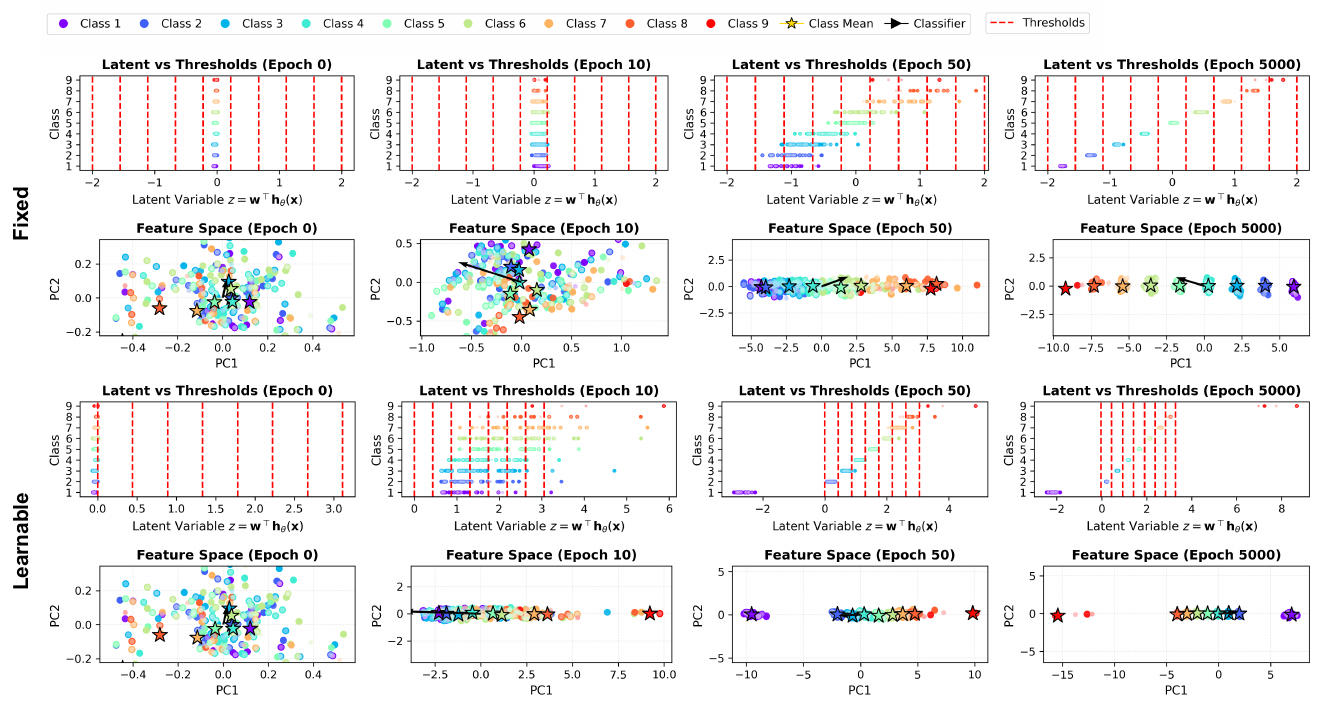}
  \captionsetup{skip=3pt}
  \caption{Visualization of the latent and feature spaces for the ER dataset using the clog-log model, comparing fixed- and learnable-threshold models.}
  \label{fig:vis_cloglog_ER}
  \vspace{-6pt} 
\end{figure}

\begin{figure}[htbp]
  \vspace{-6pt} 
  \centering
  \includegraphics[width=\textwidth]{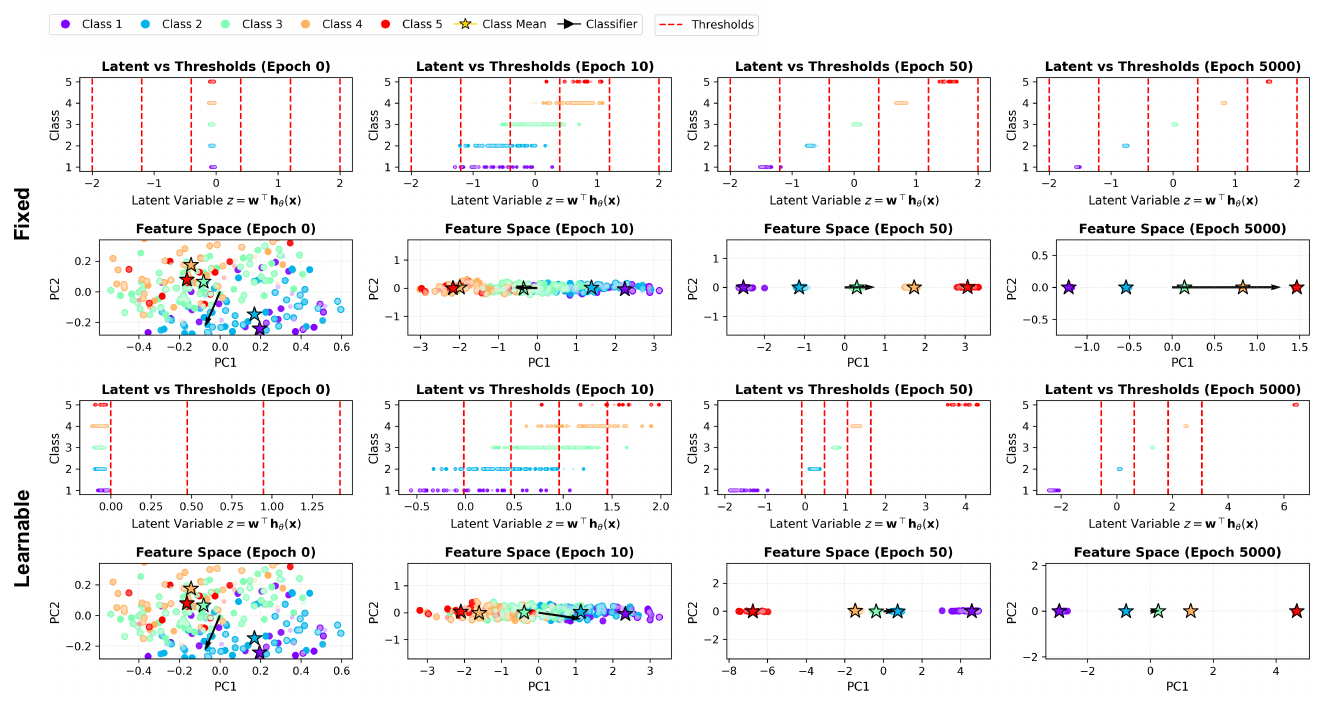}
  \captionsetup{skip=3pt}
  \caption{Visualization of the latent and feature spaces for the LE dataset using the clog-log model, comparing fixed- and learnable-threshold models.}
  \label{fig:vis_cloglog_LE}
  \vspace{-6pt} 
\end{figure}

\begin{figure}[htbp]
  \vspace{-6pt} 
  \centering
  \includegraphics[width=\textwidth]{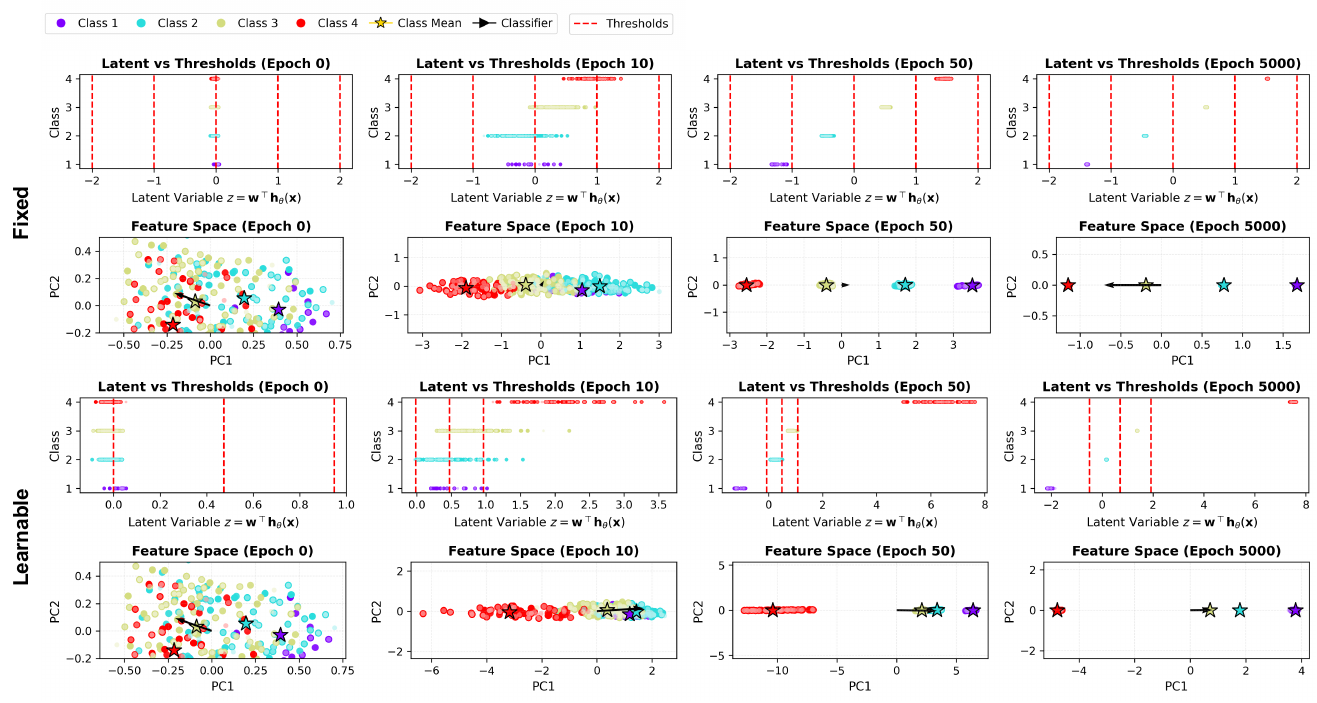}
  \captionsetup{skip=3pt}
  \caption{Visualization of the latent and feature spaces for the SW dataset using the clog-log model, comparing fixed- and learnable-threshold models.}
  \label{fig:vis_cloglog_SW}
  \vspace{-6pt} 
\end{figure}

\begin{figure}[htbp]
  \vspace{-6pt} 
  \centering
  \includegraphics[width=\textwidth]{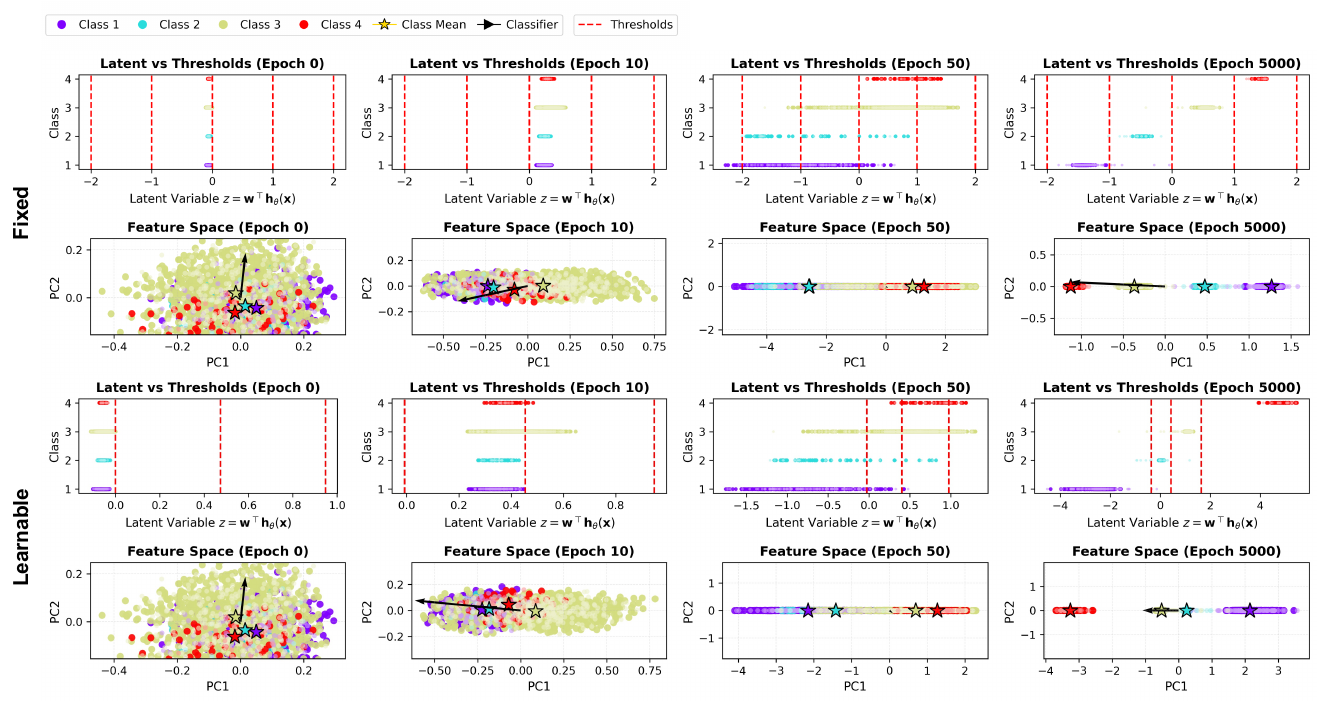}
  \captionsetup{skip=3pt}
  \caption{Visualization of the latent and feature spaces for the CA dataset using the clog-log model, comparing fixed- and learnable-threshold models.}
  \label{fig:vis_cloglog_CA}
  \vspace{-6pt} 
\end{figure}

\begin{figure}[htbp]
  \vspace{-6pt} 
  \centering
  \includegraphics[width=\textwidth]{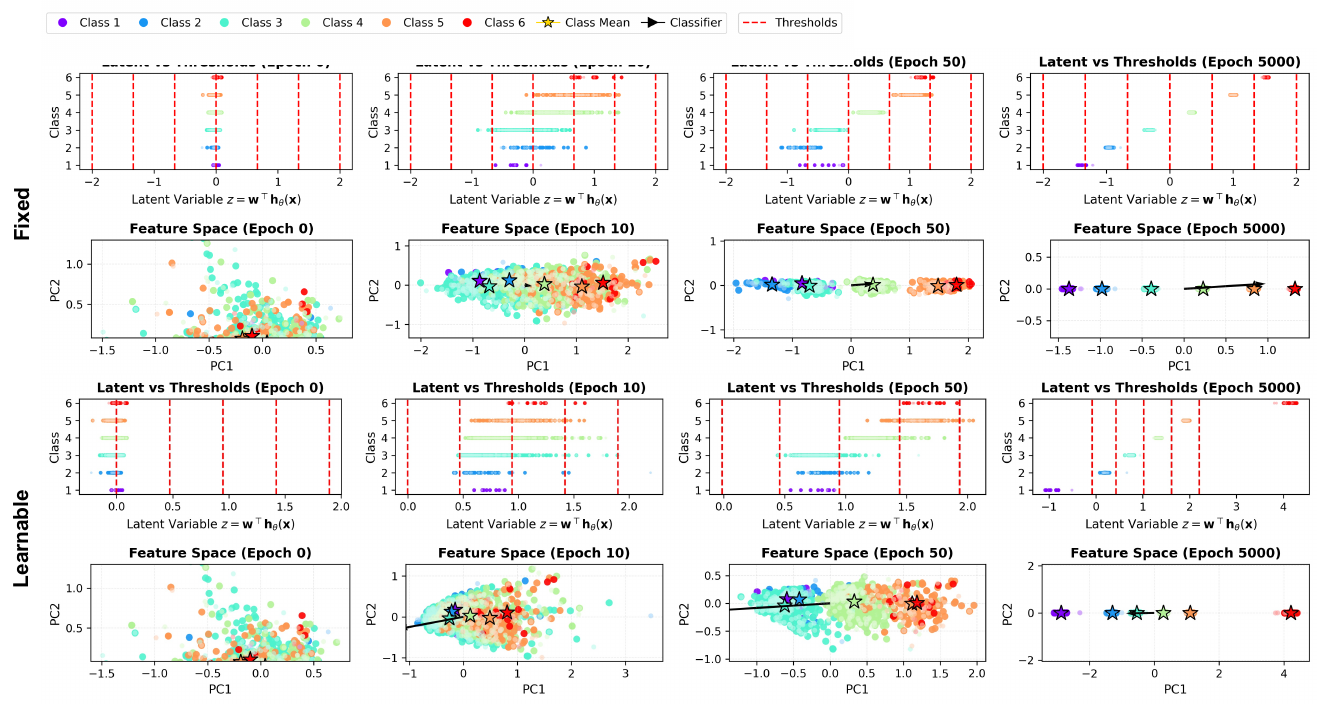}
  \captionsetup{skip=3pt}
  \caption{Visualization of the latent and feature spaces for the WR dataset using the clog-log model, comparing fixed- and learnable-threshold models.}
  \label{fig:vis_cloglog_WR}
  \vspace{-6pt} 
\end{figure}

Figures~\ref{fig:metric_curves_cloglog_ER}--\ref{fig:metric_curves_cloglog_WR} show 
the evolution of evaluation-metric curves for the datasets ER, LE, SW, CA, and WR, respectively.
Due to the limitation for learnable thresholds, we only present $\underline{\mathrm{ONC}_{3,\text{clog-log}}}$ curves for the fixed threshold model. Figures~\ref{fig:vis_cloglog_ER}--\ref{fig:vis_cloglog_WR} 
show visualization of the latent and feature spaces for the datasets ER, LE, SW, CA, and WR, respectively. These results align with the ones in the case of the symmetric link functions, suggesting the wide applicability of ONC.


\clearpage

\subsection{Results on UTKFace dataset with the ResNet50 and DenseNet201 backbones}
\begin{figure}[!htbp]
  \vspace{-6pt}                 
  \centering
  \includegraphics[width=\textwidth]{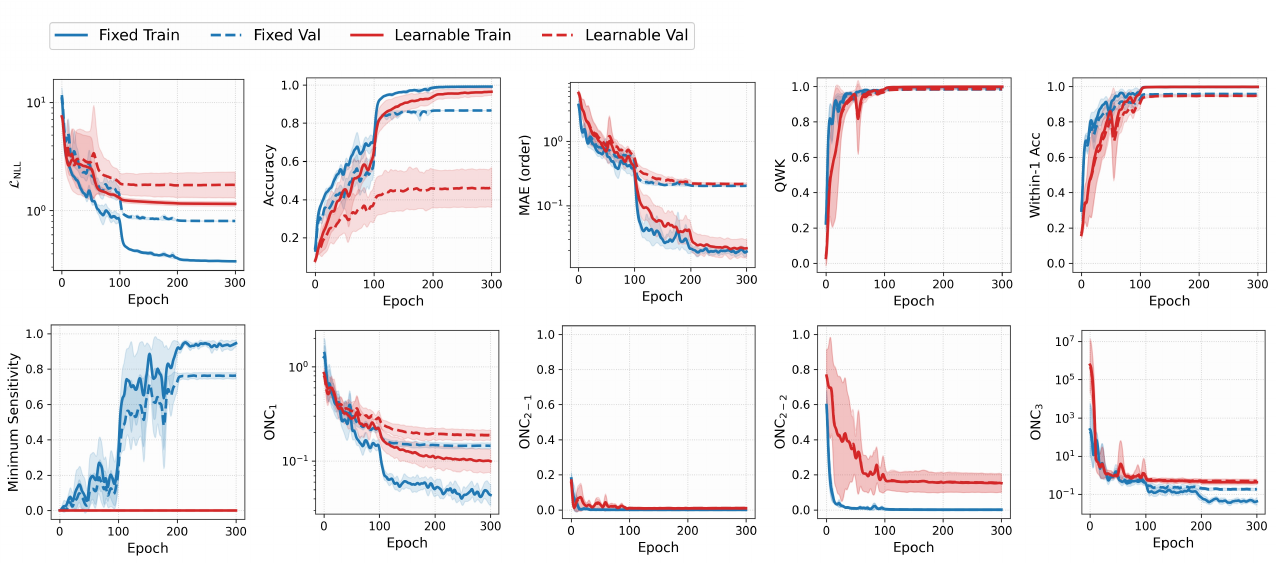}
  \caption{Epoch-wise average metrics curves for the UTKFace dataset with the ResNet50 backbone.}
  \label{fig:metric_curves_utkface_resnet50}
  \vspace{-6pt}                 
\end{figure}

\begin{figure}[!htbp]
  \vspace{-6pt}                 
  \centering
  \includegraphics[width=\textwidth]{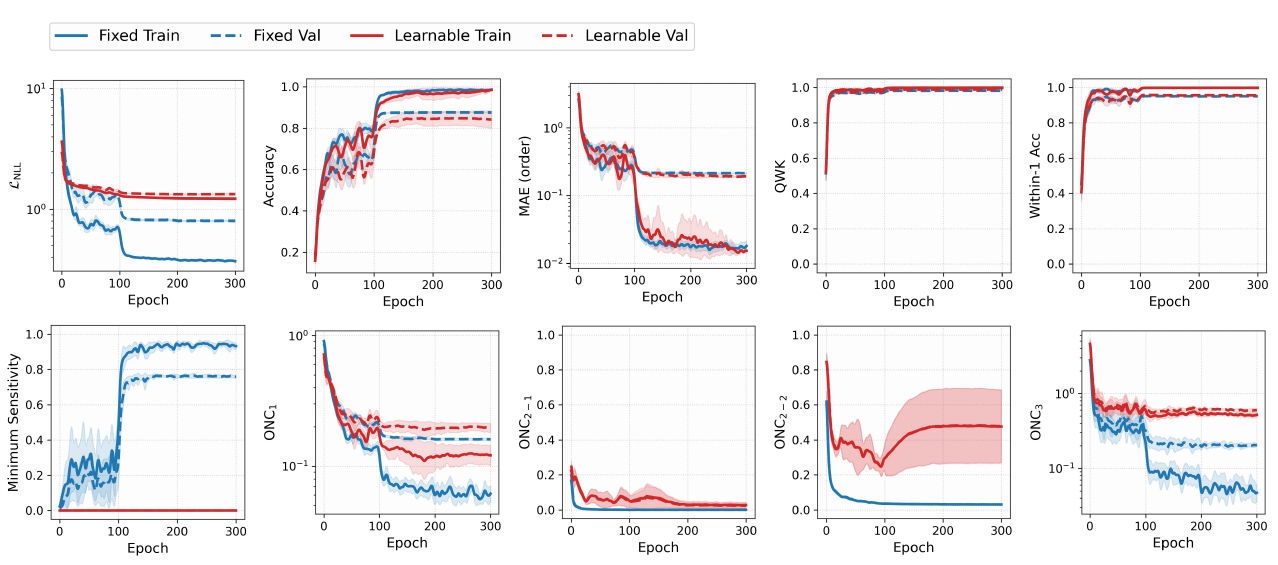}
  \caption{Epoch-wise average metrics curves for the UTKFace dataset with the DenseNet201 backbone.}
  \label{fig:metric_curves_utkface_densenet201}
  \vspace{-6pt}                 
\end{figure}

\begin{figure}[htbp]
  \vspace{-6pt} 
  \centering
  \includegraphics[width=\textwidth]{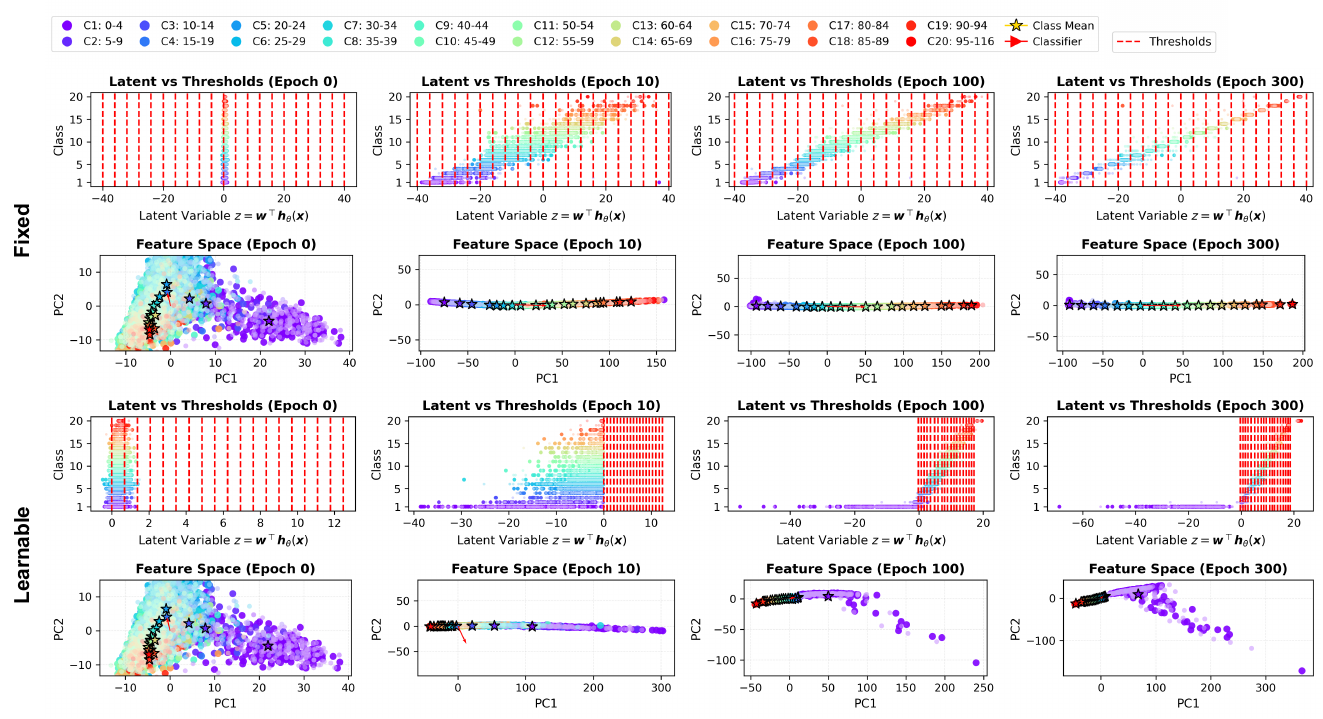}
  \captionsetup{skip=3pt}
  \caption{Visualization of the latent and feature spaces for the UTKFace dataset with the ResNet50 backbone.}
  \label{fig:vis_utkface_resnet50}
  \vspace{-6pt} 
\end{figure}

\begin{figure}[htbp]
  \vspace{-6pt} 
  \centering
  \includegraphics[width=\textwidth]{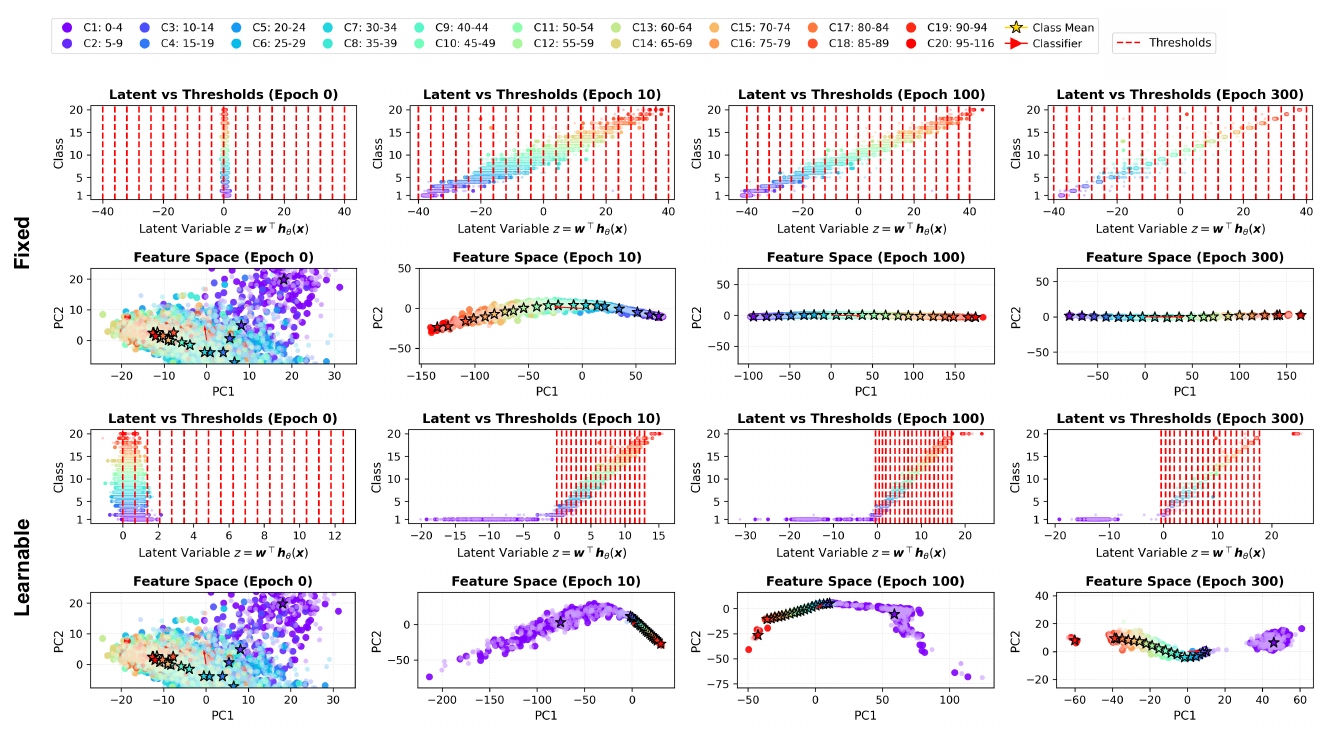}
  \captionsetup{skip=3pt}
  \caption{Visualization of the latent and feature spaces for the UTKFace dataset with the DenseNet201 backbone.}
  \label{fig:vis_utkface_densenet201}
  \vspace{-6pt} 
\end{figure}

\Lsec{utkface}
This section presents the experimental outcomes for the UTKFace dataset using the logit model.
Figures~\ref{fig:metric_curves_utkface_resnet50} and \ref{fig:metric_curves_utkface_densenet201} show the evolution of evaluation-metric curves for the ResNet50 and DenseNet201 backbones, respectively.
Figures~\ref{fig:vis_utkface_resnet50} and \ref{fig:vis_utkface_densenet201} show visualization of the latent and feature spaces for the ResNet50 and DenseNet201 backbones, respectively (one random seed per backbone). These results are essentially consistent with those obtained using ResNet101 in the main text, suggesting that the effectiveness of ONC and fixed thresholds broadly holds even for complex backbone networks used in practice.

\clearpage
\subsection{Comparison of theoretical and experimental results}\label{sec:compare}
We here compare theoretical predictions from EOS~\eqref{eq:EOS} with experimental results on the LE dataset ($Q=5$ classes) as a representative example, assuming the logit model. For the experiment, we use the same network architecture and the dataset splits as the main text, but the training objective is slightly changed to the following one:
\begin{equation}
\mathcal{L}_{\text{total}} = \mathcal{L}_{\rm NLL} + \frac{\lambda_h}{2N}\sum_{i=1}^{N}\|\bm{h}_\theta(\bm{x}_i)\|^{2}_{2} + \frac{\lambda_w}{2}\|\bm{w}\|^{2}_{2} + \lambda_{\theta} \|\bm{\theta}\|^2,
\end{equation}
where $\mathcal{L}_{\rm NLL}$ denotes the negative log-likelihood, and $\lambda_{\theta} = 5 \times 10^{-3}$ and $\lambda_h = 0.01$ are the regularization coefficients for the backbone network parameter $\theta$ and the feature vector $\V{h}_{\theta}$ (output of the backbone network), respectively: the latter regularization term was absent in the experiments in the other sections but is here introduced to make a direct quantitative comparison with our theory. We trained the whole network under the fixed thresholds set as $\bm{b} = (-20, -12, -4, 4, 12, 20)$ for different 35 values of $\lambda_w \in [0, 10^4]$.

\begin{figure}[htbp]
\centering
\includegraphics[width=0.44\textwidth]{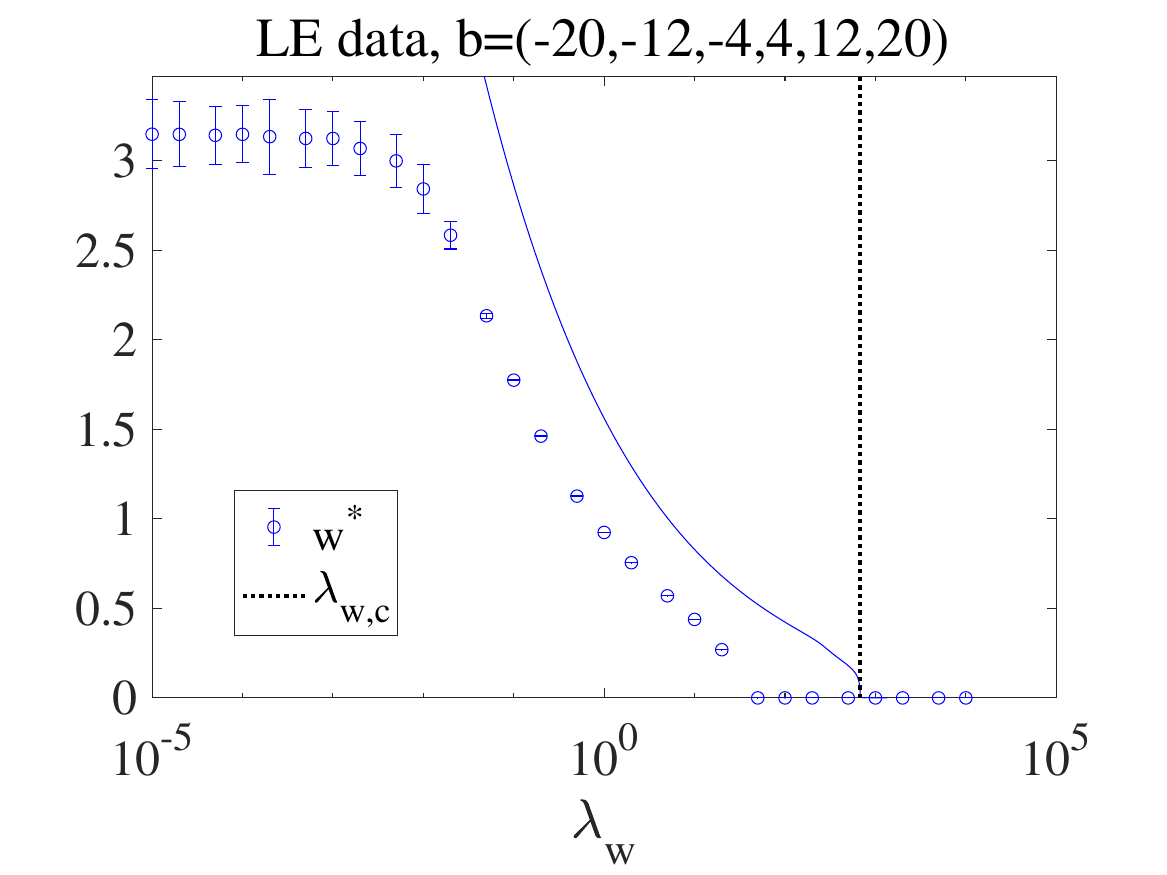}
\includegraphics[width=0.44\textwidth]{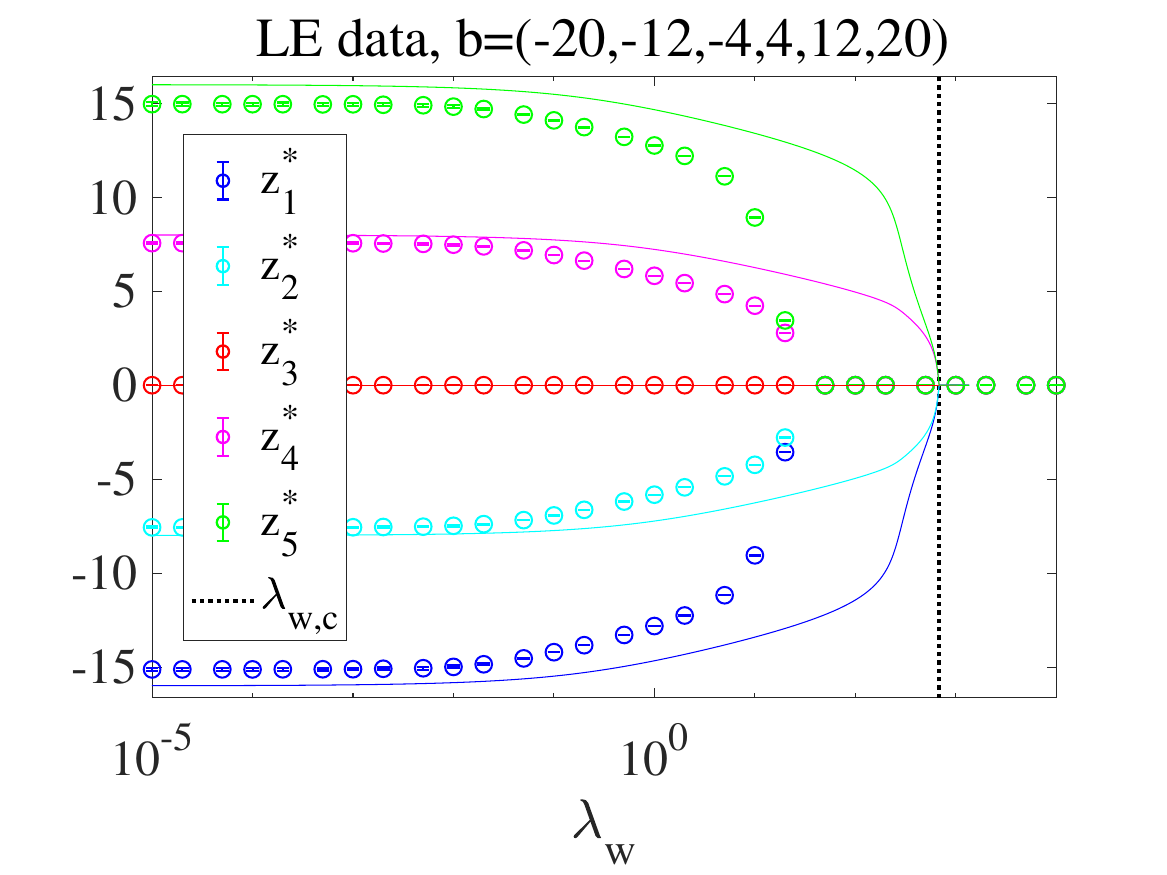}
\caption{Theory (solid curves from EOS) versus experiment (datapoints, 30-holdout average with error bars) for the LE dataset. Left: $\|\bm{w}^*\|$ versus $\lambda_w$. Right: $z_q^*$ versus $\lambda_w$.}
\label{fig:LEV_vs_thr_wz}
\end{figure}

The results of the experiment (datapoints with error bars) are given in \Rfig{LEV_vs_thr_wz}.
The left panel shows $w^*$ while the right one exhibits $\V{z}^*$. In the plots, the theoretical predictions (solid curves) are simultaneously shown: they are computed from EOS~\eqref{eq:EOS} with the above parameter values ($\V{b},\lambda_h$), the dataset size ratios $\V{\alpha} = (0.094, 0.281, 0.472, 0.231, 0.031)$ of the LE dataset, and the respective $\lambda_w$ value. The right panel shows good agreement for $\V{z}^*$ in the small-$\lambda_w$ region where the relation $z_q^* \to (b_q + b_{q-1})/2$ holds, but the left panel exhibits a huge gap in $w^*=\|\bm{w}^*\|_2$: the DNN exhibits a saturating behavior at small $\lambda_w$ while the theory predicts the divergence, implying that our theory is quantitatively not accurate. On the other hand, both the theoretical and experimental results show similar qualitative dependence on $\lambda_w$: they vary monotonically with $\lambda_w$, and in both cases, phase transitions occur at certain specific values of $\lambda_w$, though the locations of the transition points are quantitatively different. We confirmed that similar behavior is also observed for the other four tabular datasets.

In summary, although the theory developed in this study does not quantitatively predict the behavior of actual DNNs with full accuracy, it can qualitatively explain the observed phenomena.


\clearpage
\section*{NeurIPS Paper Checklist}

\begin{enumerate}

\item {\bf Limitations}
    \item[] Question: Does the paper discuss the limitations of the work performed by the authors?
    \item[] Answer: \answerYes{} 
    \item[] Justification: 
      We have provided a paragraph in the Discussion section to
      discuss the limitations of the theoretical results,
      including the assumption of the fixed threshold,
      as well as that on the number of phases
      in the $(\lambda_h,\lambda_w)$-plane.
      Additionally, we acknowledge that the current definition of ONC3 is only valid 
      when the thresholds are fixed and evenly spaced in the latent space, which is 
      the case for our fixed threshold experiments but is not for the others. 
      We note that developing generally applicable ONC3 metrics is possible but 
      the corresponding experiment is left for future work.

\item {\bf Theory assumptions and proofs}
    \item[] Question: For each theoretical result, does the paper provide the full set of assumptions and a complete (and correct) proof?
    \item[] Answer: \answerYes{} 
    \item[] Justification: 
      We have provided all the assumptions for the theorems
      in the main text
      (Theorems~\ref{thm:logcon}, \ref{thm:ONC}, and \ref{thm:EOS}),
      as well as those in the appendices
      (Theorems~\ref{th:Prekopa}, \ref{th:plineq}, \ref{th:slcn}
      and Proposition~\ref{prop:monotone}).
      We have also provided complete proofs to them
      except Theorems~\ref{th:Prekopa} and \ref{th:plineq} (which are not our original contribution), 
      and do believe that they are all correct.

    \item {\bf Experimental result reproducibility}
    \item[] Question: Does the paper fully disclose all the information needed to reproduce the main experimental results of the paper to the extent that it affects the main claims and/or conclusions of the paper (regardless of whether the code and data are provided or not)?
    \item[] Answer: \answerYes{} 
    \item[] Justification: 
    Section~\ref{sec:experiments} and Appendix~\ref{sec:Details of the Experimental Setup} provide all the information necessary for reproducibility. 
    We provide the URL for downloading the datasets in Appendix~\ref{sec:Details of the Experimental Setup}, along with detailed information for tabular datasets (see Table~\ref{tab:datasets}) and UTKFace dataset (see Table~\ref{tab:utkface_class_dist}). 
    Figure~\ref{fig:neural_network} shows the specific structure of the neural network for tabular datasets, and the backbone architectures for UTKFace are detailed in the text. 
    Other settings, including the number of epochs, learning rate (see Tables~\ref{tab:exp-settings} and \ref{tab:utkface-settings}), learning rate scheduling strategy, regularization coefficients, batch size, etc. are also clearly listed in Appendix~\ref{sec:Details of the Experimental Setup}. 
    Loss functions are defined in Section~\ref{sec:Formulation} and evaluation metrics are defined in Section~\ref{sec:experiments}. 
    The link functions, thresholding strategies, and fixed-threshold ranges are also detailed in Tables~\ref{tab:exp-settings} and \ref{tab:utkface-settings}. 
    \textbf{Finally, the supplementary material contains the complete source code, enabling exact reproduction of all the experiments.}
    

\item {\bf Open access to data and code}
    \item[] Question: Does the paper provide open access to the data and code, with sufficient instructions to faithfully reproduce the main experimental results, as described in supplemental material?
    \item[] Answer: \answerYes{} 
    \item[] Justification: 
    Our study used publicly available datasets. 
    For the tabular datasets, the URL for downloading is shown in Appendix~\ref{sec:Details of the Experimental Setup}. 
    The UTKFace dataset is also publicly available.
    We have included the complete experimental code for the tabular datasets with logit and probit models in the supplementary material, along with the necessary instructions and environment setup for reproduction.

\item {\bf Experimental setting/details}
    \item[] Question: Does the paper specify all the training and test details (e.g., data splits, hyperparameters, how they were chosen, type of optimizer, etc.) necessary to understand the results?
    \item[] Answer: \answerYes{} 
    \item[] Justification: Section~\ref{sec:experiments} and Appendix~\NRsec{Details of the Experimental Setup} provide detailed experimental settings.

\item {\bf Experiment statistical significance}
    \item[] Question: Does the paper report error bars suitably and correctly defined or other appropriate information about the statistical significance of the experiments?
    \item[] Answer: \answerYes{} 
    \item[] Justification: 
    For the tabular datasets, publicly available datasets following 30-holdout were used; every experiment was conducted on all the training–validation hold-out splits and the results were averaged, while the metric curves included the corresponding error bands.
    For UTKFace, each configuration was repeated with three different random seeds, and the results were averaged with error bands shown in the metric curves.

\item {\bf Experiments compute resources}
    \item[] Question: For each experiment, does the paper provide sufficient information on the computer resources (type of compute workers, memory, time of execution) needed to reproduce the experiments?
    \item[] Answer: \answerYes{} 
    \item[] Justification: Appendix~\NRsec{Details of the Experimental Setup} lists the computational resources for each experiment, including the GPU type, the GPU memory and runtime of each run, and other relevant details.

    
\item {\bf Code of ethics}
    \item[] Question: Does the research conducted in the paper conform, in every respect, with the NeurIPS Code of Ethics \url{https://neurips.cc/public/EthicsGuidelines}?
    \item[] Answer: \answerYes{} 
    \item[] Justification: 
      We did not do any experiments involving human subjects.
      We have confirmed the copyright information
      of all the datasets used in our experiments, 
      and mentioned it in Appendix~\ref{sec:Details of the Experimental Setup}.
      This paper is of theoretical nature,
      so that there should be no societal impact.

\item {\bf Broader impacts}
    \item[] Question: Does the paper discuss both potential positive societal impacts and negative societal impacts of the work performed?
    \item[] Answer: \answerNA{} 
    \item[] Justification: 
      This paper is of theoretical nature,
      so that there should be no societal impact.
    
\item {\bf Safeguards}
    \item[] Question: Does the paper describe safeguards that have been put in place for responsible release of data or models that have a high risk for misuse (e.g., pretrained language models, image generators, or scraped datasets)?
    \item[] Answer: \answerNA{} 
    \item[] Justification: 
      This paper is of theoretical nature,
      so that there should be no such risks of misuse.

\item {\bf Licenses for existing assets}
    \item[] Question: Are the creators or original owners of assets (e.g., code, data, models), used in the paper, properly credited and are the license and terms of use explicitly mentioned and properly respected?
    \item[] Answer: \answerYes{} 
    \item[] Justification: 
      We have properly credited the creators
      of all the datasets used in our experiments 
      in Appendix~\ref{sec:Details of the Experimental Setup}.
      All the license information has been explicitly mentioned 
      in Appendix~\ref{sec:Details of the Experimental Setup} as well.

\item {\bf New assets}
    \item[] Question: Are new assets introduced in the paper well documented and is the documentation provided alongside the assets?
    \item[] Answer: \answerYes{} 
    \item[] Justification: The code used in the experiments is uploaded in the supplementary material with run scripts, the license, and other related files.

\item {\bf Crowdsourcing and research with human subjects}
    \item[] Question: For crowdsourcing experiments and research with human subjects, does the paper include the full text of instructions given to participants and screenshots, if applicable, as well as details about compensation (if any)? 
    \item[] Answer: \answerNA{} 
    \item[] Justification: 
      We did not do any experiments involving human subjects.

\item {\bf Institutional review board (IRB) approvals or equivalent for research with human subjects}
    \item[] Question: Does the paper describe potential risks incurred by study participants, whether such risks were disclosed to the subjects, and whether Institutional Review Board (IRB) approvals (or an equivalent approval/review based on the requirements of your country or institution) were obtained?
    \item[] Answer: \answerNA{} 
    \item[] Justification: 
      We did not do any experiments involving human subjects.

\item {\bf Declaration of LLM usage}
    \item[] Question: Does the paper describe the usage of LLMs if it is an important, original, or non-standard component of the core methods in this research? Note that if the LLM is used only for writing, editing, or formatting purposes and does not impact the core methodology, scientific rigorousness, or originality of the research, declaration is not required.
    \item[] Answer: \answerNA{}
    \item[] Justification: The core methodology and experimental pipeline of this study do not employ any LLMs. LLMs were used exclusively for manuscript editing, AI search, and routine debugging assistance during code development; hence, they do not constitute an important, original, or non-standard component of the research.

\end{enumerate}

\clearpage

\end{document}